\documentclass[12pt]{article}
\newif\ifall\alltrue 
\newif\ifexpaper\expapertrue 
\newif\ifprivate\privatefalse                

\def\private#1{}            

\def\keywords#1{\small\centerline{\bf Key Words}\vspace{5mm}\centerline{\parbox{14cm}{#1}}}

\def\toinfty#1{\stackrel{#1\to\infty}{\longrightarrow}}
\def\gtapprox{\buildrel{\lower.7ex\hbox{$>$}}\over
                       {\lower.7ex\hbox{$\sim$}}}
\def\nq{\hspace{-1em}}

\def\ignore#1{}

\def\qed{\sqcap\!\!\!\!\sqcup}
\def\1d2{{\textstyle{1\over 2}}}
\def\hbar{h\!\!\!\!^{-}\,}

\def\beq{\begin{equation}}
\def\eeq{\end{equation}}
\def\beqn{\begin{displaymath}}
\def\eeqn{\end{displaymath}}
\def\bqa{\begin{equation}\begin{array}{c}}
\def\eqa{\end{array}\end{equation}}
\def\bqan{\begin{displaymath}\begin{array}{c}}
\def\eqan{\end{array}\end{displaymath}}
\def\pb{\underline}                       
\def\pb#1{\underline{#1}}                 
\def\maxarg{\mathop{\rm maxarg}}          
\def\minarg{\mathop{\rm minarg}}          
\def\hh#1{{\dot{#1}}}                     
\def\best{*}                              
\begin{document}

\begin{titlepage}
\hfill Munich, 31.03.2000


\begin{center}       \vspace*{2cm}
  {\LARGE\bf A Theory of Universal Artificial Intelligence} \\[0.5cm]
  {\LARGE\bf based on Algorithmic Complexity}               \\[2cm]
  {\bf Marcus Hutter\footnotemark}                 \\[1cm]
  {\it Bayerstr. 21, 80335 Munich, Germany} \\[1.5cm]
\end{center}
\footnotetext{Any response to {\tt marcus@hutter1.de} is welcome.}


\begin{abstract}
Decision theory formally solves the problem of rational agents in
uncertain worlds if the true environmental prior probability
distribution is known. Solomonoff's theory of universal induction
formally solves the problem of sequence prediction for unknown
prior distribution. We combine both ideas and get a parameterless
theory of universal Artificial Intelligence. We give strong
arguments that the resulting AI$\xi$ model is the most intelligent
unbiased agent possible. We outline for a number of problem
classes, including sequence prediction, strategic games, function
minimization, reinforcement and supervised learning, how the
AI$\xi$ model can formally solve them. The major drawback of the
AI$\xi$ model is that it is uncomputable. To overcome this
problem, we construct a modified algorithm AI$\xi^{tl}$, which is
still effectively more intelligent than any other time $t$ and
space $l$ bounded agent. The computation time of AI$\xi^{tl}$
is of the order $t\!\cdot\!2^l$. Other discussed topics are formal
definitions of intelligence order relations, the horizon problem
and relations of the AI$\xi$ theory to other AI approaches.
\end{abstract}

\end{titlepage}

{\parskip=0ex\tableofcontents}

\newpage
\section{Introduction}\label{int}

\paragraph{Artificial Intelligence:}
The science of Artificial Intelligence (AI) might be defined as
the construction of intelligent systems and their analysis. A
natural definition of {\it systems} is anything which has an
input and an output stream. Intelligence is more complicated. It
can have many faces like creativity, solving problems, pattern
recognition, classification, learning, induction, deduction,
building analogies, optimization, surviving in an environment,
language processing, knowledge and many more. A formal definition
incorporating every aspect of intelligence, however, seems difficult.
Further, intelligence is graded, there is a smooth transition
between systems, which everyone would agree to be not intelligent
and truely intelligent systems. One simply has to look in nature,
starting with, for instance, inanimate crystals, then come amino-acids,
then some RNA fragments, then viruses, bacteria, plants, animals,
apes, followed by the truly intelligent homo sapiens, and possibly
continued by AI systems or ET's. So the best we can expect to find
is a partial or total order relation on the set of systems, which
orders them w.r.t.\ their degree of intelligence (like
intelligence tests do for human systems, but for a limited class of
problems). Having this order we are, of course, are interested in large
elements, i.e.\ highly intelligent systems. If a largest element
exists, it would correspond to the most intelligent system which
could exist.

Most, if not all known facets of intelligence can be formulated
as goal driven or, more precisely, as maximizing some utility
function. It is, therefore, sufficient to study goal driven AI.
E.g.\ the (biological) goal of animals and humans is to survive and spread.
The goal of AI systems should be to be useful to humans. The
problem is that, except for special cases, we know neither
the utility function, nor the environment in which the
system will operate, in advance.

\paragraph{Main idea:}
We propose a theory which formally\footnote{With a formal solution
we mean a rigorous mathematically definition, uniquely specifying the solution.
In the following, a solution is
always meant in this formal sense.} solves the problem of unknown
goal and environment. It might be viewed as a unification of the ideas of
universal induction, probabilistic planning and reinforcement
learning or as a unification of sequential decision theory with algorithmic
information theory.
We apply this model to some of the facets of intelligence,
including induction, game playing, optimization, reinforcement and supervised
learning, and show how it solves these problem classes. This,
together with general convergence theorems motivates us to
believe that the constructed universal AI system is the best one
in a sense to be clarified in the sequel, i.e. that it is the most
intelligent environmental independent system possible.
The intention of this work is to introduce the universal AI model
and give an in breadth analysis. Most arguments and proofs are
succinct and require slow reading or some additional pencil
work.

\paragraph{Contents:}
{\it Section \ref{secAIfunc}:} The general framework for AI might
be viewed as the design and study of intelligent agents
\cite{Rus95}. An agent is a cybernetic system with some internal
state, which acts with output $y_k$ to some environment in cycle $k$,
perceives some input $x_k$ from the environment and updates its
internal state. Then the next cycle follows. It operates according
to some function $p$. We split the input $x_k$ into a regular part
$x'_k$ and a credit $c_k$, often called reinforcement feedback.
From time to time the environment provides non-zero credit to the
system. The task of the system is to maximize its utility, defined
as the sum of future credits. A probabilistic environment is a
probability distribution $\mu(q)$ over deterministic environments
$q$. Most, if not all environments are of this type. We give a
formal expression for the function $p^\best$, which maximizes in
every cycle the total $\mu$ expected future credit. This model is
called the AI$\mu$ model. As every AI problem can be brought into
this form, the problem of maximizing utility is hence being
formally solved, if $\mu$ is known. There is nothing remarkable or
new here, it is the essence of sequential decision theory
\cite{Che85,Pea88,Neu44}. Notation and formulas needed in
later sections are simply developed. There are two major remaining
problems. The problem of the unknown true prior probability $\mu$
is solved in section \ref{secAIxi}. Computational aspects are
addressed in section \ref{secTime}.

{\it Section \ref{secAImurec}:} Instead of talking about
probability distributions $\mu(q)$ over functions, one could
describe the environment by the conditional probability of
providing inputs $x_1...x_n$ to the system under the condition
that the system outputs $y_1...y_n$. The definition of the optimal
$p^\best$ system in this iterative form is shown to be equivalent
to the previous functional form. The functional form is more
elegant and will be used to define an intelligence order relation
and the time-bounded model in section \ref{secTime}. The iterative
form is more index intensive but more suitable for explicit
calculations and is used in most of the other sections. Further,
we introduce factorizable probability distributions.

{\it Section \ref{secAIxi}:} A special topic is the theory of
induction. In which sense prediction of the future is possible at
all, is best summarized by the theory of Solomonoff. Given the
initial binary sequence $x_1...x_k$, what is the probability of
the next bit being $1$? It can be fairly well predicted by using a
universal probability distribution $\xi$ invented and shown to
converge to the true prior probability $\mu$ by Solomonoff
\cite{Sol64,Sol78} as long as $\mu$ (which needs not be known!) is
computable. The problem of unknown $\mu$ is hence solved for
induction problems. All AI problems where the systems' output does
not influence the environment, i.e. all passive systems are of
this inductive form. Besides sequence prediction (SP),
classification(CF)
is also of this type. Active systems, like game playing (SG) and
optimization (FM), can not be reduced to induction systems. The {\bf
main idea of this work} is to generalize universal induction to
the general cybernetic model described in sections \ref{secAIfunc}
and \ref{secAImurec}. For this, we generalize $\xi$ to include
conditions and replace $\mu$ by $\xi$ in the rational agent model. In this
way the problem that the true prior probability $\mu$ is usually
unknown is solved. Universality of $\xi$ and convergence of
$\xi\!\to\!\mu$ will be shown. These are strong arguments for the
optimality of the resulting AI$\xi$ model. There are certain
difficulties in proving rigorously that and in which sense it is
optimal, i.e. the most intelligent system. Further, we introduce a
universal order relation for intelligence.

{\it Sections \ref{secSP}--\ref{secOther}} show how a number of
AI problem classes fit into the general AI$\xi$ model.
All these problems are formally solved by the AI$\xi$ model.
The solution is, however, only formal because
the AI$\xi$ model developed thus far is
uncomputable or, at best, approximable. These sections should support
the claim that every AI problem can be formulated (and hence
solved) within the AI$\xi$ model. For some classes we give
concrete examples to illuminate the
scope of the problem class. We first formulate each problem class
in its natural way (when $\mu^{\mbox{\tiny problem}}$ is known) and
then construct a formulation within the AI$\mu$ model and prove
its equivalence. We then consider the consequences of
replacing $\mu$ by $\xi$. The main goal is to understand why and
how the problems are solved by AI$\xi$. We only highlight special
aspects of each problem class. Sections
\ref{secSP}--\ref{secOther} together should give a better picture
of the AI$\xi$ model. We do not study every aspect for every
problem class. The sections might be read selectively. They are
not necessary to understand the remaining sections.

{\it Section \ref{secSP}:} Using the AI$\mu$ model for sequence
prediction (SP) is identical to Baysian sequence prediction
SP$\Theta_\mu$. One might expect, when using the AI$\xi$ model for
sequence prediction, one would recover exactly the universal
sequence prediction scheme SP$\Theta_\xi$, as AI$\xi$ was a unification of the
AI$\mu$ model and the idea of universal probability $\xi$. Unfortunately
this is not the case. One reason is that $\xi$ is only a
probability distribution in the inputs $x$ and not in the outputs
$y$. This is also one of the origins of the difficulty of proving error/credit
bounds for AI$\xi$. Nevertheless, we argue that AI$\xi$ is
equally well suited for sequence prediction as SP$\Theta_\xi$ is.
In a very limited setting we prove a (weak) error bound for
AI$\xi$ which gives hope that a general proof is attainable.

{\it Section \ref{secSG}:} A very important class of problems are
strategic games (SG). We restrict ourselves to deterministic strictly
competitive strategic games like chess. If the environment is a
minimax player, the AI$\mu$ model itself reduces to a minimax
strategy. Repeated games of fixed lengths are a special case for
factorizable $\mu$. The consequences of variable game length is
sketched. The AI$\xi$ model has to learn the rules of the game
under consideration, as it has no prior information about these
rules. We describe how AI$\xi$ actually learns these rules.

{\it Section \ref{secFM}:} There are many problems that fall into
the category 'resource bounded function minimization' (FM). They
include the Traveling Salesman Problem, minimizing production
costs, inventing new materials or even producing, e.g. nice
paintings, which are (subjectively) judged by a human. The task is to
(approximately) minimize some function $f\!:\!Y\!\to\!Z$ within
minimal number of function calls. We will see that a greedy model
trying to minimize $f$ in every cycle fails. Although the greedy
model has nothing to do with downhill or gradient techniques
(there is nothing like a gradient or direction for functions over
$Y$) which are known to fail, we discover the same difficulties.
FM has already nearly the full complexity of
general AI. The reason being that FM can actively influence the
information gathering process by its trials $y_k$ (whereas SP and
CF cannot). We discuss in detail the optimal FM$\mu$ model and
its inventiveness in choosing the $y\!\in\!Y$. A discussion of the subtleties when
using AI$\xi$ for function minimization, follows.

{\it Section \ref{secEX}:} Reinforcement learning, as the
AI$\xi$ model does, is an important learning technique but not the only one.
To improve the speed of learning, supervised learning, i.e.
learning by acquiring knowledge, or learning from a constructive
teacher is necessary. We show, how AI$\xi$ learns to learn
supervised. It actually establishes supervised learning very
quickly within $O(1)$ cycles.

{\it Section \ref{secOther}} gives a brief survey of other general
aspects, ideas and methods in AI, and their connection to the
AI$\xi$ model. Some aspects are directly included, others are or
should be emergent.

{\it Section \ref{secTime}:} Up to now we have shown the universal
character of the AI$\xi$ model but have completely ignored
computational aspects. Let us assume that there exists some
algorithm $\tilde p$ of size $\tilde l$ with computation time per
cycle $\tilde t$, which behaves in a sufficiently intelligent way
(this assumption is the very basis of AI). The
algorithm $p^\best$ should run all algorithms of length
$\leq\!\tilde l$ for $\tilde t$ time steps in every cycle and select the best
output among them. So we have an algorithm which runs in time
$\tilde l\!\cdot\!2^{\tilde t}$ and is at least as good as $\tilde
p$, i.e.\ it also serves our needs apart from the (very large
but) constant multiplicative factor in computation time. This idea
of the 'typing monkeys', one of them eventually producing 'Shakespeare', is
well known and widely used in theoretical computer science. The
difficult part is the selection of the algorithm with the best
output. A further complication is that the selection process
itself must have only limited computation time. We present a
suitable modification of the AI$\xi$ model which solves these
difficult problems. The solution is somewhat involved from an
implementational aspect. An implementation would include first
order logic, the definition of a Universal Turing machine within
it and proof theory. The assumptions behind this construction are
discussed at the end.

{\it Section \ref{secOutlook}} contains some discussion of
otherwise unmentioned topics and some (personal) remarks. It also
serves as an outlook to further research.

{\it Section \ref{secCon}} contains the conclusions.

\paragraph{History \& References:}
Kolmogorov65 \cite{Kol65} suggested to define the information
content of an object as the length of the shortest program
computing a representation of it. Solomonoff64 \cite{Sol64}
invented the closely related universal prior probability
distribution and used it for binary sequence prediction
\cite{Sol64,Sol78} and function inversion and minimization
\cite{Sol86}. Together with Chaitin66\&75 \cite{Cha66,Cha75} this
was the invention of what is now called Algorithmic Information
theory. For further literature and many applications see
\cite{LiVi93}. Other interesting 'applications' can be found in
\cite{Cha91,Sch99,Vov98}. Related topics are the Weighted Majority
Algorithm invented by Littlestone and Warmuth89 \cite{LiWa89},
universal forecasting by Vovk92 \cite{Vov92}, Levin search73
\cite{Lev73}, pac-learning introduced by Valiant84 \cite{Val84}
and Minimum Description Length \cite{LiVi92,Ris89}. Resource
bounded complexity is discussed in \cite{Dal73,Fed92,Ko86,Pin97},
resource bounded universal probability in \cite{LiVi91,LiVi93}.
Implementations are rare \cite{Con97,Sch95,Sch96}. Excellent
reviews with a philosophical touch are \cite{LiVi92a,Sol97}. For
an older, but general review of inductive inference see Angluin83
\cite{Ang83}. For an excellent introduction into algorithmic
information theory, further literature and many applications one
should consult the book of Li and Vit\'anyi97 \cite{LiVi93}. The
survey \cite{LiVi92} or the chapters 4 and 5 of \cite{LiVi93}
should be sufficient to follow the arguments and proofs
in this paper. 
The other ingredient in our AI$\xi$ model is sequential decision theory. We
do not need much more than the maximum expected utility principle
and the expecimax algorithm \cite{Mic66,Rus95}. The book of von Neumann and
Morgenstern44 \cite{Neu44} might be seen as the initiation of
game theory, which already contains the expectimax algorithm
as a special case. The literature on decision theory is
vast and we only give two possibly interesting references with
regard to this paper. Cheeseman85\&88 \cite{Che85} is a defense
of the use of probability theory in AI. Pearl88 \cite{Pea88} is a
good introduction and overview of probabilistic reasoning.

\newpage
\section{The AI$\mu$ Model in Functional Form}\label{secAIfunc}

\paragraph{The cybernetic or agent model:}
A good way to start thinking about intelligent systems is to
consider more generally cybernetic systems, in AI usually called
agents. This avoids having to struggle with the meaning of
intelligence from the very beginning. A cybernetic system is a
control circuit with input $y$ and output $x$ and an internal
state. From an external input and the internal state the system
calculates deterministically or stochastically an output. This
output (action) modifies the environment and leads to a new input
(reception). This continues ad infinitum or for a finite number of
cycles. As explained in the last section, we need some credit
assignment to the cybernetic system. The input $x$ is divided into
two parts, the standard input $x'$ and some credit input $c$. If
input and output are represented by strings, a deterministic
cybernetic system can be modeled by a Turing machine $p$. $p$ is
called the policy of the agent, which determines the action to a
receipt. If the environment is also computable it might be modeled
by a Turing machine $q$ as well. The interaction of the agent
with the environment can be illustrated as follows:

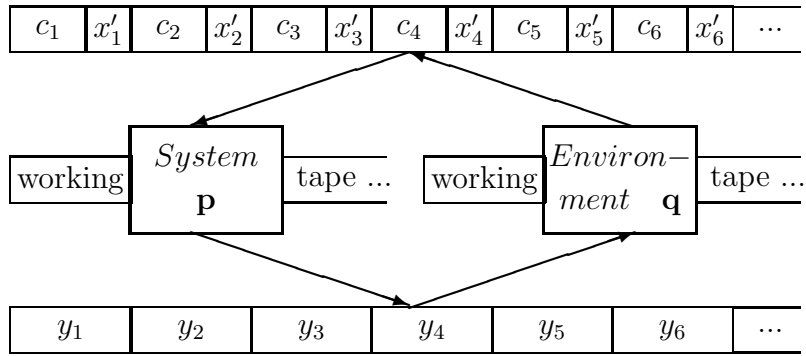
\begin{center}\label{cyberpic}
\special{em:linewidth 0.4pt}
\linethickness{0.4pt}
\begin{picture}(106,47)
\thinlines
\put(1,41){\framebox(10,6)[cc]{$c_1$}}
\put(11,41){\framebox(6,6)[cc]{$x'_1$}}
\put(17,41){\framebox(10,6)[cc]{$c_2$}}
\put(27,41){\framebox(6,6)[cc]{$x'_2$}}
\put(33,41){\framebox(10,6)[cc]{$c_3$}}
\put(43,41){\framebox(6,6)[cc]{$x'_3$}}
\put(49,41){\framebox(10,6)[cc]{$c_4$}}
\put(59,41){\framebox(6,6)[cc]{$x'_4$}}
\put(65,41){\framebox(10,6)[cc]{$c_5$}}
\put(75,41){\framebox(6,6)[cc]{$x'_5$}}
\put(81,41){\framebox(10,6)[cc]{$c_6$}}
\put(91,41){\framebox(6,6)[cc]{$x'_6$}}
\put(102,44){\makebox(0,0)[cc]{...}}
\put(1,1){\framebox(16,6)[cc]{$y_1$}}
\put(17,1){\framebox(16,6)[cc]{$y_2$}}
\put(33,1){\framebox(16,6)[cc]{$y_3$}}
\put(49,1){\framebox(16,6)[cc]{$y_4$}}
\put(65,1){\framebox(16,6)[cc]{$y_5$}}
\put(81,1){\framebox(16,6)[cc]{$y_6$}}
\put(102,4){\makebox(0,0)[cc]{...}}
\put(97,47){\line(1,0){9}}
\put(97,41){\line(1,0){9}}
\put(97,7){\line(1,0){9}}
\put(97,1){\line(0,0){0}}
\put(97,1){\line(1,0){9}}
\put(1,21){\framebox(16,6)[cc]{working}}
\thicklines
\put(17,17){\framebox(20,14)[cc]{$\displaystyle{System\atop\bf p}$}}
\thinlines
\put(37,27){\line(1,0){14}}
\put(37,21){\line(1,0){14}}
\put(39,24){\makebox(0,0)[lc]{tape ...}}
\put(56,21){\framebox(16,6)[cc]{working}}
\thicklines
\put(72,17){\framebox(20,14)[cc]{$\displaystyle{Environ-\atop ment\quad\bf q}$}}
\thinlines
\put(92,27){\line(1,0){14}}
\put(92,21){\line(1,0){14}}
\put(94,24){\makebox(0,0)[lc]{tape ...}}
\thicklines
\put(54,41){\vector(-3,-1){29}}
\put(84,31){\vector(-3,1){30}}
\put(54,7){\vector(3,1){30}}
\put(25,17){\vector(3,-1){29}}
\end{picture}
\end{center}

$p$ as well as $q$ have unidirectional input and output tapes and
bidirectional working tapes. What entangles the agent with the
environment, is the fact that the upper tape serves as input tape
for $p$, as well as output tape for $q$, and that the lower tape
serves as output tape for $p$ as well as input tape for $q$.
Further, the reading head must always be left of the writing head,
i.e. the symbols must first be written, before they are read. $p$
and $q$ have their own mutually inaccessible working tapes
containing their own 'secrets'. The heads move in the following
way. In the k$^{th}$ cycle $p$ writes $y_k$, $q$ reads $y_k$, $q$
writes $x_k\!\equiv\!c_kx_k'$, $p$ reads $x_k\!\equiv\!c_kx_k'$,
followed by the $(k+1)^{th}$ cycle and so on. The whole process
starts with the first cycle, all heads on tape start and working
tapes being empty. We want to call Turing machines behaving in
this way, {\it chronological Turing machines}, for obvious
reasons. Before continuing, some notations on strings are
appropriate.

\paragraph{Strings:}
We will denote strings over the alphabet $X$ by
$s\!=\!x_1x_2...x_n$, with $x_k\!\in\!X$, where $X$ is
alternatively interpreted as a non-empty subset of $I\!\!N$ or
itself as a prefix free set of binary strings.
$l(s)=l(x_1)\!+...+\!l(x_n)$ is the length of s. Analogous
definitions hold for $y_k\!\in\!Y$. We call $x_k$ the $k^{th}$
input word and $y_k$ the $k^{th}$ output word (rather than
letter). The string $s=y_1x_1...y_nx_n$ represents the
input/output in chronological order. Due to the prefix property of
the $x_k$ and $y_k$, $s$ can be uniquely separated into its words.
The words appearing in strings are always in chronological order.
We further introduce the following abbreviations: $\epsilon$ is the
empty string, $x_{n:m}:=x_nx_{n+1}...x_{m-1}x_m$ for $n\leq m$ and
$\epsilon$ for $n>m$. $x_{<n}:=x_1... x_{n-1}$. Analog for $y$.
Further, $y\!x_n\!:=y_nx_n$, $y\!x_{n:m}\!:=\!y_nx_n...y_mx_m$,
and so on.

\paragraph{AI model for known deterministic environment:}
Let us define for the chronological Turing machine $p$ a partial
function also named $p\!:\!X^*\!\rightarrow\!Y^*$ with
$y_{1:k}=p(x_{<k})$ where $y_{1:k}$ is the output of Turing
machine $p$ on input $x_{<k}$ in cycle k, i.e. where $p$ has read
up to $x_{k-1}$ but no further. In an analogous way, we define
$q\!:\!Y^*\!\rightarrow\!X^*$ with $x_{1:k}=q(y_{1:k})$.
Conversely, for every partial recursive chronological function we
can define a corresponding chronological Turing machine. Each
(system,environment) pair $(p,q)$ produces a unique I/O sequence
$\omega(p,q):=y_1^{pq}x_1^{pq}y_2^{pq}x_2^{pq}...$. When we look
at the definition of $p$ and $q$ we see a nice symmetry between
the cybernetic system and the environment. Until now, not much
intelligence is in our system. Now the credit assignment comes
into the game and removes the symmetry somewhat. We split the
input $x_k\!\in\!X\!:=\!C\!\times\!X'$ into a regular part
$x_k'\!\in\!X'$ and a credit $c_k\!\in\!C\!\subset\!I\!\!R$. We
define $x_k\!\equiv\!c_kx_k'$ and $c_k\equiv c(x_k)$. The goal of
the system should be to maximize received credits. This is called
reinforcement learning. The reason for the asymmetry is, that
eventually we (humans) will be the environment with which the
system will communicate and {\it we} want to dictate what is good
and what is wrong, not the other way round. This one way learning,
the system learns from the environment, and not conversely,
neither prevents the system from becoming more intelligent than the
environment, nor does it prevent the environment learning from
the system because the environment can itself interpret the
outputs $y_k$ as a regular and a credit part. The environment is
just not forced to learn, whereas the system is. In cases where we
restrict the credit to two values
$c\!\in\!C\!=\!I\!\!B\!:=\!\{0,1\}$, $c\!=\!1$ is interpreted as a
positive feedback, called {\it good} or {\it correct} and
$c\!=\!0$ a negative feedback, called {\it bad} or {\it error} in
the following. Further, let us restrict for a while the lifetime
(number of cycles) $T$ of the system to a large, but finite value.
Let $C_{km}(p,q)\!:=\!\sum_{i=k}^mc(x_i)$ be the total credit, the
system $p$ receives from the environment $q$ in the cycles $k$ to
$m$. It is now natural to call the system, which maximizes the
total credit $C_{1T}$, called utility, the {\it best} or {\it most intelligent}
one\footnote {$\maxarg_p C(p)$ is the $p$ which maximizes
$C(\cdot)$. If there is more than one maximum we might choose the
lexicographically smallest one for definiteness.}.
\beqn
 p^{\best,T,q}=\maxarg_p C_{1T}(p,q) \quad\Rightarrow\quad
 C_{kT}(p^{\best,T,q},q) \geq C_{kT}(p,q) \quad \forall p
\eeqn
For $k\!=\!1$ this is obvious and for $k\!>\!1$ easy to see.
If $T$, $Y$ and $X$ are finite, the number of different behaviours
of the system, i.e. the search space is finite. Therefore, because
we have assumed that $q$ is known, $p^{\best,T,q}$ can effectively
be determined (by pre-analyzing all behaviours). The main reason
for restricting to finite $T$ was not to ensure computability of
$p^{\best,T,q}$ but that the limit $T\!\to\infty$ might not exist.
This is nothing special, the (unrealistic) assumption of a
completely known deterministic environment $q$ has simply trivialized
everything.
\paragraph{AI model for known prior probability:}
Let us now weaken our assumptions by replacing the environment $q$
with a probability distribution $\mu(q)$ over chronological functions.
$\mu$ might be interpreted
in two ways. Either the environment itself behaves in a
probabilistic way defined by $\mu$ or the true environment is
deterministic, but we only have probabilistic information, of which
environment being the true environment. Combinations of
both cases are also possible. The interpretation does not matter in the
following. We just assume that we know $\mu$ but no more
about the environment whatever the interpretation may be.

Let us assume we are in cycle $k$ with history
$\hh y\!\hh x_1...\hh y\!\hh x_{k-1}$
and ask for the {\it best} output $y_k$.
Further, let
$\hh Q_k\!:=\!\{q:q(\hh y_{<k})=\hh x_{<k}\}$
be the set of all environments producing the above history.
The expected credit
for the next $m\!-\!k\!+\!1$ cycles (given the above history) is
given by a conditional probability:
\beq\label{eefunc}
  C^\mu_{km}(p|\hh y\!\hh x_{<k}) \;:=\;
  { \sum_{q\in \hh Q_k} \mu(q)C_{km}(p,q) \over
    \sum_{q\in \hh Q_k} \mu(q) }.
\eeq
We cannot simply determine $\maxarg_p(C_{1T})$ unlike the
deterministic case because the history is no longer
deterministically determined by $p$ and $q$, but depends on $p$
and $\mu$ {\it and} on the outcome of a stochastic process.
Every new cycle adds new information ($\hh x_i$) to the
system. This is indicated by the dots over the symbols.
In cycle $k$ we have to maximize the expected future
credit, taking into account the information in the history $\hh
y\!\hh x_{<k}$. This information is not already present
in $p$ and $q/\mu$ at the system's start unlike in the deterministic
case.

Further, we want to generalize the finite lifetime $T$ to a
dynamical (computable) farsightedness
$h_k\!\equiv\!m_k\!-\!k\!+\!1\!\geq\!1$, called horizon in the
following. For $m_k\!=\!T$ we have our original finite lifetime,
for $m_k\!=\!k\!+\!m\!-\!1$ the system maximizes in every cycle the next
$m$ expected credits. A discussion of the choices $m_k$ is delayed
to section \ref{secAIxi}.

The next $h_k$ credits are maximized by
$$
  p_k^\best \;:=\; \maxarg_{p\in \hh P_k} C^\mu_{km_k}(p|\hh y\!\hh
  x_{<k}),
$$
where $\hh P_k\!:=\!\{p:p(\hh x_{<k})=\hh y_{<k}*\}$ is the set of
systems consistent with the current history.
$p_k^\best$ depends on $k$ and is used only in step $k$ to
determine $\hh y_k$ by
$ p_k^\best(\hh x_{<k};\hh y_{<k})\!=\!\hh y_{<k}\hh y_k$.
After writing $\hh y_k$ the environment replies with $\hh x_k$
with (conditional) probability $\mu(\hh Q_{k+1})/\mu(\hh Q_k)$. This
probabilistic outcome provides new information to the system.
The cycle $k\!+\!1$ starts with determining $\hh y_{k+1}$ from
$p_{k+1}^\best$ (which differs from $p_k$ as $\hh x_k$ is
now fixed) and so on. Note that $p_k^\best$ depends also on
$\hh y_{<k}$ because $\hh P_k$ and $\hh Q_k$ do so.
But recursively inserting $p_{k-1}^\best$ and
so on, we can define
\beq\label{pbestfunc}
  p^\best(\hh x_{<k}) \;:=\;
  p_k^\best(\hh x_{<k};p_{k-1}^\best(\hh x_{<k-1}...p_1^\best)))
\eeq
It is a chronological function and computable if $X$, $Y$ and $m_k$ are
finite. The policy $p^\best$ defines our AI$\mu$ model.
For deterministic\footnote{We call a probability distribution deterministic
if it is 1 for exactly one argument and 0 for all others.}
$\mu$ this model reduces to the deterministic case.

It is important to maximize the sum of future credits and not, for instance,
to be greedy and only maximize the next credit, as is done e.g. in
sequence prediction. For example, let the environment be a
sequence of chess games and each cycle corresponds to one move.
Only at the end of each game a positive credit $c\!=\!1$ is given
to the system if it won the game (and made no illegal move).
For the system, maximizing all future credits means trying to win as
many games in as short as possible time (and avoiding illegal
moves). The same performance is reached, if we choose
$m_k\!=\!k\!+\!m$ with $m$ much larger than the typical game
lengths. Maximization of only the next credit would be a very bad
chess playing system. Even if we would make our credit $c$ finer,
e.g. by evaluating the number of chessmen, the system would play
very bad chess for $m\!=\!1$, indeed.

The AI$\mu$ model still depends on $\mu$ and $m_k$. $m_k$ is addressed
in section \ref{secAIxi}. To get our
final universal AI model the idea is to replace $\mu$ by the
universal probability $\xi$, defined later. This is motivated
by the fact that $\xi\!\to\!\mu$ in a certain sense for any $\mu$.
With $\xi$ instead of $\mu$ our model no longer depends on any
parameters, so it is truly universal. It remains to show that it
produces intelligent outputs. But let us continue step by step. In
the next section we develop an alternative but equivalent
formulation of the AI model given above. Whereas the functional
form is more suitable for theoretical considerations, especially
for the development of a timebounded version in section
\ref{secTime}, the iterative formulation of the next section will
be more appropriate for the explicit calculations in most of the
other sections.

\newpage
\section{The AI$\mu$ Model in Recursive and Iterative Form}\label{secAImurec}

\paragraph{Probability distributions:}
Throughout the paper we deal with sequences/strings and
conditional probability distributions on strings. Some
notations are therefore appropriate.

We use Greek letters for probability distributions and underline their
arguments to indicate that they are probability arguments. Let
$\rho_n(\pb x_1...\pb x_n)$ be the probability that a string starts with
$x_1...x_n$. We only consider sufficiently long strings, so the
$\rho_n$ are normalized to 1. Moreover, we drop the index on $\rho$
if it is clear from its arguments:
\beq\label{prop}
  \sum_{x_n\in X}\rho(\pb x_{1:n}) \equiv
  \sum_{x_n}\rho_n(\pb x_{1:n}) =
  \rho_{n-1}(\pb x_{<n}) \equiv
  \rho(\pb x_{<n})
  ,\quad
  \rho(\epsilon) \equiv \rho_0(\epsilon)=1.
\eeq
We also need conditional probabilities derived from Bayes' rule.
We prefer a notation which preserves the chronological order of the words, in
contrast to the standard notation $\rho(\cdot|\cdot)$ which flips it. We extend the
definition of $\rho$ to the conditional case with
the following convention for its arguments: An underlined argument
$\pb x_k$ is a probability variable and other non-underlined
arguments $x_k$ represent conditions. With this convention, Bayes'
rule has the form $\rho(x_{<n}\pb x_n)\!=\!\rho(\pb x_{1:n})/\rho(\pb
x_{<n})$.
The equation states that the probability that a string
$x_1...x_{n-1}$ is followed by $x_n$ is equal to the probability
of $x_1...x_n*$ divided by the probability of
$x_1...x_{n-1}*$. We use $x*$ as a shortcut for 'strings
starting with $x$'.

The introduced notation is also suitable for defining the
conditional probability $\rho(y_1\pb x_1...y_n\pb x_n)$ that the
environment reacts with $x_1...x_n$ under the condition that the
output of the system is $y_1...y_n$.
The environment is chronological, i.e. input $x_i$ depends on
$y\!x_{<i}y_i$ only. In the probabilistic case this means that
$\rho(y\!\pb x_{<k}y_k)\!:=\!\sum_{x_k}\rho(y\!\pb x_{1:k})$
is independent of $y_k$, hence a tailing $y_k$ in the arguments of $\rho$
can be dropped. Probability distributions with this
property will be called {\it chronological}.
The $y$ are always
conditions, i.e.\ never underlined, whereas additional
conditioning for the $x$ can be obtained with Bayes' rule
\bqa\label{bayes2}
  \rho(y\!x_{<n}y\!\pb x_n) =
  \rho(y\!\pb x_{1:n})/\rho(y\!\pb x_{<n}) \quad\mbox{and}
  \\[4mm]
  \rho(y\!\pb x_{1:n}) \;=\;
  \rho(y\!\pb x_1)\!\cdot\!\rho(y\!x_1y\!\pb x_2)\!\cdot...\cdot\!
  \rho(y\!x_{<n}y\!\pb x_n)
\eqa
The second equation is the first equation applied $n$ times.

\paragraph{Alternative Formulation of the AI$\mu$ Model:}
Let us define the AI$\mu$ model $p^\best$ in a different way. In the
next subsection we will show that the $p^\best$ model defined here
is identical to the functional definition of $p^\best$ given
in the last section.

Let $\mu(y\!\pb x_{1:k})$ be the true chronological prior probability
that the environment reacts with $x_{1:k}$ if provided with
actions $y_{1:k}$ from the system. We assume the cybernetic model depicted on page
\pageref{cyberpic} to be valid.
Next we define $C_{k+1,m}^\best(y\!x_{1:k})$ to be the $\mu$
expected credit sum in cycles $k\!+\!1$ to $m$ with outputs $y_i$
generated by system $p^\best$ and past responses $x_i$ from the
environment. Adding $c(x_k)$ we get the credit including cycle
$k$. The probability of $x_k$,
given $y\!x_{<k}y_k$, is given by the condition probability
$\mu(y\!x_{<k}y\!\pb x_k)$. So the expected credit sum
in cycles $k$ to $m$ given $y\!x_{<k}y_k$ is
\beq\label{ebesty}
  C_{km}^\best(y\!x_{<k}y_k) \;:=\;
  \sum_{x_k}[c(x_k)+C_{k+1,m}^\best(y\!x_{1:k})] \!\cdot\!
  \mu(y\!x_{<k}y\!\pb x_k)
\eeq
Now we ask about how $p^\best$ chooses
$y_k$. It should choose $y_k$ as to maximize the future credit.
So the expected number of errors in cycles $k$ to $m$ given
$y\!x_{<k}$ and $y_k$ chosen by $p^\best$ is
$ C_{km}^\best(y\!x_{<k})
\!:=\!\max_{y_k}C_{km}^\best(y\!x_{<k}y_k)$.
Together with the induction start
\beq\label{ee0}
  C_{m+1,m}^\best(y\!x_{1:m}) \;:=\; 0
\eeq
$C_{km}$ is completely defined.
We might summarize one cycle into the formula
\beq\label{airec2}
  C_{km}^\best(y\!x_{<k}) \;=\;
  \max_{y_k}\sum_{x_k}
  [c(x_k)+C_{k+1,m}^\best(y\!x_{1:k})] \!\cdot\!
  \mu(y\!x_{<k}y\!\pb x_k)
\eeq
If $m_k$ is our horizon function of $p^\best$ and
$\hh y\!\hh x_{<k}$ is the actual history in cycle
$k$, the output $\hh y_k$ of the system is explicitly given by
\beq\label{pbestrec}
  \hh y_k \;=\; \maxarg_{y_k}C_{km_k}^\best
  (\hh y\!\hh x_{<k}y_k) \;=:\;
  p^\best(\hh y\!\hh x_{<k})
\eeq
Then the environment responds $\hh x_k$ with
probability $\mu(\hh y\!\hh x_{<k}\hh y\!\pb{\hh
x}_k)$. Then cycle $k\!+\!1$ starts. We might
unfold the recursion (\ref{airec2}) further and give $\hh y_k$
non-recursive as
\beq\label{ydotrec}
  \hh y_k \;=\;
  \maxarg_{y_k}\sum_{x_k}\max_{y_{k+1}}\sum_{x_{k+1}}\;...\;
  \max_{y_{m_k}}\sum_{x_{m_k}}
  (c(x_k)\!+...+\!c(x_{m_k})) \!\cdot\!
  \mu(\hh y\!\hh x_{<k}y\!\pb x_{k:m_k})
\eeq
This has a direct interpretation: the probability of inputs
$x_{k:m_k}$ in cycle $k$ when the system outputs $y_{k:m_k}$ and
the actual history is $\hh y\!\hh x_{<k}$ is $\mu(\hh y\!\hh
x_{<k}y\!\pb x_{k:m_k})$. The future credit in this case is
$c(x_k)\!+...+\!c(x_{m_k})$. The best expected credit is obtained
by averaging over the $x_i$ ($sum_{x_i}$) and maximizing over the $y_i$.
This has to be done in chronological order to correctly
incorporate the dependency of $x_i$ and $y_i$ on the history.
This is essentially the expectimax algorithm/sequence
\cite{Mic66,Rus95}. The AI$\mu$ model is {\it optimal} in the
sense that no other policy leads to higher expected credit.

These explicit as well as recursive definitions of the AI$\mu$ model
are more index intensive as compared to the functional form but
are more suitable for explicit calculations.

\paragraph{Equivalence of Functional and Iterative AI model:}
The iterative environmental probability $\mu$ is given by the
functional form in the following way,
\beq\label{mufr}
  \mu(y\!\pb x_{1:k}) \;=\;
  \nq\sum_{q:q(y_{1:k})=x_{1:k}}\nq \mu(q)
\eeq
as is easy to see. We will prove the equivalence of
(\ref{pbestfunc}) and (\ref{pbestrec}) only for $k\!=\!2$ and
$m_2\!=\!3$. The proof of the general case is completely analog except
that the notation becomes quite messy.

Let us first evaluate (\ref{eefunc}) for fixed $\hh
y_1\hh x_1$ and some $p\!\in\!\hh P_2$, i.e. $p(\hh
x_1)=\hh y_1y_2$ for some $y_2$. If the next input to the
system is $x_2$, $p$ will respond with $p(\hh x_1
x_2)=\hh y_1y_2y_3$ for some $y_3$ depending on $x_2$. We
write $y_3(x_2)$ in the following\footnote{Dependency on dotted
words like $\hh x_1$ is not shown as the dotted words are fixed.}.
The numerator of (\ref{eefunc}) simplifies to
\beqn
  \sum_{q\in \hh Q_2} \mu(q)C_{23}(p,q) \;=\;
  \nq\sum_{q:q(\hh y_1)=\hh x_1}\nq \mu(q)C_{23}(p,q)
  \;=\; \sum_{x_2x_3}(c(x_2)\!+\!c(x_3))
  \nq\nq\sum_{q:q(\hh y_1y_2y_3(x_2))=\hh x_1x_2x_3}\nq\nq
  \mu(q) \;=\;
\eeqn
\beqn
  \;=\; \sum_{x_2x_3}(c(x_2)\!+\!c(x_3)) \!\cdot\!
  \mu(\hh y_1\pb{\hh x}_1y_2\pb x_2y_3(x_2)\pb x_3)
\eeqn
In the first equality we inserted the definition of $\hh Q_2$. In
the second equality we split the sum over $q$ by first summing
over $q$ with fixed $x_2x_3$. This allows us to pull
$C_{23}\!=c(x_2)\!+\!c(x_3)$ out of the inner sum. Then we sum
over $x_2x_3$. Further, we have inserted $p$, i.e. replaced $p$
by $y_2$ and $y_3(\cdot)$. In the last equality we used
(\ref{mufr}). The denominator reduces to
\beqn
  \sum_{q\in \hh Q_2} \mu(q) \;=\;
  \nq\sum_{q:q(\hh y_1)=\hh x_1}\nq \mu(q)
  \;=\; \mu(\hh y_1\pb{\hh x}_1).
\eeqn
For the quotient we get
$$
  C_{23}(p|\hh y_1\hh x_1) \;=\;
  \sum_{x_2x_3}(c(x_2)\!+\!c(x_3))\!\cdot\!
  \mu(\hh y_1\hh x_1
      y_2\pb x_2y_3(x_2)\pb x_3)
$$
We have seen that the relevant behaviour of $p\!\in\!\hh P_2$ in cycle 2 and 3
is completely determined by $y_2$ and the function $y_3(\cdot)$
$$
  \max_{p\in\hh P_2}C_{23}(p|\hh y_1\hh x_1) \;=\;
  \max_{y_2}\max_{y_3(\cdot)}\sum_{x_2x_3}(c(x_2)\!+\!c(x_3))\!\cdot\!
  \mu(\hh y_1\hh x_1y_2\pb x_2y_3(x_2)\pb c_3) \;=\;
$$
$$
  \;=\;
  \max_{y_2}\sum_{x_2}\max_{y_3}\sum_{x_3}(c(x_2)\!+\!c(x_3))\!\cdot\!
  \mu(\hh y_1\hh x_1y_2\pb x_2y_3\pb x_3)
$$
In the last equality we have used the fact that the functional
minimization over $y_3(\cdot)$ reduces to a simple minimization
over the word $y_3$ when interchanging with the sum over its
arguments
$(\max_{y_3(\cdot)}\sum_{x_2}\equiv\sum_{x_2}\max_{y_3})$.
In the functional case $\hh y_2$ is therefore determined by
$$
  \hh y_2 \;=\;
  \maxarg_{y_2}\sum_{x_2}\max_{y_3}\sum_{x_3}(c(x_2)\!+\!c(x_3))\!\cdot\!
  \mu(\hh y_1\hh x_1y_2\pb x_2y_3\pb x_3)
$$
This is identical to the iterative definition (\ref{ydotrec}) with
$k\!=\!2$ and $m_2\!=\!3$ $\qed$.

\paragraph{Factorizable $\mu$:}
Up to now we have made no restrictions on the form of the prior
probability $\mu$ apart from being a chronological probability
distribution. On the other hand, we will see that, in order to
prove rigorous credit bounds, the prior probability must satisfy
some separability condition to be defined later. Here we introduce
some very strong form of separability, when $\mu$ factorizes into
products. We start with a
factorization into two factors. Let us assume that $\mu$ is of the
form
\beq\label{fac12}
  \mu(y\!\pb x_{1:n}) \;=\;
  \mu_1(y\!\pb x_{<l}) \cdot
  \mu_2(y\!\pb x_{l:n})
\eeq
for some fixed $l$ and sufficiently large $n\!\geq\!m_k$.
For this $\mu$ the output $\hh y_k$ in cycle
$k$ of the AI$\mu$ system (\ref{ydotrec}) for $k\!\geq\!l$ depends on
$\hh y\!\hh x_{l:k-1}$ and $\mu_2$ only and
is independent of $\hh y\!\hh x_{<l}$
and $\mu_1$. This is easily seen when inserting
\beq\label{fac11}
  \mu(\hh y\!\hh x_{<k}y\!\pb x_{k:m_k}) =
  \underbrace{\mu_1(\hh y\!\hh x_{<l})}_{\equiv 1}
  \cdot
  \mu_2(\hh y\!\hh x_{l:k-1}y\!\pb x_{k:m_k})
\eeq
into (\ref{ydotrec}). For $k\!<\!l$ the output $\hh y_k$ depends
on $\hh y\!\hh x_{<k}$ (this is trivial) and $\mu_1$
only (trivial if $m_k\!<\!l$) and is independent of $\mu_2$.
The non-trivial case, where the horizon $m_k\!\geq\!l$ reaches
into the region $\mu_2$, can be proved as follows (we abbreviate
$m\!:=\!m_k$ in the following). Inserting (\ref{fac12}) into the
definition of $C_{lm}^\best(y\!x_{<l})$ the factor
$\mu_1$ is $1$ as in (\ref{fac11}). We abbreviate
$C_{lm}^\best\!:=\!C_{lm}^\best(y\!x_{<l})$ as
it is independent of its arguments. One can
decompose
\beq\label{decompE}
  C_{km}^\best(y\!x_{<k}) \;=\;
  C_{k,l-1}^\best(y\!x_{<k}) \;+\; C_{lm}^\best
\eeq
For $k\!=\!l$ this is true because the first term on the r.h.s.\ is
zero.
For $k\!<\!l$ we prove the decomposition by induction from $k\!+\!1$ to $k$.
\beqn
  C_{km}^\best(y\!x_{<k}) \;=\;
  \max_{y_k}\sum_{x_k}
  [c(x_k)+C_{k+1,l-1}^\best(y\!x_{1:k})+C_{lm}^\best] \!\cdot\!
  \mu_1(y\!x_{<k}y\!\pb x_k) \;=\;
\eeqn
\beqn
  \;=\; \max_{y_k}\bigg[\sum_{x_k}
  (c(x_k)+C_{k+1,l-1}^\best(y\!x_{<k})) \!\cdot\!
  \mu_1(y\!x_{<k}y\!\pb x_k) + C_{lm}^\best\bigg]
   \;=\;
\eeqn
\beqn
  \;=\; C_{k,l-1}^\best(y\!x_{<k}) + C_{lm}^\best
\eeqn
Inserting (\ref{decompE}), valid for $k$ by induction hypothesis,
into (\ref{airec2}) gives the first equality. In the second
equality we have performed the $x_k$ sum for the
$C_{lm}^\best\!\cdot\!\mu_1$ term which is now independent of $y_k$. It can
therefore be pulled out of $\max_{y_k}$. In the last
equality we used again the definition (\ref{airec2}). This completes
the induction step and proves
(\ref{decompE}) for $k\!<\!l$. $\hh y_k$ can now be represented
as
\beq
  \hh y_k \;=\; \maxarg_{y_k}C_{km}^\best
  (\hh y\!\hh x_{<k}y_k) \;=\;
  \maxarg_{y_k}C_{k,l-1}^\best(\hh y\!\hh x_{<k}y_k)
\eeq
where (\ref{pbestrec}) and (\ref{decompE}) and the fact that
an additive constant $C_{lm}^\best$ does not change
$\maxarg_{y_k}$ has been used. $C_{k,l-1}^\best(\hh y\!\hh x_{<k}y_k)$ and
hence $\hh y_k$ is independent of $\mu_2$ for $k\!<\!l$. Note,
that $\hh y_k$ is also independent of the choice of $m$, as
long as $m\!\geq\!l$.

In the general case the cycles are grouped into
independent episodes $r\!=\!1,2,3,...$, where each episode $r$
consists of the cycles $k\!=\!n_r\!+\!1,...,n_{r+1}$ for some
$0=n_0<n_1<...<n_s=n$:
\beq\label{facmu}
  \mu(y\!\pb x_{1:n}) \;=\;
  \prod_{r=0}^{s-1} \mu_r(y\!\pb x_{n_r+1:n_{r+1}})
\eeq
In the simplest case, when all episodes have the
same length $l$ then $n_r=r\!\cdot\!l$. $\hh y_k$ depends on
$\mu_r$ and $x$ and $y$ of episode $r$ only, with $r$ such
that $n_r\!<\!k\!\leq\!n_{r+1}$.
\beq\label{facydot}
  \hh y_k =
  \maxarg_{y_k}\sum_{x_k}...
  \max_{y_t}\sum_{x_t}
  (c(x_k)\!+...+\!c(x_t)) \!\cdot\!
  \mu_r(\hh y\!\hh x_{n_r+1:k-1}y\!\pb x_{k:n_{r+1}}) \\[-3mm]
\eeq
with $t\!:=\!\min\{m_k,n_{r+1}\}$. The different episodes are
completely independent in the following sense. The inputs $x_k$
of different episodes are statistically independent and
depend only on $y_k$ of the same episode. The outputs $y_k$ depend on the
$x$ and $y$ of the corresponding episode $r$ only, and are
independent of the actual I/O of the other episodes.

If all episodes have a length of at most $l$, i.e.
$n_{r+1}\!-\!n_r\!\leq\!l$ and if we choose the horizon
$h_k$ to be at least $l$, then
$m_k\!\geq\!k\!+\!l\!-\!1\!\geq\!n_r\!+\!l\!\geq\!n_{r+1}$ and
hence $t=n_{r+1}$ independent of $m_k$. This means that for
factorizable $\mu$ there is no problem in taking the limit
$m_k\!\to\!\infty$. Maybe this limit can also be performed in the
more general case of a separable $\mu$. The (problem of the)
choice of $m_k$ will be discussed in more detail later.

Although factorizable $\mu$ are too restrictive to cover all AI
problems, it often occurs in practice in the form of repeated
problem solving, and hence, is worth being studied. For example, if
the system has to play games like chess repeatedly, or has to
minimize different functions, the different games/functions might
be completely independent, i.e. the environmental probability
factorizes, where each factor corresponds to a game/function
minimization. For details, see the appropriate sections on
strategic games and function minimization.

Further, for factorizable $\mu$ it is probably easier to derive
suitable credit bounds for the universal AI$\xi$ model defined in
the next section, than for the general separable case which will be
introduced later. This could be a first step toward a definition
and proof for the general case of separable problems. One goal of
this paragraph was to show, that the notion of a factorizable
$\mu$ could be the first step toward a definition and analysis of
the general case of separable $\mu$.

\paragraph{Constants and Limits:}
We have in mind a universal system with complex
interactions that is as least as intelligent and complex as a human
being. One might think of a system whose input $y_k$ comes from a
digital video camera, the output $x_k$ is some image to a
monitor\footnote{Humans can only simulate a screen as
output device by drawing pictures.}, only for the valuation we
might restrict to the most primitive binary one, i.e. $c_k\!\in I\!\!B$. So we think of the
following constant sizes:
$$
\begin{array}{ccccccccc}
  1 & \ll & \langle l(y_kx_k)\rangle & \ll & k & \leq & T & \ll & |Y\times X| \\
  1 & \ll & 2^{16} & \ll & 2^{24} & \le & 2^{32} & \ll & 2^{65536}
\end{array}
$$
The first two limits say that the actual number $k$ of
inputs/outputs should be reasonably large, compared to the typical
size $\langle l\rangle$ of the input/output words, which itself
should be rather sizeable. The last limit expresses the fact that
the total lifetime $T$ (number of I/O cycles) of the system is far
too small to allow every possible input to occur, or to try every
possible output, or to make use of identically repeated
inputs or outputs. We do not expect any useful outputs for
$k\le\langle l\rangle$. More interesting than the lengths of the
inputs is the complexity $K(x_1...x_k)$ of all inputs until now,
to be defined later. The environment is usually not "perfect". The
system could either interact with a non-perfect human or tackle a
non-deterministic world (due to quantum mechanics or chaos)
world\footnote{Whether there exist stochastic processes at all is
a difficult question. At least the quantum indeterminacy comes
very close to it.}. In either case, the sequence contains some
noise, leading to $K\sim \langle l\rangle\!\cdot\!k$. The
complexity of the probability distribution of the input sequence
is something different. We assume that this noisy world operates
according to some simple computable, though not finite rules.
$K(\mu_k)\ll \langle l\rangle\!\cdot\!k$, i.e. the rules of the
world can be highly compressed. On the other hand, there may
appear new aspects of the environment for $k\!\to\!\infty$ causing
a non-bounded $K(\mu_k)$.

In the following we never use these limits, except when explicitly
stated. In some simpler models and examples the size of the
constants will even violate these limits (e.g. $l(x_k)=l(y_k)=1$),
but it is the limits above that the reader should bear in mind. We are
only interested in theorems which do not degenerate under the
above limits.

\paragraph{Sequential decision theory:}
In the following we clarify the connection of (\ref{airec2}) and
(\ref{pbestrec}) to sequential decision theory and discuss similarities and
differences. With probability $M^a_{ij}$, the system under
consideration should reach (environmental) state $i\!\in\!S$ when
taking action $a\!\in\!A$ depending on the current state
$j\!\in\!S$. If the system receives reward $R(i)$,
the optimal policy $p^*$, maximizing expected utility (defined as
sum of future rewards), and the utility $U(i)$ of policy
$p^*$ are
\beq\label{dt}
  p^*(i)=\maxarg_a\sum_j M^a_{ij}U(j) \quad,\quad
  U(i)=R(i)+\max_a\sum_j M^a_{ij}U(j)
\eeq
See \cite{Rus95} for details and further references. Let us identify
\bqan
  S=(Y\!\times\!X)^*,\quad A=Y,\quad
  a=y_k, \quad M^a_{ij}=\mu(y\!x_{<k}y\!\pb x_k), \\[4pt]
  i=y\!x_{<k}, \quad R(i)=c(x_{k-1}), \quad
  U(i)=C^*_{k-1,m}(y\!x_{<k})=c(x_{k-1})+C^*_{km}(y\!x_{<k}), \\[4pt]
  j=y\!x_{1:k}, \quad R(j)=c(x_k), \quad
  U(j)=C^*_{km}(y\!x_{1:k})=c(x_k)+C^*_{k+1,m}(y\!x_{1:k}),
\eqan
where we further set $M^a_{ij}\!=\!0$ if $i$ is not a starting
substring of $j$ or if $a\!\neq\!y_k$. This ensures the sum over
$j$ in (\ref{dt}) to reduce to a sum over $x_k$. If we set
$m_k\!=\!m$ and use
$C^*_{km}(y\!x_{<k}y_k)\!=\!\sum_{x_k}C^*_{km}(y\!x_{1:k})$ in
(\ref{pbestrec}), it is easy to see that (\ref{dt}) coincides with
(\ref{airec2}) and (\ref{pbestrec}).

Note that despite of this formal equivalence, we were forced to use
the complete history $y\!x_{<k}$ as environmental state $i$. The
AI$\mu$ model neither assumes stationarity, nor Markov property,
nor complete accessibility of the environment, as any assumption
would restrict the applicability of AI$\mu$. The consequence is
that every state occurs at most once in the lifetime of the
system. Every moment in the universe is unique! Even if the state
space could be identified with the input space $X$, inputs would
usually not occur twice by assumption $k\!\ll\!|X|$, made in the
last subsection. Further, there is no (obvious) universal
similarity relation on $(X\!\times\!Y)^*$ allowing an effective
reduction of the size of the state space. Although many algorithms
(e.g. value and policy iteration) have problems in solving
(\ref{dt}) for huge or infinite state spaces in practice,
there is no principle problem in determining $p^*$ and $U$, as
long as $\mu$ is known and $|X|$, $|Y|$ and $m$ are finite.

Things dramatically change if $\mu$ is unknown. Reinforcement
learning algorithms \cite{Kae96} are commonly used in this case to
learn the unknown $\mu$. They succeed if the state space is either
small or has effectively been made small by so called generalization
techniques. In any case, the solutions are either ad hoc, or work
in restricted domains only, or have serious problems with state
space exploration versus exploitation, or have non-optimal
learning rate. There is no universal and optimal solution to this
problem so far. In the next section we present a new model and
argue that it formally solves all these problems in an optimal
way. It will not concern with learning of $\mu$ directly. All we
do is to replace the true prior probability $\mu$ by a universal
probability $\xi$, which is shown to converge to $\mu$ in a sense.

\newpage
\section{The Universal AI$\xi$ Model}\label{secAIxi}

\paragraph{Induction and Algorithmic Information theory:}
One very important and highly non-trivial aspect of intelligence is
inductive inference. Before formulating the AI$\xi$ model,
a short introduction to the history of induction is given, culminating
into the sequence prediction theory by Solomonoff. We emphasize
only those aspects which will be of importance for the development
of our universal AI$\xi$ model.

Simply speaking, induction is the process of
predicting the future from the past or, more precisely, it is the
process of finding rules in (past) data and using these rules to
guess future data. On the one hand, induction seems to happen in
every day life by finding regularities in past observations and
using them to predict the future. On the other hand, this procedure
seems to add knowledge about the future from past observations.
But how can we know something about the future? This dilemma and
the induction principle in general have a long philosophical
history
\begin{itemize}\parskip=0ex\parsep=0ex\itemsep=0ex
  \item Hume's negation of Induction (1711-1776) \cite{Hume},
  \item Epicurus' principle of multiple explanations (342?-270?
  BC),
  \item Occams' razor (simplicity) princple (1290?-1349?),
  \item Bayes' rule for conditional probabilites \cite{Bay63}
\end{itemize}
and a short but important mathematical history: a clever
unification of all these aspects into one formal theory of
inductive inference has been done by Solomonoff \cite{Sol64} based
on Kolmogorov's \cite{Kol65} definition of complexity. For an
excellent introduction into Kolmogorov complexity and Solomonoff
induction one should consult the book of Li and Vit\'anyi
\cite{LiVi93}. In the rest of this subsection we state all results
which are needed or generalized later.

Let us choose some universal prefix Turing machine $U$ with
unidirectional binary input and output tapes and a bidirectional
working tape. We can then define the (prefix) Kolomogorov complexity
\cite{Cha75,Gac74,Kol65,Lev74} as the shortest prefix program $p$, for which $U$
outputs $x\!=\!x_{1:n}$ with $x_i\in\!I\!\!B$:
$$
  K(x) \;:=\; \min_p\{l(p): U(p)=x\}
$$
The universal semimeasure $\xi(\pb x)$ is defined as the probability
that the output of the universal Turing machine $U$ starts with
$x$ when provided with fair coin flips on the input tape \cite{Sol64,Sol78}. It is
easy to see that this is equivalent to the formal definition
\beq\label{xidef}
  \xi(\pb x)\;:=\;\sum_{p\;:\;U(p)=x*}\nq 2^{-l(p)}
\eeq
where the sum is over minimal programs $p$ for which $U$
outputs a string starting with $x$. $U$ might be non-terminating.
As the shortest programs dominate the sum, $\xi$ is closely
related to $K(x)$ ($\xi(\pb x)=2^{-K(x)+O(K(l(x))}$).
$\xi$ has the important universality property \cite{Sol64}, that it
majorizes every computable probability distribution $\rho$ up
to a multiplicative factor
depending only on $\rho$ but not on $x$:
\beq\label{uni}
  \xi(\pb x) \;\stackrel{\times}{\geq}\; 2^{-K(\rho)}\!\cdot\!\rho(\pb x).
\eeq
A '$\times$' above an (in)equality denotes (in)equality within a
universal multiplicative constant,
a '$+$' above an (in)equality denotes (in)equality within a
universal additive constant, both depending only on the choice of the
universal reference machine $U$.
$\xi$ itself is {\it not} a probability
distribution\footnote{It is possible to normalize $\xi$ to a
probability distribution as has been done in
\cite{Wil70,Sol78,Hut99} by giving up the enumerability of $\xi$.
Error bounds (\ref{eukdist}) and (\ref{spebound}) hold for both
definitions.}.
We have $\xi(\pb{x0})\!+\!\xi(\pb{x1})\!<\!\xi(\pb
x)$ because there are programs $p$, which output just $x$, neither
followed by $0$ nor $1$. They just stop after printing $x$ or
continue forever without any further output. We will call a
function $\rho\!\geq 0$ with the properties
$\rho(\epsilon)\!\leq\!1$ and $\sum_{x_n}\rho(\pb
x_{1:n})\!\leq\!\rho(\pb x_{<n})$ a {\it semimeasure}. $\xi$ is a
semimeasure and (\ref{uni}) actually holds for all enumerable
semimeasures $\rho$.

(Binary) sequence prediction algorithms try to predict the
continuation $x_n$ of a given sequence $x_1...x_{n-1}$. In the
following we will assume that the sequences are drawn according to
a probability distribution and that the true prior probability of
$x_{1:n}$ is $\mu(\pb{x_1...x_n})$. The probability of $x_n$ given
$x_{<n}$ hence is $\mu(x_{<n}\pb x_n)$. The best possible system
predicts the $x_n$ with higher probability. Usually $\mu$ is
unknown and the system can only have some belief $\rho$ about the
true prior probability $\mu$. Let SP$\rho$ be a probabilistic
sequence predictor, predicting $x_n$ with probability
$\rho(x_{<n}\pb x_n)$. Further we define a deterministic sequence
predictor SP$\Theta_\rho$ predicting the $x_n$ with higher $\rho$
probability. $\Theta_\rho(x_{<n}\pb x_n)\!:=\!1$ if
$\rho(x_{<n}\pb x_n)\!>\!{1\over 2}$ and $\Theta_\rho(x_{<n}\pb x_n)\!:=\!0$
otherwise.  If $\rho$ is only a semimeasure the SP$\rho$ and
SP$\Theta_\rho$ systems might refuse any output in some cycles
$n$. The SP$\Theta_\mu$ is the best prediction scheme when $\mu$
is known.

If $\rho(x_{<n}\pb x_n)$ converges quickly to $\mu(x_{<n}\pb x_n)$ the
number of additional prediction errors introduced by using
$\Theta_\rho$ instead of $\Theta_\mu$ for prediction should be
small in some sense. Now the universal probability $\xi$
comes into play as it has been proved
by Solomonoff \cite{Sol78} that the $\mu$ expected Euclidean
distance betweewn $\xi$ and $\mu$ is finite
\beq\label{eukdist}
  \sum_{k=1}^\infty\sum_{x_{1:k}}\mu(\pb x_{1:k})
  (\xi(x_{<k}\pb x_k)-\mu(x_{<k}\pb x_k))^2 \;\stackrel{+}{<}\;
  {\1d2}\ln 2\!\cdot\!K(\mu)
\eeq
The '$+$' atop '$<$' means up to additive terms of order 1.
So indeed the difference does tend to zero, i.e.
$\xi(x_{<n}\pb x_n)\toinfty{n}\mu(x_{<n}\pb x_n)$ with $\mu$ probability
$1$ for {\it any} computable probability distribution $\mu$. The reason for the
astonishing property of a single (universal) function to
converge to {\it any} computable probability distribution lies in the fact that the
set of $\mu$ random sequences differ for different $\mu$.
The universality property (\ref{uni}) is the central ingredient for
proving (\ref{eukdist}).

Let us define the total number of expected erroneous predictions
the SP$\rho$ system makes for the first $n$ bits
\beq\label{esp}
  E_{n\rho} \;:=\; \sum_{k=1}^n\sum_{x_{1:k}}\mu(\pb x_{1:k})
  (1\!-\!\rho(x_{<k}\pb x_k))
\eeq
The SP$\Theta_\mu$ system is best in the sense that
$E_{n\Theta_\mu}\!\leq\!E_{n\rho}$
for any $\rho$. In \cite{Hut99} it has been shown that
SP$\Theta_\xi$ is not much worse
\beq\label{spebound}
  E_{n\Theta_\xi}\!-\!E_{n\rho} \;\leq\;
  H+\sqrt{4E_{n\rho}H+H^2} \;=\;
  O(\sqrt{E_{n\rho}})\quad,\quad
  H\;\stackrel{+}{<}\;\ln 2\!\cdot\!K(\mu)
\eeq
with the tightest bound for $\rho\!=\!\Theta_\mu$. For finite
$E_{\infty\Theta_\mu}$, $E_{\infty\Theta_\xi}$ is finite too. For
infinite $E_{\infty\Theta_\mu}$,
$E_{n\Theta_\xi}/E_{n\Theta_\mu}\toinfty{n}1$ with rapid
convergence. One can hardly imagine any better prediction
algorithm without extra knowledge about the environment. In
\cite{Hut00e}, (\ref{eukdist}) and (\ref{spebound}) have been
generalized from binary to arbitrary alphabet. Apart from
computational aspects, which are of course very important, the
problem of sequence prediction could be viewed as essentially
solved.

\paragraph{Definition of the AI$\xi$ Model:}
We have developed enough formalism to suggest our universal
AI$\xi$ model\footnote{Speak 'aixi' and write AIXI without Greek letters.}.
All we have to do is to suitably generalize the universal
semimeasure $\xi$ from the last subsection and replace the true
but unknown prior probability $\mu^{AI}$ in the AI$\mu$ model by this
generalized $\xi^{AI}$. In what sense this AI$\xi$ model is universal
will be discussed later.

In the functional formulation we define the universal probability
$\xi^{AI}$ of an environment $q$ just as $2^{-l(q)}$
\beqn
  \xi(q) \;:=\; 2^{-l(q)}
\eeqn
The definition could not be easier\footnote{It is not necessary
to use $2^{-K(q)}$ or something similar as some reader may expect
at this point. The reason is that for every program $q$ there
exists a functionally equivalent program $q'$ with
$K(q')=l(q')$.}!\footnote{Here and later we identify objects with
their coding relative to some fixed Turing machine $U$. For example, if $q$ is
a function $K(q):=K(\lceil q\rceil)$ with $\lceil q\rceil$ being a
binary coding of $q$ such that $U(\lceil q\rceil,y):=q(y)$. On the
other hand, if $q$ already is a binary string we define $q(y)\!:=U(q,y)$.}
Collecting the formulas of section \ref{secAIfunc}
and replacing $\mu(q)$ by $\xi(q)$
we get the definition of the AI$\xi$ system in
functional form. Given the history $\hh y\!\hh x_{<k}$ the
functional AI$\xi$ system outputs
\beq\label{eefuncxi}
  \hh y_k \;:=\;
  \maxarg_{y_k}\max_{p:p(\hh x_{<k})=\hh y_{<k}y_k}
  \sum_{q:q(\hh y_{<k})=\hh x_{<k}}
  \nq 2^{-l(q)}\cdot C_{km_k}(p,q)
\eeq
in cycle $k$, where $C_{km_k}(p,q)$ is the total credit of cycles $k$ to $m_k$ when
system $p$ interacts with environment $q$. We have dropped the
denominator $\sum_q\mu(q)$ from (\ref{eefunc}) as it is
independent of the $p\!\in\!\hh P_k$ and a constant multiplicative
factor does not change $\maxarg$.

For the iterative formulation the universal probability
$\xi$ can be obtained by inserting the functional $\xi(q)$ into
(\ref{mufr})
\beq\label{uniMAI}
  \xi(y\!\pb x_{1:k}) \;=\;
  \nq\sum_{q:q(y_{1:k})=x_{1:k}}\nq 2^{-l(q)}
\eeq
Replacing $\mu$ by $\xi$ in (\ref{ydotrec}) the
iterative AI$\xi$ system outputs
\beq\label{ydotxi}
  \hh y_k \;=\;
  \maxarg_{y_k}\sum_{x_k}\max_{y_{k+1}}\sum_{x_{k+1}}\;...\;
  \max_{y_{m_k}}\sum_{x_{m_k}}
  (c(x_k)\!+...+\!c(x_{m_k})) \!\cdot\!
  \xi(\hh y\!\hh x_{<k}y\!\pb x_{k:m_k})
\eeq
in cycle $k$ given the history $\hh y\!\hh x_{<k}$.

One subtlety has been passed over. Like in the
SP case, $\xi$ is not a probability distribution but satisfies only the weaker inequalities
\beq\label{chrf}
  \sum_{x_n}\xi(y\!\pb x_{1:n}) \;\leq\; \xi(y\!\pb x_{<n})
  \quad,\quad
  \xi(\epsilon) \;\leq\; 1
\eeq
Note, that the sum on the l.h.s.\ is {\it not}
independent of $y_n$ unlike for chronological probability
distributions. Nevertheless, it is bounded by something (the r.h.s)
which is independent of $y_n$. The reason is that the sum in
(\ref{uniMAI}) runs over (partial recursive) chronological
functions only and the functions $q$ which satisfy
$q(y_{1:n})=x_{<n}*$ are a subset of the functions satisfying
$q(y_{<n})=x_{<n}$. Therefore we will in general call functions satisfying
(\ref{chrf}) {\it chronological semimeasures}. The important point
is that the conditional probabilities (\ref{bayes2}) are $\leq\!1$
like for true probability distributions.

The equivalence of the functional and iterative AI model proven in
section \ref{secAImurec} is true for every chronological
semimeasure $\rho$, esp.\ for $\xi$, hence we can talk about {\it
the} AI$\xi$ model in this respect. It (slightly) depends on the
choice of universal Turing machine. $l(q)$ is defined only up to
an additive constant. It also depends on the choice of
$X\!=\!C\!\times\!X'$ and $Y$, but we do not expect any bias when
the spaces are chosen sufficiently simple, e.g. all strings of
length $2^{16}$. Choosing $I\!\!N$ as word space would be optimal,
but whether the maxima (suprema) exist in this case, has to be
shown beforehand. The only non-trivial dependence is on the
horizon function $m_k$ which will be discussed later. So apart
from $m_k$ and unimportant details the AI$\xi$ system is uniquely
defined by (\ref{eefuncxi}) or (\ref{ydotxi}).
It doesn't depend
on assumptions about the environment apart from being generated
from some computable (but unknown!) probability distribution.

\paragraph{Universality of $\xi^{AI}$:}
In which sense the AI$\xi$ model is optimal will be clarified
later. In this and the next two subsections we show that $\xi^{AI}$
defined in (\ref{uniMAI}) is universal and converges to $\mu^{AI}$ analog to the
SP case (\ref{uni}) and (\ref{eukdist}). The proofs are
generalizations from the SP case. The $y$ are pure spectators and
cause no difficulties in the generalization. The replacement of
the binary alphabet $I\!\!B$ used in SP by the (possibly infinite)
alphabet $X$ is possible, but needs to be done with care. In
(\ref{uni}) $U(p)=x*$ produces strings starting with $x$, whereas
in (\ref{uniMAI}) we can demand $q$ to output exactly $n$ words $x_{1:n}$ as
$q$ knows $n$ from the number of input words $y_1...y_n$.
For proofs of (\ref{uni}) and (\ref{eukdist}) see \cite{Sol78} and
\cite{LiVi92}.

There is an alternative
definition of $\xi$ which coincides with (\ref{uniMAI}) within a
multiplicative constant of $O(1)$,
\beq\label{xirhodef}
  \xi(y\!\pb x_{1:n}) \;\stackrel{\times}{=}\; \sum_\rho 2^{-K(\rho)}\rho(y\!\pb
  x_{1:n})
\eeq
where the sum runs over all enumerable chronological semimeasures.
The $2^{-K(\rho)}$ weighted sum over probabilistic environments
$\rho$, coincides with the sum over $2^{-l(q)}$ weighted
deterministic environments $q$, as will be proved below.
In the next subsection we show that an enumeration of all
enumerable functions can be converted into an enumeration of
enumerable chronological semimeasures $\rho$. $K(\rho)$ is co-enumerable,
therefore $\xi$ defined in (\ref{xirhodef}) is itself enumerable.
The representation (\ref{uniMAI}) is also enumerable. As
$\sum_\rho2^{-K(\rho)}\!\leq\!1$ and the $\rho's$ satisfy (\ref{chrf}), $\xi$
is a chronological semimeasure as well. If we pick one $\rho$ in
(\ref{xirhodef}) we get the universality property ''for free''
\beq\label{uniaixi}
  \xi(y\!\pb x_{1:n}) \;\stackrel{\times}{\geq}\; 2^{-K(\rho)}\rho(y\!\pb x_{1:n})
\eeq
$\xi$ is a universal element in the sense of (\ref{uniaixi}) in
the set of all enumerable chronological semimeasures.

To prove universality of $\xi$ in the form (\ref{uniMAI}) we have
to show that for every  enumerable chronological semimeasure
$\rho$ there exists a Turing machine $T$ with
\beq\label{reprho}
  \rho(y\!\pb x_{1:n}) \;=\; \sum_{q:T(qy_{1:n})=x_{1:n}}\nq 2^{-l(q)}
  \quad\mbox{and}\quad l(T)\stackrel{+}{=}K(\rho).
\eeq

This will not be done here. Given $T$ the universality of
$\xi$
follows from
\beqn
  \xi(y\!\pb x_{1:n}) \;=\;
  \nq\nq\sum_{\quad\quad q:U(qy_{1:n})=x_{1:n}}\nq\nq 2^{-l(q)}
  \;\geq\;
  \nq\nq\sum_{\quad\quad q:U(Tq'y_{1:n})=x_{1:n}}\nq\;\nq\nq 2^{-l(Tq')}
  \;=\;
  2^{-l(T)}\nq\nq\sum_{q:T(q'y_{1:n})=x_{1:n}}\nq\nq 2^{-l(q')}
  \stackrel{\times}{\;=\;}
  2^{-K(\rho)}\rho(y\!\pb x_{1:n})
\eeqn
The first equality and (\ref{uniMAI}) are identical by definition.
In the inequality we have restricted the sum over all $q$ to $q$
of the form $q\!=\!Tq'$. The third relation is true as running $U$
on $Tz$ is a simulation of $T$ on $z$. The last equality follows
from (\ref{reprho}). All enumerable, universal, chronological
semimeasures coincide up to a multiplicative constant, as they
mutually dominate each other. Hence, definitions (\ref{uniMAI}) and
(\ref{xirhodef}) are, indeed, equivalent.

\paragraph{Converting general functions into chronological semi-measures:}
To complete the proof of the universality (\ref{uniaixi}) of $\xi$
we need to convert enumerable functions
$\psi:I\!\!B^*\!\to\!I\!\!R^+$ into enumerable chronological
semi-measures $\rho:(Y\!\times\!X)^*\!\to\!I\!\!R^+$ with certain
additional properties. Every enumerable function like $\psi$ and
$\rho$ can be approximated from below by definition\footnote{Defining
enumerability as the supremum of total primitive recursive
functions is more suitable for our purpose than the equivalent
definition as a limit of monotone increasing partial
recursive functions. In terms of Turing machines, the recursion
parameter is the time after which a computation is terminated.} by
primitive recursive functions
$\varphi:I\!\!B^*\!\times\!I\!\!N\!\to\!I\!\!\!Q^+$ and
$\phi:(Y\!\times\!X)^*\!\times\!I\!\!N\!\to\!I\!\!\!Q^+$ with
$\psi(s)=\sup_t\varphi(s,t)$ and $\rho(s)=\sup_t\phi(s,t)$ and
recursion parameter $t$. For arguments of the form
$s\!=\!y\!x_{1:n}$ we recursively (in $n$) construct $\phi$ from
$\varphi$ as follows:
\begin{eqnarray}\label{ccsm1}
  \varphi'(y\!x_{1:n},t) &\!:=\!&
  \left\{
  \begin{array}{c@{\quad\mbox{for}\quad}l}
    \varphi(y\!x_{1:n},t) & x_n<t     \\
    0                   & x_n\geq t
  \end{array} \right.
  \quad,\quad \varphi'(\epsilon,t) \;:=\; \varphi(\epsilon,t)
\\ \label{ccsm2}
  \phi(\epsilon,t) &\!:=\!& \max_{0\leq i\leq t}
  \Big\{\varphi'(\epsilon,i):\varphi'(\epsilon,i)\leq 1 \Big\}
\\ \label{ccsm3}
  \phi(y\!\pb x_{1:n},t) &\!:=\!& \max_{0\leq i\leq t}
  \Big\{ \varphi'(y\!x_{1:n},i):{\textstyle\sum_{x_n}}\varphi'(y\!x_{1:n},i)\leq
     \phi(y\!\pb x_{<n},t) \Big\}
\end{eqnarray}
With $x_n\!<\!t$ we mean that the natural number associated with
string $x_n$ is smaller than $t$.
According to (\ref{ccsm1}) with $\varphi$ also $\varphi'$ as well as
$\sum_{x_n}\varphi'$ are primitive recursive functions. Further, if we
allow $t\!=\!0$ we have $\varphi'(s,0)=0$. This ensures that
$\phi$ is a total function.

In the following we prove by induction over $n$ that $\phi$ is a
primitive recursive chronological semimeasure
monotone increasing in $t$. All necessary properties hold for
$n\!=\!0$ ($y\!x_{1:0}\!=\!\epsilon$) according to (\ref{ccsm2}).
For general $n$ assume that the induction hypothesis is true for
$\phi(y\!\pb x_{<n},t)$. We can see from (\ref{ccsm3}) that
$\phi(y\!\pb x_{1:n},t)$ is monotone  increasing in $t$. $\phi$ is
total as $\varphi'(y\!x_{1:n},i\!=\!0)\!=\!0$ satisfies the
inequality. By assumption $\phi(y\!x_{<n},t)$ is
primitive recursive, hence with $\sum_{x_n}\varphi'$ also the order relation
$\sum\varphi'\!\leq\!\phi$ is primitive recursive. This ensures
that the non-empty finite set
$\{\varphi'\!:\!\sum\varphi'\!\leq\!\phi\}_i$ and its maximum
$\phi(y\!\pb x_{1:n},t)$ are primitive recursive. Further,
$\phi(y\!\pb x_{1:n},t)\!=\!\varphi'(y\!x_{1:n},i)$ for some $i$ with
$i\!\leq\!t$ independent of $x_n$. Thus,
$\sum_{x_n}\phi(y\!\pb x_{1:n},t)$ $=$ $\sum_{x_n}\varphi'(y\!x_{1:n},i)$
$\leq$ $\phi(y\!\pb x_{<n},t)$ which is the condition for $\phi$ being a
chronological semimeasure. Inductively we have proved that $\phi$ is
indeed a primitive recursive chronological semimeasure
monotone increasing in $t$.

In the following we show that every (total)\footnote{Semimeasures
are, by definition, total functions.} enumerable chronological
semimeasure $\rho$ can be enumerated by some $\phi$. By definition
of enumerability there exist primitive recursive functions
$\tilde\varphi$ with $\rho(s)\!=\!\sup_t\tilde\varphi(s,t)$. The
function $\varphi(s,t)\!:=\!(1\!-\! ^1\!/_t)\!\cdot\!
\max_{i<t}\tilde\varphi(s,i)$ also enumerates $\rho$ but has
the additional advantage of being strictly monotone increasing in $t$.

$\varphi'(y\!x_{1:n},\infty)\!=
\!\varphi(y\!x_{1:n},\infty)\!=\!\rho(y\!x_{1:n})$ by definition
(\ref{ccsm1}). $\phi(\epsilon,t)\!=\!\varphi'(\epsilon,t)$ by
(\ref{ccsm2}) and the fact that
$\varphi'(\epsilon,i\!-\!1)<\varphi'(\epsilon,i)\!\leq\!
\varphi(\epsilon,i)\!\leq\!\rho(\epsilon)\!\leq\!1$, hence
$\phi(\epsilon,\infty)\!=\!\rho(\epsilon)$. $\phi(y\!\pb
x_{1:n},t)\!\leq\!\varphi'(y\!x_{1:n},t)$ by (\ref{ccsm3}), hence
$\phi(y\!\pb x_{1:n},\infty)\!\leq\!\rho(y\!\pb x_{1:n})$. We prove
the opposite direction $\phi(y\!\pb
x_{1:n},\infty)\!\geq\!\rho(y\!x_{1:n})$ by induction over $n$. We
have
\beq\label{upineq}
  \sum_{x_n}\varphi'(y\!x_{1:n},i) \;\leq\;
  \sum_{x_n}\varphi(y\!x_{1:n},i)  \;<\;
  \sum_{x_n}\varphi(y\!x_{1:n},\infty) \;=\;
  \sum_{x_n}\rho(y\!x_{1:n}) \;\leq\; \rho(y\!\pb x_{<n})
\eeq
The strict monotony of $\varphi$ and the semimeasure
property of $\rho$ have been used. By induction hypothesis
$\lim_{t\to\infty}\phi(y\!\pb x_{<n},t)\!\geq\!\rho(y\!\pb x_{<n})$ and
(\ref{upineq}) for sufficiently large $t$ we have
$\phi(y\!\pb x_{<n},t)\!>\!\sum_{x_n}\varphi'(y\!x_{1:n},i)$. The
condition in (\ref{ccsm3}) is, hence, satisfied and therefore
$\phi(y\!\pb x_{1:n},t)\!\geq\!\varphi'(y\!x_{1:n},i)$ for sufficiently
large $t$, especially
$\phi(y\!\pb x_{1:n},\infty)\!\geq\!\varphi'(y\!x_{1:n},i)$ for all $i$.
Taking the limit $i\!\to\!\infty$ we get
$\phi(y\!\pb x_{1:n},\infty)\!\geq\!\varphi'(y\!x_{1:n},\infty)\!=\!\rho(y\!\pb x_{1:n})$.

Combining all results, we have shown that the constructed
$\phi(\cdot,t)$ are primitive recursive chronological semimeasures
monotone increasing in $t$, which converge to the enumerable
chronological semimeasure $\rho$. This finally proves the
enumerability of the set of enumerable chronological
semimeasures.

\paragraph{Convergence of $\xi^{AI}$ to $\mu^{AI}$:}
In \cite{Hut00e} the following inequality is proved
\beq\label{entro2}
  2\sum_{i=1}^{|X|} y_i(y_i\!-\!z_i)^2 \;\leq\!
  \sum_{i=1}^{|X|} y_i\ln{y_i\over z_i} \quad\mbox{with}\quad
  \sum_{i=1}^{|X|} y_i=1, \quad \sum_{i=1}^{|X|} z_i\leq 1
\eeq
If we identify $i\!=\!x_k$ and $y_i\!=\!\mu(y\!x_{<k}y\!\pb x_k)$ and
$z_i\!=\!\xi(y\!x_{<k}y\!\pb x_k)$, multiply both sides with
$\mu(y\!\pb x_{<k})$, take the sum over $x_{<k}$, then the sum
over $k$ and use Bayes' rule $\mu(y\!\pb x_{<k})\!\cdot\!\mu(y\!x_{<k}y\!\pb
x_k)=\mu(y\!\pb x_{1:k})$ we get
\beq\label{eukdistxi}
  2\sum_{k=1}^n\sum_{x_{1:k}}\mu(y\!\pb x_{1:k})
  \Big(\mu(y\!x_{<k}\pb x_k)-\xi(y\!x_{<k}\pb x_k)\Big)^2 \;\leq\;
  \sum_{k=1}^n\sum_{x_{1:k}}\mu(y\!\pb x_{1:k})
  \ln{\mu(y\!x_{<k}\pb x_k)\over\xi(y\!x_{<k}\pb x_k)}
  =\; ...
\eeq
In the r.h.s.\ we can replace $\sum_{x_{1:k}}\mu(y\!\pb
x_{1:k})$ by $\sum_{x_{1:n}}\mu(y\!\pb x_{1:n})$ as the argument
of the logarithm is independent of $x_{k+1:n}$. The $k$ sum can now be
brought into the logarithm and converts to a product. Using Bayes'
rule (\ref{bayes2}) for $\mu$ and $\xi$ we get
\beq\label{eukdistxi2}
  ...\;=\;
  \sum_{x_{1:n}}\mu(y\!\pb x_{1:n})
  \ln\prod_{k=1}^n{\mu(y\!x_{<k}\pb x_k)\over\xi(y\!x_{<k}\pb x_k)}
  \;=\;
  \sum_{x_{1:n}}\mu(y\!\pb x_{1:n})
  \ln{\mu(y\!\pb x_{1:n})\over\xi(y\!\pb x_{1:n})}
  \;\stackrel{+}{<}\; \ln 2\!\cdot\!K(\mu)
\eeq
where we have used the universality property (\ref{uniaixi})
of $\xi$ in the last step. The main complication for generalizing
(\ref{eukdist}) to (\ref{eukdistxi},\ref{eukdistxi2}) was the
generalization of (\ref{entro2}) from $|X|\!=\!2$ to a general
alphabet, the $y$ are, again, pure spectators. This will change when
we analyze error/credit bounds analog to (\ref{spebound}).

(\ref{eukdistxi},\ref{eukdistxi2}) shows that the $\mu$ expected
squared difference of $\mu$ and $\xi$ is finite for computable
$\mu$. This, in turn, shows that $\xi(y\!x_{<k}y\!\pb x_k)$
converges to $\mu(y\!x_{<k}y\!\pb x_k)$ for $k\!\to\!\infty$ with $\mu$
probability 1. If we take a finite product of $\xi's$ and use
Bayes' rule, we see that also $\xi(y\!x_{<k}y\!\pb x_{k:k+r})$
converges to $\mu(y\!x_{<k}y\!\pb x_{k:k+r})$. More generally, in case of
a bounded horizon $h_k$, it follows that
\beq\label{aixitomu}
  \xi(y\!x_{<k}y\!\pb x_{k:m_k}) \toinfty{k} \mu(y\!x_{<k}y\!\pb x_{k:m_k})
  \quad\mbox{if}\quad h_k\equiv m_k\!-\!k\!+\!1 \leq h_{max} < \infty
\eeq
This gives makes us confident that the outputs $\hh y_k$
of the AI$\xi$ model (\ref{ydotxi}) could converge to the outputs $\hh
y_k$ from the AI$\mu$ model (\ref{ydotrec}), at least for bounded
horizon.

We want to call an AI model {\it universal}, if it is $\mu$
independent (unbiased, model-free) and is able
to solve any solvable problem and learn any learnable task.
Further, we call a universal model, {\it universally optimal}, if
there is no program, which can solve or learn significantly faster
(in terms of interaction cycles). As the AI$\xi$ model is
parameterless, $\xi$ converges to $\mu$ (\ref{aixitomu}), the
AI$\mu$ model is itself optimal, and we expect no other model to
converge faster to AI$\mu$ by analogy to SP (\ref{spebound}),
\beqn
  \mbox{\it we expect AI$\xi$ to be universally optimal.}
\eeqn
This is our main claim. In a sense, the intention of the remaining
(sub)sections is to define this statement more rigorously and
to give further support.

\paragraph{Intelligence order relation:}
We define the $\xi$ expected credit in cycles $k$ to $m$ of a
policy $p$ similar to (\ref{eefunc}) and (\ref{eefuncxi}).
We extend the definition to programs $p\!\not\in\!\hh P_k$ which
are not consistent with the current history.
\beq\label{cxi}
  C^\xi_{km}(p|\hh y\!\hh x_{<k}) \;:=\;
  {1\over\cal N}
  \sum_{q:q(\hh y_{<k})=\hh x_{<k}}
  \nq 2^{-l(q)}\cdot C_{km}(\tilde p,q)
\eeq
The normalization $\cal N$ is again only necessary for
interpreting $C_{km}$ as the expected credit but otherwise
unneeded. For consistent policies $p\!\in\!\hh P_k$ we define
$\tilde p\!:=\!p$. For $p\!\not\in\!\hh P_k$, $\tilde p$ is a
modification of $p$ in such a way that its output is consistent
with the current history $\hh y\!\hh x_{<k}$, hence $\tilde
p\!\in\!\hh P_k$, but unaltered for the current and future cycles
$\geq\!k$. Using this definition of $C_{km}$ we could take the
maximium over all systems $p$ in (\ref{eefuncxi}), rather than only the
consistent ones.

We call $p$ {\it more or equally intelligent} than $p'$ if
\beq\label{aiorder}
  p\succeq p' \;:\Leftrightarrow
  \forall k\forall\hh y\!\hh x_{<k}:
  C^\xi_{km_k}(p|\hh y\!\hh x_{<k}) \geq
  C^\xi_{km_k}(p'|\hh y\!\hh x_{<k})
\eeq
i.e.\ if $p$ yields in any circumstance higher $\xi$ expected
credit than $p'$. As the algorithm $p^\best$ behind the AI$\xi$
system maximizes $C^\xi_{km_k}$ we have $p^\best\!\succeq\!p$ for all
$p$. The AI$\xi$ model is hence the most intelligent system
w.r.t.\ $\succeq$. $\succeq$ is a universal order relation in the
sense that it is free of any parameters (except $m_k$) or specific
assumptions about the environment. A proof, that $\succeq$ is a
reliable intelligence order (what we believe to be true), would
prove that AI$\xi$ is universally optimal. We could further ask:
how useful is $\succeq$ for ordering policies of practical
interest with intermediate intelligence, or how can $\succeq$ help
to guide toward constructing more intelligent systems with
reasonable computation time. An effective intelligence order
relation $\succeq^c$ will be defined in section \ref{secTime},
which is more useful from a practical point of view.

\paragraph{Credit bounds and separability concepts:}
The credits $C_{km}$ associated with the AI systems correspond
roughly to the negative error measure $-E_{n\rho}$ of the SP
systems. In SP, we were interested in small bounds for the error
excess $E_{n\Theta_\xi}\!-\!E_{n\rho}$. Unfortunately, simple
credit bounds for AI$\xi$ in terms of $C_{km}$ analog to the error
bound (\ref{spebound}) do not hold. We even have difficulties in
specifying what we can expect to hold for AI$\xi$ or any AI system
which claims to be universally optimal. Consequently, we cannot
have a proof if we don't know what to prove. In SP, the only
important property of $\mu$ for proving error bounds was its
complexity $K(\mu)$. We will see that in the AI case, there are no
useful bounds in terms of $K(\mu)$ only. We either have to study
restricted problem classes or consider bounds depending on other
properties of $\mu$, rather than on its complexity only. In the
following, we will exhibit the difficulties by two examples and
introduce concepts which may be useful for proving credit bounds.
Despite the difficulties in even claiming useful credit bounds, we
nevertheless, firmly believe that the order relation
(\ref{aiorder}) correctly formalizes the intuitive meaning of
intelligence and, hence, that the AI$\xi$ system is universally optimal.

In the following, we choose $m_k\!=\!T$. We want to compare the
true, i.e. $\mu$ expected credit $C^\mu_{1T}$ of a $\mu$
independent universal policy $p^{best}$ with any other policy $p$.
Naively, we might expect the existence of a policy $p^{best}$ which
maximizes $C^\mu_{1T}$, apart from additive
corrections of lower order for $T\!\to\!\infty$
\beq\label{cximu}
  C^\mu_{1T}(p^{best}) \;\geq\; C^\mu_{1T}(p) - o(...)
  \quad \forall\mu,p
\eeq
Note, that the policy $p^{*\xi}$ of the AI$\xi$ system
maximizes $C^\xi_{1T}$ by definition ($p^{*\xi}\succeq p$). As
$C^\xi_{1T}$ is thought to be a guess of $C^\mu_{1T}$, we might
expect $p^{best}\!=\!p^{*\xi}$ to approximately maximize
$C^\mu_{1T}$, i.e. (\ref{cximu}) to hold. Let us consider the
problem class (set of environments) $\{\mu_0,\mu_1\}$ with
$Y\!=\!C\!=\{0,1\}$ and $c_k\!=\delta_{iy_1}$ in environment
$\mu_i$. The first output $y_1$ decides whether you go to heaven
with all future credits $c_k$ being $1$ (good) or to hell with all
future credits being $0$ (bad). It is clear, that if
$\mu_i$, i.e. $i$ is known, the optimal policy $p^{*\mu_i}$
is to output $y_1\!=\!i$ in the first cycle with
$C^\mu_{1T}(p^{*\mu_i})\!=\!T$. On the other hand, any unbiased
policy $p^{best}$ independent of the actual $\mu$ either outputs
$y_1\!=\!1$ or $y_1\!=\!0$. Independent of the actual choice
$y_1$, there is always an environment ($\mu\!=\!\mu_{1-y_1}$)
for which this choice is catastrophic
($C^\mu_{1T}(p^{best})\!=\!0$). No single system can perform well in both
environments $\mu_0$ {\it and} $\mu_1$. The r.h.s.\ of
(\ref{cximu}) equals $T\!-\!o(T)$ for $p\!=\!p^{*\mu}$. For all
$p^{best}$ there is a $\mu$ for which the l.h.s.\ is zero. We have
shown that no $p^{best}$ can satisfy (\ref{cximu}) for all $\mu$
and $p$, so we cannot expect $p^{*\xi}$ to do so. Nevertheless,
there are problem classes for which (\ref{cximu}) holds, for
instance SP and CF. For SP, (\ref{cximu}) is just a reformulation
of (\ref{spebound}) with an appropriate choice for $p^{best}$
(which differs from $p^{*\xi}$, see next section). We expect
(\ref{cximu}) to hold for all inductive problems in which the
environment is not influenced\footnote{Of course, the credit
feedback $c_k$ depends on the system's output. What we have in mind
is, like in sequence prediction, that the true sequence is not
influenced by the system} by the output of the system. We want to
call these $\mu$, {\it passive} or {\it inductive} environments.
Further, we want to call $\mu$ satisfying (\ref{cximu}) with
$p^{best}\!=\!p^{*\xi}$ {\it pseudo passive}. So we expect
inductive $\mu$ to be pseudo passive.

Let us give a further example to demonstrate the difficulties in
establishing credit bounds. Let $C\!=\{0,1\}$ and $|Y|$ be large. We
consider all (deterministic) environments in which a single complex output
$y^*$ is correct ($c\!=\!1$) and all others are wrong ($c\!=\!0$).
The problem class $M$ is defined by
$$
  M:=\{\mu:\mu(y\!x_{<k}y_k\pb 1)=
       \delta_{y_ky^*},\; y^*\!\in\!Y,\; K(y^*)\!=\!_\lfloor\log_2|Y|_\rfloor\}
$$
There are $N\stackrel\times=|Y|$ such $y^*$. The only way a
$\mu$ independent policy $p$ can find the correct $y^*$´, is
by trying one $y$ after the other in a certain order. In the first
$N\!-\!1$ cycles at most, $N\!-\!1$ different $y$ are tested. As
there are $N$ different possible $y^*$, there is always a
$\mu\!\in\!M$ for which $p$ gives erroneous outputs in the first
$N\!-\!1$ cycles. The number of errors are $E_{\infty
p}\!\geq\!N\!-\!1\!\stackrel\times=|Y|\stackrel\times=2^{K(y^*)}\stackrel\times=2^{K(\mu)}$
for this $\mu$. As this is true for any $p$, it is also true
for the AI$\xi$ model, hence $E_{k\xi}\!\leq\!2^{K(\mu)}$ is the
best possible error bound we can expect, which depends on $K(\mu)$
only. Actually, we will derive such a bound in section
\ref{secSP} for SP. Unfortunately, as we are mainly interested in
the cycle region $k\ll|Y|\stackrel\times=2^{K(\mu)}$ (see section
\ref{secAImurec}) this bound is trivial.
There are no interesting bounds depending on $K(\mu)$
only, unlike the SP case for deterministic $\mu$. Bounds must
either depend on additional properties of $\mu$ or we have to
consider specialized bounds for restricted problem classes. The
case of probabilistic $\mu$ is similar. Whereas for SP there are
useful bounds in terms of $E_{k\mu}$ and $K(\mu)$, there are no
such bounds for AI$\xi$. Again, this is not a drawback of AI$\xi$
since for no unbiased AI system the errors/credits could be bound in
terms of $K(\mu)$ and the errors/credits of AI$\mu$ only.

There is a way to make use of gross (e.g. $2^{K(\mu)}$) bounds.
Assume that after a reasonable number of cycles $k$, the
information $\hh x_{<k}$ perceived by the AI$\xi$ system contains
a lot of information about the true environment $\mu$. The
information in $\hh x_{<k}$ might be coded in any form. Let us
assume that the complexity $K(\mu|\hh x_{<k})$ of $\mu$ under the
condition that $\hh x_{<k}$ is known, is of order 1. Consider a
theorem, bounding the sum of credits or of other quantities over
cycles $1...\infty$ in terms of $f(K(\mu))$ for a function $f$
with $f(O(1))\!=\!O(1)$, like $f(n)\!=\!2^n$. Then, there will be
a bound for cycles $k...\infty$ in terms of $f(K(\mu|\hh
x_{<k}))\!=\!O(1)$. Hence, a bound like $2^{K(\mu)}$ can be
replaced by small bound $2^{K(\mu|\hh x_{<k})}\!=\!O(1)$ after
a reasonable number of cycles. All one has to
show/ensure/assume is that enough information about $\mu$ is
presented (in any form) in the first $k$ cycles. In this way, even
a gross bound could become useful. In section \ref{secEX} we use a
similar argument to prove that AI$\xi$ is able to learn
supervised.

In the following, we weaken (\ref{cximu}) in the hope of getting a
bound applicable to wider problem classes than the passive one.
Consider the I/O sequence $\hh y_1\hh x_1...\hh y_n\hh x_n$ caused
by AI$\xi$. On history $\hh y\!\hh x_{<k}$, AI$\xi$ will output
$\hh y_k\!\equiv\hh y^\xi_k$ in cycle $k$. Let us compare this to
$\hh y^\mu_k$ what AI$\mu$ would output, still on the same history
$\hh y\!\hh x_{<k}$ produced by AI$\xi$. As AI$\mu$ maximizes the
$\mu$ expected credit, AI$\xi$ causes lower (or at best equal)
$C^\mu_{km_k}$, if $\hh y^\xi_k$ differs from $\hh y^\mu_k$. Let
$D_{n\mu\xi}\!:=\!\langle\sum_{k=1}^n 1\!-\!\delta_{\hh
y^\mu_k,\hh y^\xi_k}\rangle_\mu$ be the $\mu$ expected number of
suboptimal choices of AI$\xi$, i.e. outputs different from AI$\mu$
in the first $n$ cycles. One might weigh the deviating cases by
their severity. Especially when the $\mu$ expected credits
$C^\mu_{km_k}$ for $\hh y^\xi_k$ and $\hh y^\mu_k$ are equal or
close to each other, this should be taken into account in the
definition of $D_{n\mu\xi}$. These details do not matter in the
following qualitative discussion. The important difference to
(\ref{cximu}) is that here we stick on the history produced by
AI$\xi$ and count a wrong decision as, at most, one error. The
wrong decision in the Heaven\&Hell example in the first cycle no
longer counts as losing $T$ credits, but counts as one wrong
decision. In a sense, this is fairer. One shouldn't blame somebody
too much who makes a single wrong decision for which he just has
too little information available, in order to make a correct
decision. The AI$\xi$ model would deserve to be called
asymptotically optimal, if the probability of making a wrong
decision tends to zero, i.e.\ if
\beq\label{Doon}
  D_{n\mu\xi}/n\to 0 \quad\mbox{for}\quad n\to\infty, \quad\mbox{i.e.}\quad
  D_{n\mu\xi} \;=\; o(n).
\eeq
We say that $\mu$ can be {\it asymptotically learned} (by AI$\xi$)
if (\ref{Doon}) is satisfied. We claim that AI$\xi$ (for
$m_k\!\to\!\infty$) can asymptotically learn every problem $\mu$
of relevance, i.e. AI$\xi$ is asymptotically optimal. We included
the qualifier {\it of relevance}, as we are not sure whether there
could be strange $\mu$ spoiling (\ref{Doon}) but we expect those
$\mu$ to be irrelevant from the perspective of AI. In the field of
Learning, there are many asymptotic learnability theorems, often
not too difficult to prove. So a proof of (\ref{Doon}) might also
be accessible. Unfortunately, asymptotic learnability theorems are
often too weak to be useful from a practical point. Nevertheless,
they point in the right direction.

From the convergence (\ref{aixitomu}) of $\mu\!\to\!\xi$ we might
expect $C^\xi_{km_k}\!\to\!C^\mu_{km_k}$ and hence, $\hh y^\xi_k$
defined in (\ref{ydotxi}) to converge to $\hh y^\mu_k$ defined in
(\ref{ydotrec}) with $\mu$ probability 1 for $k\!\to\!\infty$.
The first problem is, that if the $C_{km_k}$ for
the different choices of $y_k$ are nearly equal, then even if
$C^\xi_{km_k}\!\approx\!C^\mu_{km_k}$, $\hh y^\xi_k\!\neq\!\hh
y^\mu_k$ is possible due to the non-continuity of $\maxarg_{y_k}$. This
can be cured by a weighted $D_{n\mu\xi}$ as described above. More
serious is the second problem we explain for $h_k\!=\!1$ and
$X\!=\!C\!=\!\{0,1\}$. For $\hh
y^\xi_k\!\equiv\!\maxarg_{y_k}\xi(\hh y\!\hh c_{<k}y_k\pb 1)$ to
converge to $\hh y^\mu_k\!\equiv\!\maxarg_{y_k}\mu(\hh y\!\hh
c_{<k}y_k\pb 1)$, it is not sufficient to know that $\xi(\hh
y\!\hh c_{<k}\hh y\!\hh{\pb c}_k)\!\to\!\mu(\hh y\!\hh c_{<k}\hh
y\!\hh{\pb c}_k)$ as has been proved in (\ref{aixitomu}). We need
convergence not only for the true output $\hh y_k$ and credit $\hh
c_k$, but also for alternate outputs $y_k$ and credit 1.
$\hh y^\xi_k$ converges to $\hh y^\mu_k$
if $\xi$ converges uniformly to $\mu$, i.e. if in addition to
(\ref{aixitomu})
\beq\label{uniform}
  \big|\mu(y\!x_{<k}y'_k\pb x'_k)-\xi(y\!x_{<k}y'_k\pb x'_k)\big|
  \;<\; c\!\cdot\!
  \big|\mu(y\!x_{<k}y\!\pb x_k)-\xi(y\!x_{<k}y\!\pb x_k)\big|
  \quad\forall y'_kx'_k
\eeq
holds for some constant $c$ (at least in some $\mu$ expected sense).
We call $\mu$ satisfying (\ref{uniform}) {\it uniform}. For
uniform $\mu$ one can show (\ref{Doon}) with appropriately weighted
$D_{n\mu\xi}$ and bounded horizon $h_k\!<\!h_{max}$. Unfortunately
there are relevant $\mu$ which are not uniform.
Details will be given elsewhere.

In the following, we briefly mention some further
concepts. A {\it Markovian} $\mu$ is defined as depending only on the
last output, i.e. $\mu(y\!x_{<k}y\!\pb x_k)\!=\!\mu_k(y\!\pb x_k)$. We
say $\mu$ is {\it generalized Markovian}, if $\mu(y\!x_{<k}y\!\pb
x_k)\!=\!\mu_k(y\!x_{k-l:k-1}y\!\pb x_k)$ for fixed $l$. This
property has some similarities to {\it factorizable} $\mu$ defined
in (\ref{facmu}). If further $\mu_k\!\equiv\!\mu_1\forall k$,
$\mu$ is called {\it stationary}. Further, for all enumerable
$\mu$, $\mu(y\!x_{<k}y\!\pb x_k)$ and $\xi(y\!x_{<k}y\!\pb x_k)$
get independent of $y\!x_{<l}$ for fixed $l$ and $k\!\to\!\infty$
with $\mu$ probability 1. This property, which we want to call
{\it forgetfulness}, will be proved elsewhere.
Further, we say $\mu$ is {\it farsighted}, if
$\lim_{m_k\to\infty}\hh y_k^{(m_k)}$ exists. More details will be given in
the next subsection, where we also give an example of a
possibly relevant $\mu$, which is not farsighted.

We have introduced several concepts, which might be useful for
proving credit bounds, including forgetful, relevant, asymptotically
learnable, farsighted, uniform, (generalized) Markovian, factorizable
and (pseudo) passive $\mu$. We have sorted them here, approximately in
the order of decreasing generality. We want to call them {\it
separability concepts}. The more general (like relevant,
asymptotically learnable and farsighted) $\mu$ will be called
weakly separable, the more restrictive (like (pseudo) passive and
factorizable) $\mu$ will be called strongly separable, but we will
use these qualifiers in a more qualitative, rather than rigid
sense. Other (non-separability) concepts are deterministic $\mu$
and, of course, the class of all chronological $\mu$.

\paragraph{The choice of the horizon:}
The only significant arbitrariness in the AI$\xi$ model lies in
the choice of the horizon function
$h_k\!\equiv\!m_k\!-\!k\!+\!1$. We discuss some choices which seem
to be natural and give preliminary conclusions at the end.
We will not discuss ad hoc choices of $h_k$ for
specific problems (like the discussion in section \ref{secSG} in
the context of finite games). We are interested in universal
choices of $m_k$.

If the lifetime of the system is known to be $T$, which is in
practice always large but finite, then the choice $m_k\!=\!T$
maximizes correctly the expected future credit. $T$ is usually not
known in advance, as in many cases the time we are willing to run
a system depends on the quality of its outputs. For this reason,
it is often desirable that good outputs are not delayed too much,
if this results in a marginal credit increase only. This can be
incorporated by damping the future credits. If, for instance, we
assume that the survival of the system in each cycle is
proportional to the past credit an exponential damping
$c_k\!:=\!c'_k\!\cdot\!e^{-\lambda k}$ is appropriate, where
$c'_k$ are bounded, e.g. $c'_k\!\in\![0,1]$. The expression
(\ref{ydotxi}) converges for $m_k\!\to\!\infty$ in this case. But
this does not solve the problem, as we introduced a new arbitrary
time-scale $^1\!/_\lambda$. Every damping introduces a time-scale.

Even the time-scale invariant damping factor $k^{-\alpha}$
introduces a dynamic time-scale. In cycle $k$ the contribution of
cycle $2^{1/\alpha}\!\cdot\!k$ is damped by a factor $\1d2$. The
effective horizon $h_k$ in this case is $\sim k$. The choice
$h_k\!=\!\beta\!\cdot\!k$ with $\beta\!\sim\!2^{1/\alpha}$
qualitatively models the same behaviour. We have not introduced an
arbitrary time-scale $T$, but limited the farsightedness to some
multiple (or fraction) of the length of the current history. This
avoids the pre-selection of a global time-scale $T$ or
$^1\!/_\lambda$. This choice has some appeal, as it seems that
humans of age $k$ years usually do not plan their lives for more
than, perhaps, the next $k$ years ($\beta_{human}\!=\!1$). From a
practical point of view this model might serve all needs, but from
a theoretical point we feel uncomfortable with such a limitation
in the horizon from the very beginning. Note, that we have to
choose $\beta\!=\!O(1)$ because otherwise we would again introduce
a number $\beta$, which has to be justified.

The naive limit $m_k\!\to\!\infty$ in
(\ref{ydotxi}) may turn out to be well defined and the previous discussion
superfluous. In the following, we define a limit which is always
well defined (for finite $|Y|$). Let $\hh y_k^{(m)}$ be defined as
in (\ref{ydotxi}) with $m_k$ replaced by $m$. Further, let $\hh
Y_k^{(m)}\!:=\!\{\,\hh y_k^{(m)}\!:\!m_k\!\geq\!m\}$ be the set of
outputs in cycle $k$ for the choices $m_k\!=\!m,m+1,m+2,...$.
Because $\hh Y_k^{(m)}\!\supseteq\!\hh Y_k^{(m+1)}\!\neq\!\{\}$, we
have $\hh Y_k^{(\infty)}\!:=\!\bigcap_{m=k}^\infty\hh
Y_k^{(m)}\!\neq\!\{\}$. We define the $m_k\!=\!\infty$ model to
output any $\hh y_k^{(\infty)}\!\in\!\hh Y_k^{(\infty)}$. This is
the best output consistent with any choice of $m_k$, esp.
$m_k\!\to\!\infty$. Choosing the lexicographically smallest $\hh
y_k^{(\infty)}\!\in\!\hh Y_k^{(\infty)}$ would correspond to the
limes inferior $\underline\lim_{m\to\infty}\hh y_k^{(m)}$. $\hh
y_k^{(\infty)}$ is unique, i.e. $|\hh Y_k^{(\infty)}|\!=\!1$ iff
the naive limit $\lim_{m\to\infty}\hh y_k^{(m)}$ exists. Note,
that the limit $\lim_{m\to\infty}C_{km}^\best(y\!x_{<k})$ needs
not to exist for this construction.

The construction above leads to a mathematically elegant,
no-parameter AI$\xi$ model. Unfortunately this is not the end of
the story. The limit $m_k\!\to\!\infty$ can cause undesirable
results in the AI$\mu$ model for special $\mu$ which might also happen
in the AI$\xi$ model whatever we define $m_k\!\to\!\infty$.
Consider $Y\!=\!C\!=\!\{0,1\}$ and $X'\!=\!\{\}$. Output $y_k\!=\!0$ shall give credit
$c_k\!=\!0$, output $y_k\!=\!1$ shall give $c_k\!=\!1$ iff $\hh
y_{k-l-\sqrt l}...\hh y_{k-l}\!=\!0...0$ for some $l$. I.e. the system can
achieve $l$ consecutive positive credits if there was a sequence
of length at least $\sqrt l$ with $y_k\!=\!c_k\!=\!0$. If the lifetime of the
AI$\mu$ system is $T$, it outputs $\hh y_k\!=\!0$ in the first $r$ cycles
and then $\hh y_k\!=\!1$ for the remaining $r^2$ cycles with
$r$ such that $r+r^2=T$. This will lead to the highest possible
total credit $C_{1T}\!=\!\sqrt{T+^1\!\!/_4}-^1\!\!/_2$. Any fragmentation of the
$0$ and $1$ sequences would reduce this. For $T\!\to\!\infty$ the
AI$\mu$ system can and will delay the point $r$ of switching to
$\hh y_k\!=\!1$ indefinitely and always output $0$ with total
credit $0$, obviously the worst possible behaviour. The AI$\xi$
system will explore the above rule after a while of trying
$y_k\!=\!0/1$ and then applies the same behaviour as the AI$\mu$
system, since the simplest rules covering past data dominate $\xi$.
For finite $T$ this is exactly what we want, but for infinite $T$
the AI$\xi$ model fails just as the AI$\mu$ model does. The good point
is, that this is not a weakness of the AI$\xi$ model, as AI$\mu$
fails too and no system can be better than AI$\mu$. The bad point
is that $m_k\!\to\!\infty$ has far reaching consequences, even when
starting from an already very large $m_k\!=\!T$. The reason being that
the $\mu$ of this example is highly non-local in time, i.e. it
may violate one of our weak separability conditions.

In the last paragraph we have considered the consequences of
$m_k\!\to\!\infty$ in the AI$\mu$ model. We now consider
whether the AI$\xi$ model is a good approximation of the
AI$\mu$ model for large $m_k$. Another objection against too large
choices of $m_k$ is that $\xi(y\!x_{<k}y\!\pb x_{k:m_k})$ has been proved to be a
good approximation of $\mu(y\!x_{<k}y\!\pb x_{k:m_k})$ only for
$k\!\gg\!h_k$, which is never satisfied for $m_k\!=\!T$ or
$m_k\!=\!\infty$. We have seen that, for factorizable
$\mu$, the limit $h_k\!\to\!\infty$ causes
no problem, as from a certain $h_k$ on the output $\hh y_k$ is
independent of $h_k$. As $\xi\!\to\!\mu$ for bounded $h_k$, $\xi$
will develop this separability property too. So, from a
certain $k_0$ on the limit $h_k\!\to\!\infty$ might also be safe
for $\xi$. Therefore, taking the limit from the very beginning worsens
the behaviour of AI$\xi$ maybe only for finitely many cycles
$k\!\leq k_0$, which would be acceptable. We suppose that the
valuations $c_{k'}$ for $k'\!\gg\!k$, where $\xi$ can no longer
be trusted as a good approximation to $\mu$, are in some sense
randomly disturbed with decreasing influence on the choice of $\hh
y_k$. This claim is supported by the forgetfulness property of $\xi$.

We are not sure whether the choice of $m_k$ is of marginal
importance, as long as $m_k$ is chosen sufficiently large and of
low complexity, $m_k=2^{2^{16}}$ for instance, or whether the choice of
$m_k$ will turn out to be a central topic for the AI$\xi$ model or
for the planning aspect of any AI system in general. We suppose
that the limit $m_k\!\to\!\infty$ for the AI$\xi$ model results in
correct behaviour for weakly separable $\mu$, and that even the naive
limit exists, but to prove this would probably give interesting
insights.

\newpage
\section{Sequence Prediction (SP)}\label{secSP}
We have introduced the
AI$\xi$ model as a unification of the ideas of decision theory and
universal probability distribution. We might expect AI$\xi$ to
behave identically to SP$\Theta_\xi$, when faced with a sequence
prediction problem, but things are not that simple, as we will see.

\paragraph{Using the AI$\mu$ Model for Sequence Prediction:}
We have seen in the last section how to predict sequences for
known and unknown prior distribution $\mu^{SP}$. Here we consider binary
sequences\footnote{We use $z_k$ to avoid notational conflicts with
the systems inputs $x_k$.} $z_1z_2z_3...\in I\!\!B^\infty$ with known prior
probability $\mu^{SP}(\pb{z_1z_2z_3...})$.

We want to show
how the AI$\mu$ model can be used for sequence prediction.
We will see that it gives the same prediction as the SP$\Theta_\mu$ system.
First, we have to specify {\it how} the AI$\mu$ model should be used
for sequence prediction. The following choice is natural:

The systems output $y_k$ is interpreted as a prediction for the
$k^{th}$ bit $z_k$ of the string, which has to be predicted. This
means that $y_k$ is binary ($y_k\!\in\!I\!\!B\!=:\!Y$). As a
reaction of the environment, the system receives credit $c_k\!=\!1$
if the prediction was correct ($y_k\!=\!z_k$), or $c_k\!=\!0$ if
the prediction was erroneous ($y_k\!\neq\!z_k$). The question is
what the input $x'_k$ of the next cycle should be. One choice
would be to inform the system about the correct $k^{th}$ bit of
the last cycle of the string and set $x'_k=z_k$. But as from
the credit $c_k$ in conjunction with the prediction $y_k$, the true
bit $z_k=\delta_{y_kc_k}$ can be inferred, this information is
redundant. $\delta$ is the Kronecker symbol, defined as
$\delta_{ab}\!=\!1$ for $a\!=\!b$ and $0$ otherwise. There is no
need for this additional feedback. So we set
$x'_k\!=\!\epsilon\!\in\!X\!=\!\{\epsilon\}$ thus having $x_k\!\equiv\!c_k$. The
system's performance does not change when we include this
redundant information, it merely complicates the notation. The prior
probability $\mu^{AI}$ of the AI$\mu$ model is
\beq\label{muaisp}
  \mu^{AI}(y_1\pb x_1 ...y_k\pb x_k) \;=\;
  \mu^{AI}(y_1\pb c_1...y_k\pb c_k) \;=\;
  \mu^{SP}(\pb{\delta_{y_1 c_1}...\delta_{y_k c_k}}) \;=\;
  \mu^{SP}(\pb{z_1...z_k})
\eeq
In the following, we will drop the superscripts of $\mu$ because they
are clear from the arguments of $\mu$ and the $\mu$ equal in any case.

The formula (\ref{airec2}) for the expected credit reduces to
\beq\label{eerecsp}
  C_{km}^\best(y\!x_{<k}) \;=\;
  \max_{y_k}\sum_{c_k}
  [c_k+C_{k+1,m}^\best(y\!x_{1:k})] \!\cdot\!
  \mu(\delta_{y_1c_1}...\delta_{y_{k-1}c_{k-1}}\pb{\delta_{y_kc_k}})
\eeq
The first observation we can make, is that for this special
$\mu$, $C_{km}^\best$ only depends on $\delta_{y_ic_i}$, i.e.
replacing $y_i$ and $c_i$ simultaneously with their complements
does not change the value of $C_{km}^\best$. We have a symmetry in
$y_ic_i$. For $k\!=\!m\!+\!1$ this is definitely true as
$C_{m+1,m}^\best\!=\!0$ in this case (see (\ref{ee0})). For
$k\!\leq\!m$ we prove it by induction. The r.h.s.\ of
(\ref{eerecsp}) is symmetric in $y_ic_i$ for $i\!<\!k$ because
$\mu$ possesses this symmetry and $C_{k+1,m}^\best$ possesses it by induction
hypothesis, so the symmetry holds for the l.h.s., which completes
the proof. The prediction $\hh y_k$ is
\beq\label{ebestysp}
  \hh y_k \;=\; \maxarg_{y_k}
  C_{km_k}^\best(\hh y\!\hh x_{<k}y_k) \;=\;
  \maxarg_{y_k}\sum_{c_k}[c_k+C_{k+1,m_k}^\best(y\!x_{1:k})]
  \!\cdot\!\mu(...\pb{\delta_{y_kc_k}}) \;=\;
\eeq
$$
  \;=\; \maxarg_{y_k}\sum_{c_k}c_k
  \!\cdot\!\mu(\delta_{\hh y_1\hh c_1}...\pb{\delta_{y_kc_k}}) \;=\;
  \maxarg_{y_k}\mu(\hh z_1...\hh z_{k-1}\pb y_k) \;=\;
  \maxarg_{z_k}\mu(\hh z_1...\hh z_{k-1}\pb z_k)
$$
The first equation is the definition of the system's prediction
(\ref{pbestrec}). In the second equation, we have inserted
(\ref{ebesty}) which gives the r.h.s.\ of (\ref{eerecsp}) with
$\max_{y_k}$ replaced by $\maxarg_{y_k}$. $\sum_c
f(...\delta_{yc}...)$ is independent of $y$ for any function,
depending on the combination $\delta_{yc}$ only. Therefore, the
$\sum_cC^\best\mu$ term is independent of $y_k$ because
$C_{k+1,m}^\best$ as well as $\mu$ depend on $\delta_{y_kc_k}$ only. In
the third equation, we can therefore drop this term, as adding a
constant to the argument of $\maxarg_{y_k}$ does not change the
location of the maximum. In the second last equation we evaluated
the $\sum_{c_k}$. Further, if the true credit to $\hh y_i$ is $\hh
c_i$ the true $i^{th}$ bit of the string must be $\hh
z_i\!=\!\delta_{\hh y_i\hh c_i}$. The last equation is just a renaming.

So, the AI$\mu$ model predicts that $z_k$ that has maximal $\mu$
probability, given $\hh z_1...\hh z_{k-1}$. This prediction is
independent of the choice of $m_k$. It is exactly the prediction
scheme of the deterministic sequence prediction with known prior
SP$\Theta_\mu$ described in the last section. As this model was
optimal, AI$\mu$ is optimal, too, i.e. has minimal number of
expected errors (maximal expected credit) as compared to any other
sequence prediction scheme.

From this, it is already clear that the total expected credit
$C_{km}$ must be related to the expected sequence prediction error
$E_{m\Theta_\mu}$ (\ref{esp}). Let us prove directly that
$C_{1m}(\epsilon)\!+\!E_{m\Theta_\mu\!}=m$.
We rewrite $C_{km}^\best$ in (\ref{eerecsp})
as a function of $z_i$ instead of $y_ic_i$ as it
is symmetric in $y_ic_i$. Further, we can pull $C_{km}^\best$ out of
the maximization, as it is independent of $y_k$ similar to
(\ref{ebestysp}). Renaming the bounded variables $y_k$ and $c_k$
we get
\beq\label{ebr2}
  C_{km}^\best(z_{<k}) \;=\;
  \max_{z_k}\mu(z_{<k}\pb z_k) +
  \sum_{z_k}C_{k+1,m}^\best(z_{1:k})
  \!\cdot\!\mu(z_{<k}\pb z_k)
\eeq
Recursively inserting the l.h.s.\ into the r.h.s.\ we get
\beq\label{ebi2}
  C_{km}^\best(z_{<k}) \;=\;
  \sum_{i=k}^m\nq\;\sum_{\quad z_{k:i-1}}\nq\max_{z_i}
  \mu(z_{<k}\pb{z_{k:i}})
\eeq
This is most easily proven by induction. For $k\!=\!m$
we have $C_{mm}^\best(z_{<m})\!=\!\max_{z_m}\mu(z_{<m}\pb
z_m)$ from (\ref{ebr2}) and (\ref{ee0}), which equals (\ref{ebi2}). By induction
hypothesis, we assume that
(\ref{ebi2}) is true for $k$. Inserting this into
(\ref{ebr2}) we get
$$
  C_{km}^\best(z_{<k})
  \;=\;
  \max_{z_k}\mu(z_{<k}\pb z_k) +
  \sum_{z_k}\left[
  \sum_{i=k+1}^m\nq\;\sum_{\quad z_{k+1:i-1}}\max_{z_i}
  \mu(z_{1:k}\pb z_{k+1:i})
  \right]\mu(z_{<k}\pb z_k) \;=\;
$$
$$
  \;=\; \max_{z_k}\mu(z_{<k}\pb z_k) +
  \sum_{i=k+1}^m\nq\;\sum_{\quad z_{k:i-1}}\max_{z_i}
  \mu(z_{<k}\pb z_{k:i})
$$
which equals (\ref{ebi2}). This was the induction step and hence
(\ref{ebi2}) is proven.

By setting $k\!=\!0$ and slightly reformulating (\ref{ebi2}),
we get the total expected credit in the first $m$ cycles
$$
  C_{1:m}^\best(\epsilon) \;=\;
  \sum_{i=1}^m\;\sum_{z_{<i}}\mu(\pb z_{<i})
  \max\{\mu(z_{<i}\pb 0),\mu(z_{<i}\pb 1)\} \;=\;
  m-E_{m\Theta_\mu}
$$
with $E_{m\Theta_\mu}$ defined in (\ref{esp}).

\paragraph{Using the AI$\xi$ Model for Sequence Prediction:}
Now we want to use the universal AI$\xi$ model instead of
AI$\mu$ for sequence prediction and try to derive error bounds
analog to (\ref{spebound}).
Like in the AI$\mu$ case, the systems output $y_k$ in cycle $k$ is
interpreted as a prediction for the k$^{th}$ bit $z_k$ of the
string, which has to be predicted. The credit is
$c_k=\delta_{y_kz_k}$ and there are no other inputs
$x_k=\epsilon$. What makes the analysis more difficult is that $\xi$ is not
symmetric in $y_ic_i\leftrightarrow(1-y_i)(1-c_i)$ and
(\ref{muaisp}) does not hold for $\xi$. On the other hand,
$\xi^{AI}$ converges to $\mu^{AI}$ in the limit (\ref{aixitomu}), and
(\ref{muaisp}) should hold asymptotically for $\xi$ in some sense.
So we expect that everything proven for AI$\mu$ holds
approximately for AI$\xi$. The AI$\xi$ model should behave
similarly to SP$\Theta_\xi$, the deterministic variant of Solomonoff prediction.
Especially we expect error bounds similar to (\ref{spebound}). Making
this rigorous seems difficult. Some general remarks have been made
in the last section.

Here we concentrate on the special case of a deterministic
computable environment, i.e. the environment is a sequence
$\hh z\!=\!\hh z_1\hh z_2...$, $K(\hh z_1...\hh z_n*)\!\leq\!K(\hh
z)\!<\!\infty$. Furthermore, we only consider the simplest
horizon model $m_k\!=\!k$, i.e. maximize only the next
credit. This is sufficient for sequence prediction, as the credit
of cycle $k$ only depends on output $y_k$ and not on earlier
decisions. This choice is in no way sufficient and satisfactory
for the full AI$\xi$ model, as {\it one} single choice of $m_k$ should
serve for {\it all} AI problem classes. So AI$\xi$ should allow
good sequence prediction for some universal choice of $m_k$ and not
only for $m_k\!=\!k$, which definitely does not suffice for more
complicated AI problems. The analysis of this general case is a challenge for the future.
For $m_k\!=\!k$ the AI$\xi$ model
(\ref{ydotxi}) with $x'_i\!=\!\epsilon$ reduces to
\beq\label{ydotxisp}
  \hh y_k \;=\; \maxarg_{y_k}\sum_{c_k}c_k\!\cdot\!
  \xi(\hh y\!\hh c_{<k}y\!\pb c_k) \;=\;
  \maxarg_{y_k}\xi(\hh y\!\hh c_{<k}y_k\pb 1) \;=\;
  \maxarg_{y_k}\xi(\hh y\!\hh{\pb c}_{<k}y_k\pb 1)
\eeq
The environmental response $\hh c_k$ is given by $\delta_{\hh y_k\hh
z_k}$; it is 1 for a correct prediction $(\hh y_k\!=\!\hh z_k)$ and 0
otherwise. In the following, we want to bound the number of errors
this prediction scheme makes. We need the following inequality
\beq\label{spineq}
  \xi(y\!\pb c_1...y\!\pb c_k) \;>\;
  2^{-K(\delta_{y_1c_1}...\delta_{y_kc_k}*)-O(1)}
\eeq
We have to find a short program in the sum
(\ref{uniMAI}) calculating $c_1...c_k$ from $y_1...y_k$. If we
knew $z_i:=\delta_{y_ic_i}$ for $1\!\leq\!i\!\leq\!k$ a program of
size $O(1)$ could calculate
$c_1...c_k=\delta_{y_1z_1}...\delta_{y_kz_k}$. So combining this program with
a shortest coding of $z_1...z_k$ leads to a program of size
$K(z_1...z_k*)\!+\!O(1)$, which proves (\ref{spineq}).

Let us now assume that we make a wrong prediction in cycle $k$,
i.e. $\hh c_k\!=\!0$, $\hh y_k\neq \hh z_k$. The goal is to
show that $\hh\xi$ defined by
\beqn
  \hh\xi_k \;:=\; \xi(\hh y\!\pb{\hh c}_{1:k}) \;=\;
  \xi(\hh y\pb{\hh c}_{<k}\hh y_k\pb 0) \;\leq\;
  \xi(\hh y\pb{\hh c}_{<k}) -
  \xi(\hh y\pb{\hh c}_{<k}\hh y_k\pb 1) \;<\;
  \hh\xi_{k-1}-\alpha
\eeqn
decreases for every wrong prediction, at least by some $\alpha$.
The $\leq$ arises from the fact that $\xi$ is only a semimeasure.
\beqn
  \xi(\hh y\!\pb{\hh c}_1...\hh y\pb 1) \;>\;
  \xi(\hh y_1\pb{\hh c}_1...(1\!-\!\hh y_k)\pb 1) \;\stackrel{\times}{>}\;
  2^{-K(\delta_{\hh y_1\hh c_1}...\delta_{(1-\hh y_k)1}*)}
  \;=\;
\eeqn
\beqn
  \;=\; 2^{-K(\hh z_1...\hh z_k*)} \;>\;
  2^{-K(\hh z)-O(1)} \;=:\; \alpha
\eeqn
In the first inequality we have used the fact that $\hh y_k$
maximizes by definition (\ref{ydotxisp}) the argument, i.e.
$1\!-\!\hh y_k$ has lower probability than $\hh y_k$. (\ref{spineq}) has been
applied in the second inequality. The equality holds, because
$\hh z_i\!=\!\delta_{\hh y_i\hh c_i}$ and
$\delta_{(1-\hh y_k)1}\!=\!\delta_{\hh y_k0}\!=\!\delta_{\hh y_k\hh
c_k}\!=\!\hh z_k$. The last inequality follows from the
definition of $\hh z$.

We have shown that each erroneous prediction reduces $\hh\xi$ by at
least the $\alpha$ defined above. Together with $\hh\xi_0\!=\!1$ and
$\hh\xi_k\!>\!0$ for all $k$ this shows that the system can make
at most $1/\alpha$ errors, since otherwise $\hh\xi_k$ would become
negative. So the number of wrong predictions $E_{n\xi}^{AI}$ of system
(\ref{ydotxisp}) is bounded by
\beq\label{Ebndsp}
  E_{n\xi}^{AI} \;<\; {\textstyle{1\over\alpha}} \;=\;
  2^{K(\hh z)+O(1)} \;<\; \infty
\eeq
for a computable deterministic environment string $\hh z_1\hh
z_2...$. The intuitive interpretation is that each wrong
prediction eliminates at least one program $p$ of size
$l(p)\!\stackrel+<\!K(\hh z)$. The size is smaller than $K(\hh z)$, as
larger policies could not mislead the system to a wrong
prediction, since there is a program of size $K(\hh z)$ making a correct
prediction. There are at most $2^{K(\hh z)+O(1)}$ such policies,
which bounds the total number of errors.

We have derived a finite bound for $E_{n\xi}^{AI}$, but unfortunately, a
rather weak one as compared to (\ref{spebound}). The reason for the
strong bound in the SP case was that every error at least halves
$\hh\xi$ because the sum of the $\maxarg_{x_k}$ arguments was 1.
Here we have
\bqan
  \xi(\hh y_1\hh c_1...\hh y_{k-1}\hh c_{k-1}0\pb 0) +
  \xi(\hh y_1\hh c_1...\hh y_{k-1}\hh c_{k-1}0\pb 1) = 1 \\
  \xi(\hh y_1\hh c_1...\hh y_{k-1}\hh c_{k-1}1\pb 0) +
  \xi(\hh y_1\hh c_1...\hh y_{k-1}\hh c_{k-1}1\pb 1) = 1
\eqan
but $\maxarg_{y_k}$ runs over the right top and right bottom
$\xi$, for which no sum criterion holds.

The AI$\xi$ model would not be sufficient for
realistic applications if the bound (\ref{Ebndsp}) were sharp,
but we have the strong feeling (but only weak
arguments) that better bounds proportional to $K(\hh z)$
analog to (\ref{spebound}) exist. The technique used above may not
be appropriate for achieving this. One argument for a better bound is
the formal similarity between $\maxarg_{z_k}(\hh z_{<k}z_k)$ and (\ref{ydotxisp}),
the other is that we were unable to construct an example sequence
for which (\ref{ydotxisp}) makes more than $O(K(\hh z))$ errors.

\newpage
\section{Strategic Games (SG)}\label{secSG}

\paragraph{Introduction:}
A very important class of problems are strategic games, like chess.
In fact, what is subsumed under game theory nowadays, is so
general, that it includes not only a huge variety of games, from simple
games of chance like roulette, combined with strategy like
Backgammon, up to purely strategic games like chess or checkers or
go. Game theory can also describe political and economic competitions and
coalitions, even Darwinism and many more have been modeled within game theory.
It seems that nearly every AI problem could be brought into
the form of a game. Nevertheless,
the intention of a game is that several players perform
some actions with (partial) observable consequences.
The goal of each player is to maximize some utility
function (e.g.\ to win the game). The players are assumed to be
rational, taking into account all information they posses. The
different goals of the players are usually in conflict.
For an introduction into game theory, see \cite{Fud91,Osb94,Rus95,Neu44}.

If we interpret the AI system as one player and the environment
models the other rational player {\it and} the environment provides
the reinforcement feedback $c_k$, we see that the system-environment
configuration satisfies all criteria of a game. On the other hand,
we know that the AI system can handle more general situations,
since it interacts optimally with an environment, even if the environment
is not a rational player with conflicting goals.

\paragraph{Strictly competitive strategic games:}
In the following, we restrict ourselves to deterministic, strictly
competitive strategic\footnote{In game theory, games like chess
are often called 'extensive', whereas 'strategic' is reserved for a
different kind of game.} games with alternating moves. Player 1
makes move $y_k'$ in round $k$, followed by the move $x_k'$ of player
2. So a game with $n$ rounds consists of a sequence of alternating
moves $y'_1x'_1y'_2x'_2...y'_nx'_n$. At the end of the game in cycle $n$
the game or final board state is evaluated with
$C(y'_1x'_1...y'_nx'_n)$. Player 1 tries to maximize $C$, whereas player 2
tries to minimize $C$. In the simplest case, $C$ is $1$ if player 1
won the game, $C\!=\!-1$ if player 2 won and $C\!=\!0$ for a draw. We
assume a fixed game length $n$ independent of the actual move
sequence. For games with variable length but maximal possible number of
moves $n$, we could add dummy moves
and pad the length to $n$. The optimal strategy (Nash equilibrium)
of both players is a minimax strategy
\beq\label{sgxdot}
  \hh x'_k=\minarg_{x'_k}\max_{y'_{k+1}}\min_{x'_{k+1}}...\max_{y'_n}\min_{x'_n}
  C(\hh y'_1\hh x'_1...\hh y'_kx'_k...y'_nx'_n)
\eeq
\beq\label{sgydot}
  \hh y'_k=\maxarg_{y'_k}\min_{x'_k}...\max_{y'_n}\min_{x'_n}
  C(\hh y'_1\hh x'_1...\hh y'_{k-1}\hh x'_{k-1}y'_kx'_k...y'_nx'_n)
\eeq
But note, that the minimax strategy is only optimal if both players
behave rationally. If, for instance, player 2 has limited capabilites or makes
errors and player 1 is able to discover these (through past moves) he
could exploit these and improve his performance
by deviating from the minimax strategy. At least, the classical
game theory of Nash equilibria does not take into account limited
rationality, whereas the AI$\xi$ system should.

\paragraph{Using the AI$\mu$ model for game playing:}
In the following, we demonstrate the applicability of the AI model
to games. The AI system takes the position of player 1. The
environment provides the evaluation $C$. For a symmetric situation
we could take a second AI system as player 2, but for simplicity we
take the environment as the second player and assume that this
environmental player behaves according to the minimax strategy (\ref{sgxdot}).
The environment serves as a perfect player {\it and} as a teacher, albeit a
very crude one as it tells the system at the end of the game,
only whether it won or lost.

The minimax behaviour of player 2 can be expressed by a
(deterministic) probability distribution $\mu^{SG}$ as the
following
\beq\label{defmusg}
  \mu^{SG}(y'_1\pb x'_1...y'_n\pb x'_n) \;:=\;
  \left\{
  \begin{array}{l}
    \displaystyle
    1 \quad\mbox{if}\quad
    x'_k=\minarg_{x''_k}...\max_{y''_n}\min_{x''_n}
    C(y'_1...x'_{k-1}y''_k...x''_n)
    \;\;\forall\; 1\!\leq\!k\!\leq\!n
    \\
    0 \quad\mbox{otherwise}
  \end{array} \right.
\eeq
The probability that player 2 makes move $x'_k$ is
$\mu^{SG}(\hh y'_1\!\hh x'_1...\hh y'_k\pb x'_k)$ which is 1 for
$x'_k\!=\!\hh x'_k$ as defined in (\ref{sgxdot}) and 0 otherwise.

Clearly, the AI system receives no feedback, i.e.
$c_1\!=...=\!c_{n-1}\!=\!0$, until the end of the game, where it should
receive positive/negative/neutral feedback on a win/loss/draw, i.e.
$c_n=C(...)$. The environmental prior probability is therefore
\beq\label{muaisg}
  \mu^{AI}(y_1\pb x_1...y_n\pb x_n) \;=\;
  \left\{
  \begin{array}{cl}
    \displaystyle
    \mu^{SG}(y'_1\pb x'_1...y'_n\pb x'_n) & \mbox{if}\quad
    c_1\!=...=\!c_{n-1}\!=\!0 \;\mbox{and}\; c_n=C(y'_1x'_1...y'_nx'_n)
    \\
    0 & \mbox{otherwise}
  \end{array} \right.
\eeq
where $y_i\!=\!y'_i$ and $x_i\!=\!c_ix'_i$.
If the environment is a minimax player (\ref{sgxdot}) plus a crude
teacher $C$, i.e. if $\mu^{AI}$ is the true prior probability, the
question now is, what is the behaviour $\hh y_k^{AI}$ of the AI$\mu$
system. It turns out that if we set $m_k\!=\!n$ the AI$\mu$ system
is also a minimax player (\ref{sgydot}) and hence optimal
\beqn
  \hh y_k^{AI} \;=\;
  \maxarg_{y_k}\sum_{x'_k}...\max_{y_n}\sum_{x'_n}
  C(\hh y\!\hh x'_{<k}y\!x'_{k:n})\!\cdot\!
  \mu^{SG}(\hh y\!\hh x'_{<k}y\!\pb x'_{k:n}) \;=
\eeqn
\beq\label{yaisg2}
  =\; \maxarg_{y_k}\sum_{x'_k}...\max_{y_{n-1}}\sum_{x'_{n-1}}\max_{y_n}\min_{x'_n}
  C(\hh y\!\hh x'_{<k}y\!x'_{k:n})\!\cdot\!
  \mu^{SG}(\hh y\!\hh x'_{<k}y\!\pb x'_{k:n-1}) \;=
\eeq
\beqn
 =\;...\;=\; \maxarg_{y_k}\min_{x'_{k+1}}...\max_{y_n}\min_{x'_n}
     C(\hh y\!\hh x'_{<k}y\!x'_{k:n}) \;=\;
     \hh y_k^{SG}
\eeqn
In the first line we inserted $m_k\!=\!n$ and (\ref{muaisg}) into
the definition (\ref{ydotrec}) of $\hh y_k^{AI}$. This removes all
sums over the $c_k$. Further, the sum over $x'_n$ gives only a
contribution for $x'_n\!=\!\minarg_{x'_n}C(\hh x'_1\hh
y'_1...x'_ny'_n)$ by definition (\ref{defmusg}) of $\mu^{SG}$.
Inserting this $x'_n$ gives the second line. $\mu^{SG}$ is
effectively reduced to a lower number of arguments and the sum
over $x'_n$ replaced by $\min_{x'_n}$.  Repeating this procedure
for $x'_{n-1},...,x'_{k+1}$ leads to the last line, which is just
the minimax strategy of player 1 defined in (\ref{sgydot}).

Let us now assume that the game under consideration is played $s$
times. The prior probability then is
\beq\label{sgrep}
  \mu^{AI}(y\!\pb x_1...y\!\pb x_{sn}) \;=\;
  \prod_{r=0}^{s-1} \mu_1^{AI}(y\!\pb x_{rn+1}...
  y\!\pb x_{(r+1)n})
\eeq
where we have renamed the prior probability (\ref{muaisg}) for
one game to $\mu_1^{AI}$. (\ref{sgrep}) is a special case of a
factorizable $\mu$ (\ref{facmu}) with identical factors
$\mu_r\!=\mu_1^{AI}$ for all $r$ and equal episode lengths
$n_{r+1}\!-\!n_r\!=\!n$. The AI$\mu$ system (\ref{sgrep}) for repeated
game playing also implements the minimax strategy,
\beq\label{yaisgrep}
  \hh y_k^{AI} \;=\;
  \maxarg_{y_k}\min_{x'_k}...
     \max_{y_{(r+1)n}}\min_{\;x'_{(r+1)n}}
     C(\hh y\!\hh x'_{rn+1:k-1}...y\!x'_{k:(r+1)n})
\eeq
with $r$ such that $rn\!<\!k\!\leq\!(r\!+\!1)n$ and for any choice of $m_k$
as long as the horizon $h_k\!\geq\!n$. This can be
proved by using (\ref{facydot}) and (\ref{yaisg2}).
See section (\ref{secAIxi}) for a discussion on separable and
factorizable $\mu$.

\paragraph{Games of variable length:}
In the unrepeated case we have argued that games of variable but
bounded length can be padded to a fixed length without effect. We
now analyze in a sequence of games the effect of replacing the games with fixed
length by games of variable length.
The sequence $y'_1x'_1...y'_nx'_n$ can still be grouped into episodes
corresponding to the moves of separated consecutive games, but now
the length and total number of games that fit into the $n$
moves depend on the actual moves taken\footnote{If the sum of
game lengths do not fit exactly into $n$ moves, we pad the last
game appropriately.}. $C(y'_1x'_1...y'_nx'_n)$
equals the number of games where the
system wins, minus the number of games where the environment wins.
Whenever a loss, win or draw has been achieved by the
system or the environment, a new game starts. The player whose turn it would next
be, begins the next game. The games are still separated in
the sense that the behaviour and credit of the current game does
not influence the next game. On the other hand, they are
slightly entangled, because the length of the current
game determines the time of start of the next. As the rules of the
game are time invariant, this does not influence the next game
directly. If we play a fixed number of games, the games are
completely independent, but if we play a fixed number of total moves
$n$, the number of games depends on their lengths. This has the
following consequences: the better player tries to keep the games
short, to win more games in the given time $n$. The poorer player
tries to draw the games out, in order loose less games. The better
player might further prefer a quick draw, rather than to win a long game.
Formally, this entanglement is represented by the fact that the
prior probability $\mu$ does no longer factorize. The reduced
form (\ref{yaisgrep}) of $\hh y_k^{AI}$ to one episode is no
longer valid. Also, the behaviour $\hh y_k^{AI}$ of the system
depends on $m_k$, even if the horizon $h_k$ is
chosen larger than the longest possible game (unless $m_k\!\geq\!n$).
The important point is that the system realizes that
keeping games short/long can lead to increased credit. In
practice, a horizon much larger than the average game length
should be sufficient to incorporate this effect. The details of
games in the distant future do not affect the current game and can,
therefore, be ignored. A more quantitative analysis could be interesting, but
would lead us too far astray.

\paragraph{Using the AI$\xi$ model for game playing:}
When going from the specific AI$\mu$ model, where the rules of the
game have been explicitly modeled into the prior probability
$\mu^{AI}$, to the universal model AI$\xi$ we have to ask whether
these rules can be learned from the assigned credits $c_k$. Here,
another (actually the main) reason for studying the case of
repeated games, rather than just one game arises. For a single game
there is only one cycle of non-trivial feedback namely the end of
the game - too late to be useful except when there are further
games following.

Even in the case of repeated games, there is only very limited
feedback, at most $\log_2 3$ bits of information per game if the 3
outcomes win/loss/draw have the same frequency. So there are at
least $O(K(game))$ number of games necessary to learn a game of
complexity $K(game)$. Apart from extremely simple games, even this
estimate is far too optimistic. As the AI$\xi$ system has no
information about the game to begin with, its moves will be more
or less random and it can win the first few games merely by pure luck.
So the probability that the system looses is near to one and
hence the information content $I$ in the feedback $c_k$ at the end
of the game is much less than $\log_2 3$. This situation remains
for a very large number of games. On the other hand, in principle,
every game should be learnable after a very long sequence of games
even with this minimal feedback only, as long as $I\not\equiv 0$.

The important point is that no other learning scheme with no extra
information can learn the game more quickly. We expect this to be
true as $\mu^{AI}$ factorizes in the case of games of fixed
length, i.e. $\mu^{AI}$ satisfies a strong separability condition.
In the case of variable game length the entanglement is also low.
$\mu^{AI}$ should still be sufficiently separable allowing
to formulate and prove good credit bounds for AI$\xi$.

To learn realistic games like tic-tac-toe (noughts and crosses) in
realistic time one has to provide more feedback. This could be
achieved by intermediate help during the game. The environment
could give positive(negative) feedback for every good(bad) move
the system makes. The demand on whether a move is to be valued as
good should be adopted to the gained experience of the system in
such a way that approximately half of the moves are valuated as
good and the other half as bad, in order to maximize the
information content of the feedback.

For more complicated games like chess, even more feedback is
necessary from a practical point of view. One way to increase the
feedback far beyond a few bits per cycle is to train the system by
teaching it good moves. This is called supervised learning.
Despite the fact that the AI model has only a credit feedback
$c_k$, it is able to learn by teaching, as will be shown in section
\ref{secEX}. Another way would be to start with more simple games
containing certain aspects of the true game and to switch to the true
game when the system has learned the simple game.

No other difficulties are expected when going from
$\mu$ to $\xi$. Eventually $\xi^{AI}$ will converge to the
minimax strategy $\mu^{AI}$. In the more realistic case, where the
environment is not a perfect minimax player, AI$\xi$ can
detect and exploit the weakness of the opponent.

Finally, we want to comment on the input/output space $X$/$Y$ of
the AI system. In practical applications, $Y$ will possibly include
also illegal moves. If $Y$ is the set of moves of e.g. a robotic
arm, the system could move a wrong figure or even knock over the
figures. A simple way to handle illegal moves $y_k$ is by
interpreting them as losing moves, which terminate the game.
Further, if e.g. the input $x_k$ is the image of a video camera
which makes one shot per move, $X$ is not the set of moves by the
environment but includes the set of states of the game board. The
discussion in this section handles this case as well. There is no
need to explicitly design the systems I/O space $X/Y$ for a
specific game.

The discussion above on the AI$\xi$ system was rather informal for
the following reason: game playing (the SG$\xi$ system) has
(nearly) the same complexity as fully general AI, and quantitative
results for the AI$\xi$ system are difficult (but not impossible)
to obtain.

\newpage
\section{Function Minimization (FM)}\label{secFM}

\paragraph{Applications/Examples:}
There are many problems that can be reduced to a minimization
problem (FM). The minimum of a (real valued) function
$f\!:\!Y\!\to\!I\!\!R$ over some domain $Y$ or a good approximate
of it has to be found, usually with some limited resources.

One popular example is the traveling salesman problem (TSP). $Y$
is the set of different routes between towns and $f(y)$ the length
of route $y\!\in\!Y$. The task is to find a route of minimal
length visiting all cities. This problem is NP hard. Getting good
approximations in limited time is of great importance in various
applications. 
Another example is the minimization of production costs (MPC),
e.g.\ of a car, under several constraints. $Y$ is the set of all
alternative car designs and production methods compatible with the
specifications and $f(y)$ the overall cost of alternative
$y\!\in\!Y$. 
A related example is finding materials or (bio)molecules with
certain properties (MAT). E.g. solids with minimal electrical
resistance or maximally efficient chlorophyll modifications or
aromatic molecules that taste as close as possible to strawberry.
We can also ask for nice paintings (NPT). $Y$ is the set of all
existing or imaginable paintings and $f(y)$ characterizes how much
person $A$ likes painting $y$. The system should present
paintings, which $A$ likes.

For now, these are enough examples. The TSP is very rigorous from a
mathematical point of view, as $f$, i.e. an algorithm of $f$, is
usually known. In principle, the minimum could be found by
extensive search, were it not for computational resource
limitations. For MPC, $f$ can often be modeled in a reliable and
sufficiently accurate way. For MAT you need very accurate physical
models, which might be unavailable or too difficult to solve or
implement. For NPT the most we have is the judgement of person $A$ on
every presented painting. The evaluation function $f$ cannot be
implemented without scanning $A's$ brain, which is not possible with
todays technology.

So there are different limitations, some depending on the
application we have in mind. An implementation of $f$ might not be
available, $f$ can only be tested at some arguments $y$ and $f(y)$
is determined by the environment. We want to (approximately)
minimize $f$ with as few function calls as possible or, conversely,
find an as close as possible approximation for the
minimum within a fixed number of function evaluations. If $f$ is
available or can quickly be inferred by the system and evaluation
is quick, it is more important to minimize the total time needed to
imagine new trial minimum candidates plus the evaluation time for
$f$. As we do not consider computational aspects of AI$\xi$ till
section \ref{secTime} we concentrate on the first
case, where $f$ is not available or dominates the computational
requirements.

\paragraph{The Greedy Model FMG$\mu$ :}
The FM model consists of a sequence $\hh y_1\hh z_1\hh y_2\hh
z_2...$ where $\hh y_k$ is a trial of the FM system for a minimum
of $f$ and $\hh z_k=f(\hh y_k)$ is the true function value
returned by the environment. We randomize the model by assuming a
probability distribution $\mu(f)$ over the functions. There are
several reasons for doing this. We might really not know the exact
function $f$, as in the NPT example, and model our uncertainty by
the probability distribution $\mu$. More importantly, we want to
parallel the other AI classes, like in the SP$\mu$ model, where we
always started with a probability distribution $\mu$ that was finally
replaced by $\xi$ to get the universal Solomonoff prediction
SP$\xi$. We want to do the same thing here. Further, the probabilistic
case includes the deterministic case by choosing
$\mu(f)\!=\!\delta_{ff_0}$, where $f_0$ is the true function. A
final reason is that the deterministic case is trivial when $\mu$
and hence $f_0$ is known, as the system can internally (virtually)
check all function arguments and output the correct minimum from the very
beginning.

We will assume that $Y$ is countable or finite and that $\mu$ is a
discrete measure, e.g. by taking only computable functions. The
probability that the function values of $y_1,...,y_n$ are
$z_1,...,z_n$ is then given by
\beq\label{fmmudef}
  \mu^{FM}(y_1\pb z_1...y_n\pb z_n) \;:=\;
  \sum_{f:f(y_i)=z_i\;\forall 1\leq i\leq n} \nq\mu(f)
\eeq
We start with a model that minimizes the expectation
$z_k$ of the function value $f$ for the next output
$y_k$, taking into account previous information:
\beqn
  \hh y_k \;:=\; \minarg_{y_k}\sum_{z_k} z_k\!\cdot\!
  \mu(\hh y_1\hh z_1...\hh y_{k-1}\hh z_{k-1}y_k\pb z_k)
\eeqn
This type of greedy algorithm, just minimizing the next
feedback, was sufficient for sequence prediction (SP) and is also
sufficient for classification (CF). It is, however, not sufficient for
function minimization as the following example demonstrates.

Take $f:\{0,1\}\!\to\!\{1,2,3,4\}$. There are 16 different
functions which shall be equiprobable, $\mu(f)\!=\!{1\over 16}$.
The function expectation in the first cycle
\beqn
  \langle z_1\rangle \;:=\; \sum_{z_1} z_1\!\cdot\!\mu(y_1\pb z_1) \;=\;
  {\textstyle{1\over 4}}\sum_{z_1}z_1 \;=\;
  {\textstyle{1\over 4}}(1\!+\!2\!+\!3\!+\!4) \;=\; 2.5
\eeqn
is just the arithmetic average of the possible function values and
is independent of $y_1$. Therefore, $\hh y_1\!=\!0$, as $\minarg$
is defined to take the lexicographically first minimum in an
ambiguous case. Let us assume that $f_0(0)\!=\!2$, where $f_0$ is the
true environment function, i.e. $\hh z_1\!=\!2$. The expectation of $z_2$ is then
\beqn
  \langle z_2\rangle \;:=\; \sum_{z_2} z_2\!\cdot\!\mu(02y_2\pb z_2)
  \;=\; \left\{
  \begin{array}{c@{\quad\mbox{for}\quad}l}
    2                      & y_2=0 \\
    2.5                    & y_2=1
  \end{array} \right.
\eeqn
For $y_2\!=\!0$ the system already knows $f(0)\!=\!2$, for
$y_2\!=\!1$ the expectation is, again, the arithmetic average. The
system will again output $\hh y_2\!=\!0$ with feedback $\hh
z_2\!=\!2$. This will continue forever. The system is not
motivated to explore other $y's$ as $f(0)$ is already smaller than the
expectation of $f(1)$. This is obviously not what we
want. The greedy model fails. The system ought to be inventive and
try other outputs when given enough time.

The general reason for the failure of the greedy approach is that
the information contained in the feedback $z_k$ depends on the
output $y_k$. A FM system can actively influence the knowledge it
receives from the environment by the choice in $y_k$. It may be
more advantageous to first collect certain knowledge about $f$ by
an (in greedy sense) non-optimal choice for $y_k$, rather than to
minimize the $z_k$ expectation immediately. The non-minimality of
$z_k$ might be over-compensated in the long run by
exploiting this knowledge. In SP, the received information is
always the current bit of the sequence, independent of what SP
predicts for this bit. This is the reason why a greedy
strategy in the SP case is already optimal.

\paragraph{The general FM$\mu/\xi$ Model:}
To get a useful model we have to think more carefully about what we
really want. Should the FM system output a good minimum in the last output
in a limited number of
cycles $T$, or should the average of the $z_1,...,z_T$ values be minimal, or
does it suffice that just one of the $z$ is as small as possible?
Let us define the FM$\mu$ model as to minimize the $\mu$ averaged weighted
sum $\alpha_1 z_1\!+...+\!\alpha_T z_T$ for some given
$\alpha_k\!\geq\!0$. Building the $\mu$ average by summation over
the $z_i$ and minimizing w.r.t.\ the $y_i$ has to be performed in
the correct chronological order. With a similar reasoning as in
(\ref{ebesty}) to (\ref{ydotrec}) we get
\beq\label{fmydot}
  \hh y_k^{FM} \;=\; \minarg_{y_k}\sum_{z_k}...\min_{y_T}\sum_{z_T}
  (\alpha_1 z_1\!+...+\!\alpha_T z_T)\!\cdot\!
  \mu(\hh y_1\hh z_1...\hh y_{k-1}\hh z_{k-1}y_k\pb z_k...y_T\pb z_T)
\eeq
If we want the final output $\hh y_T$ to be optimal we should
choose $\alpha_k\!=\!0$ for $k\!<\!T$ and $\alpha_T\!=\!1$ (final
model FMF$\mu$). If we want to already have a good
approximation during intermediate cycles, we should demand that the
output of all cycles together are optimal in some average sense,
so we should choose $\alpha_k\!=\!1$ for all $k$ (sum model
FMS$\mu$). If we want to have something in between, for instance, increase
the pressure to produce good outputs, we could choose the
$\alpha_k\!=\!e^{\gamma(k-T)}$ exponentially increasing for some
$\gamma\!>\!0$ (exponential model FME$\mu$). For
$\gamma\!\to\!\infty$ we get the FMF$\mu$, for $\gamma\!\to\!0$
the FMS$\mu$ model. If we want to demand that the best of the
outputs $y_1...y_k$ is optimal, we must replace the $\alpha$
weighted $z$-sum by $\min\{z_1,...,z_T\}$ (minimum Model
FMM$\mu$). We expect the behaviour to be very similar to the
FMF$\mu$ model, and do not consider it further.

By construction, the FM$\mu$ models guarantee optimal results in
the usual sense that no other model knowing only $\mu$
can be expected to produce better results. The variety of FM
variants is not a fault of the theory. They just reflect the fact
that there is some interpretational freedom of what is meant by
minimization within $T$ function calls. In most applications, probably FMF is
appropriate. In the NPT application one might prefer the FMS model.

The interesting case (in AI) is when $\mu$ is unknown. We
define for this case, the FM$\xi$ model by replacing $\mu(f)$
with some $\xi(f)$, which should assign high probability to
functions $f$ of low complexity. So we might define\footnote
{$\xi^{FM}(f)$ is a true
probability distribution if we include partial functions in the
domain. So normalization is not necessary.}
$\xi(f)\!=\!\sum_{q:\forall x[U(qx)=f(x)]}2^{-l(q)}$.
The problem with this definition is that it is, in general,
undecidable whether a TM $q$ is an implementation of a function
$f$. $\xi(f)$ defined in this way is uncomputable,
not even approximable. As we only need a $\xi$ analog to the
l.h.s.\ of (\ref{fmmudef}), the following definition is natural
\beq\label{fmxidef}
  \xi^{FM}(y_1\pb z_1...y_n\pb z_n) \;:=\;
  \sum_{q:q(y_i)=z_i\;\forall 1\leq i\leq n} \nq 2^{-l(q)}
\eeq
$\xi^{FM}$ is
actually equivalent to inserting the incomputable $\xi(f)$ into
(\ref{fmmudef}). $\xi^{FM}$ is an enumerable semi-measure and
universal, relative to all probability distributions of the form
(\ref{fmmudef}). We will not prove this here.

Alternatively, we could have constrained the sum in (\ref{fmxidef})
by $q(y_1...y_n)\!=\!z_1...z_n$ analog to (\ref{uniMAI}), but these
two definitions are not equivalent. Definition (\ref{fmxidef})
ensures the symmetry\footnote{See \cite{Sol99} for a discussion
on symmetric universal distributions on unordered data.} in its
arguments and $\xi^{FM}(...y\pb z...y\pb z'...)\!=\!0$ for $z\neq z'$.
It incorporates all general knowledge we have about function
minimization, whereas (\ref{uniMAI}) does not. But this extra
knowledge has only low information content (complexity of $O(1)$),
so we do not expect FM$\xi$ to perform much worse when using
(\ref{uniMAI}) instead of (\ref{fmxidef}). But there is no reason
to deviate from (\ref{fmxidef}) at this point.

We can now define an ''error'' measure $E_{T\mu}^{FM}$ as
(\ref{fmydot}) with $k\!=\!1$ and $\minarg_{y_1}$ replaced by
$\min_{y_1}$ and, additionally, $\mu$ replaced by $\xi$ for
$E_{T\xi}^{FM}$. We expect $|E_{T\xi}^{FM}\!-\!E_{T\mu}^{FM}|$ to
be bounded in a way that justifies the use of $\xi$ instead of
$\mu$ for computable $\mu$, i.e. computable $f_0$ in the
deterministic case. The arguments are the same as for the AI$\xi$
model.

\paragraph{Is the general model inventive?}
In the following we will show that FM$\xi$ will never cease
searching for minima, but will test an infinite set of different
$y's$ for $T\!\to\!\infty$.

Let us assume that the system tests only a finite number of
$y_i\!\in\!A\!\subset Y$, $|A|\!<\!\infty$. Let $t\!-\!1$ be the
cycle in which the last new $y\!\in\!A$ is selected (or some later
cycle). Selecting $y's$ in cycles $k\!\geq\!t$ a second time, the
feedback $z$ does not provide any new information, i.e. does not
modify the probability $\xi^{FM}$. The system can
minimize $E_{T\xi}^{FM}$ by outputting in cycles $k\geq t$ the
best $y\!\in\!A$ found so far (in the case $\alpha_k\!=\!0$, the output
does not matter).
Let us fix $f$ for a moment. Then we have
\beqn
  E^a \;:=\; \alpha_1 z_1\!+...+\!\alpha_T z_T \;=\;
  \sum_{k=1}^{t-1}\alpha_kf(y_k)+f_1\!\cdot\!\sum_{k=t}^T\alpha_k
  \quad,\quad f_1:=\min_{1\leq k<t}f(y_k)
\eeqn
Let us now assume that the system tests one additional
$y_t\!\not\in\!A$ in cycle $t$, but no other $y\!\not\in\!A$.
Again, it will keep to the best output for $k\!>\!t$, which is
either the one of the previous system or $y_t$.
\beqn
  E^b \;=\;
  \sum_{k=1}^t\alpha_kf(y_k) +
  \min\{f_1,f(y_t)\}\!\cdot\nq\;\sum_{k=t+1}^T\alpha_k
\eeqn
The difference can be represented in the form
\beqn
  E^a-E^b \;=\; \left(\sum_{k=t}^T\alpha_k\right)\!\cdot\!f^+ -
  \alpha_t\!\cdot\!f^- \quad,\quad
  f^\pm \;:=\; \max\{0,\pm(f_1\!-\!f(y_t))\} \;\geq\; 0
\eeqn
As the true FM$\xi$ strategy is the one which minimizes $E$, assumption
$a$ is ruled out if $E^a>E^b$. We will say that $b$ is favored over $a$,
which does not mean that $b$ is the correct strategy, only that
$a$ is not the true one. For probability distributed $f$, $b$ is
favored over $a$ when
\beqn
  E^a-E^b \;=\; \left(\sum_{k=t}^T\alpha_k\right)\!\cdot\!\langle f^+\rangle -
  \alpha_t\!\cdot\!\langle f^-\rangle \;>\; 0
  \quad\Leftrightarrow\quad
  \sum_{k=t}^T\alpha_k > \alpha_t{\langle f^-\rangle\over\langle
  f^+\rangle}
\eeqn
where $\langle f^\pm\rangle$ is the $\xi$ expectation of $\pm f_1\mp f(y_t)$
under the condition that $\pm f_1\!\geq\!\pm f(y_t)$ and under the constrains
imposed in cycles $1...t\!-\!1$. As $\xi$ assigns a strictly
positive probability to every non-empty event, $\langle
f^+\rangle\!\neq\!0$.
Inserting $\alpha_k\!=\!e^{\gamma(k-T)}$, assumption $a$ is ruled
out in model FME$\xi$ if
\beqn
  T-t \;>\; {1\over\gamma}\ln\left[1+
  {\langle f^-\rangle\over\langle f^+\rangle}(e^\gamma-1)\right]-1
  \;\to\; \left\{
  \begin{array}{c@{\quad\mbox{for}\quad}l}
    0 & \gamma\to\infty\mbox{ (FMF$\xi$ model)} \\
    \langle f^-\rangle/\langle f^+\rangle-1
    & \gamma\to 0\;\;\mbox{ (FMS$\xi$ model)}
  \end{array} \right.
\eeqn
We see that if the condition is not satisfied for some $t$, it will
remain wrong for all $t'\!>\!t$. So the FMF$\xi$ system will test each $y$
only once up to a point from which on it always outputs the best
found $y$. Further, for $T\!\to\!\infty$ the condition always gets
satisfied. As this is true for any finite $A$, the assumption of a
finite $A$ is wrong. For $T\!\to\!\infty$ the system
tests an increasing number of different $y's$, provided $Y$ is
infinite. The FMF$\xi$ model will never repeat any $y$ except in
the last cycle $T$ where it chooses the best found $y$. The
FMS$\xi$ model will test a new $y_t$ for fixed $T$, only if the
expected value of $f(y_t)$ is not too large.

The above does not necessarily hold for different choices of
$\alpha_k$. The above also holds for the FMF$\mu$ system if
$\langle f^+\rangle\!\neq\!0$. $\langle f^+\rangle\!=\!0$ if the
system can already exclude that $y_t$ is a better guess, so there
is no reason to test it explicitly.

Nothing has been said about the quality of the guesses, but for
the FM$\mu$ system they are optimal by definition.
If $K(\mu)$ for the true distribution $\mu$ is finite, we expect
the FM$\xi$ system to solve the ''exploration versus
exploitation'' problem in a universally optimal way, as $\xi$
converges to $\mu$.

\paragraph{Using the AI models for Function Mininimization:}
The AI model can be used for function minimization in the
following way. The output $y_k$ of cycle $k$ is a guess for a
minimum of $f$, like in the FM model. The credit $c_k$ should
be high for small function values $z_k\!=\!f(y_k)$.
The credit should also be weighted with $\alpha_k$ to reflect the
same strategy as in the FM case. The choice of $c_k\!=\!-\alpha_k z_k$
is natural. Here, the feedback is not binary but
$c_k\!\in\!C\!\subset\!I\!\!R$, with $C$ being a countable subset of
$I\!\!R$, e.g. the computable reals or all rational numbers. The
feedback $x'_k$ should be the function value $f(y_k)$.
So we set $x'_k\!=\!z_k$. Note, that there is a redundancy
if $\alpha_{()}$ is a computable function with no zeros, as
$c_k\!=-\alpha_kx'_k$. So, for small $K(\alpha_{()})$ like in
the FMS model, one might set $x_k\equiv\epsilon$. If we keep $x'_k$
the AI prior probability is
\beq\label{muAIfm}
  \mu^{AI}(y_1\pb x_1...y_n\pb x_n)
  \;=\; \left\{
  \begin{array}{cl}
    \mu^{FM}(y_1\pb z_1...y_n\pb z_n)
    & \mbox{for } c_k=-\alpha_kz_k,\; x'_k=z_k,\; x_k=c_kx_k' \\
    0 & \mbox{else}.
  \end{array} \right.
\eeq
Inserting this into (\ref{ydotrec}) with $m_k=T$ we get
\beqn
  \hh y_k^{AI} \;=\;
  \maxarg_{y_k}\sum_{x_k}...\max_{y_T}\sum_{x_T}
  (c_k\!+...+\!c_T)\!\cdot\!
  \mu^{AI}(\hh y_1\hh x_1...y_k\pb x_k...y_T\pb x_T)
  \;=\;
\eeqn
\beqn
  \;=\; \minarg_{y_k}\sum_{z_k}...\min_{y_T}\sum_{z_T}
  (\alpha_kz_k\!+...+\!\alpha_Tz_T)\!\cdot\!
  \mu^{FM}(\hh y_1\hh z_1...y_k\pb z_k...y_T\pb z_T) \;=\; \hh y_k^{FM}
\eeqn
where $\hh y_k^{FM}$ has been defined in (\ref{fmydot}).
The proof of equivalence was so simple because the FM model has already a
rather general structure, which is similar to the full AI model.

One might expect no problems when going from the already very
general FM$\xi$ model to the universal AI$\xi$ model (with
$m_k=T$), but there is a pitfall in the case of the FMF model. All
credits $c_k$ are zero in this case, except for the last one being $c_T$.
Although there is a feedback $z_k$ in every cycle, the AI$\xi$
system cannot learn from this feedback as it is not told that in
the final cycle $c_T$ will equal to $-z_T$. There is no problem in
the FM$\xi$ model because in this case this knowledge is hardcoded into
$\xi^{FM}$. The AI$\xi$ model must first learn that it
has to minimize a function but it can only learn if there is a
non-trivial credit assignment $c_k$. FMF works for repeated
minimization of (different) functions, such as minimizing $N$
functions in $N\!\cdot\!T$ cycles. In this case there are $N$ non-trivial
feedbacks and AI$\xi$ has time to learn that there is a relation
between $c_{k\!\cdot\!T}$ and $x'_{k\!\cdot\!T}$ every T$^{th}$
cycle. This situation is similar to the case of strategic games
discussed in section \ref{secSG}.

There is no problem in applying AI$\xi$ to FMS because the $c$
feedback provides enough information in this case. The only thing
the AI$\xi$ model has to learn, is to ignore the $x$ feedbacks as
all information is already contained in $c$. Interestingly the
same argument holds for the FME model if $K(\gamma)$ and $K(T)$
are small\footnote{If we set $\alpha_k=e^{\gamma k}$ the condition
on $K(T)$ can be dropped.}. The AI$\xi$ model has additionally only to learn
the relation $c_k\!=\!-e^{-\gamma(k-T)}x'_k$. This
task is simple as every cycle provides one data point for a simple
function to learn. This argument is no longer valid for
$\gamma\!\to\!\infty$ as $K(\gamma)\!\to\!\infty$ in this case.

\paragraph{Remark:}
TSP seems to be trivial in the AI$\mu$ model but non-trivial in
the AI$\xi$ model. The reason being that (\ref{fmydot}) just
implements an internal complete search as
$\mu(f)\!=\!\delta_{ff^{TSP}}$ contains all necessary information.
AI$\mu$ outputs from the very beginning, the exact minimum of $f^{TSP}$. This
''solution'' is, of course, unacceptable from performance
perspective. As long as we give no efficient approximation $\xi^c$
of $\xi$, we have not contributed anything to a solution of the
TSP by using AI$\xi^c$. The same is true for any other problem
where $f$ is computable and easily accessible. Therefore, TSP is not (yet)
a good example because all we have done is to replace a NP
complete problem with the uncomputable AI$\xi$ model or by a
computable AI$\xi^c$ model, for which we have said nothing about
computation time yet. It is simply an overkill to reduce simple
problems to AI$\xi$. TSP is a simple problem in this respect, until
we consider the AI$\xi^c$ model seriously. For the other examples,
where $f$ is inaccessible or complicated, AI$\xi^c$ provides a
true solution to the minimization problem as an explicit
definition of $f$ is not needed for AI$\xi$ and AI$\xi^c$.

\newpage
\section{Supervised Learning by Examples (EX)}\label{secEX}

The AI models provide a frame for reinforcement learning. The
environment provides a feedback $c$, informing the system about the
quality of its last output $y$; it assigns credit $c$ to output
$y$. In this sense, reinforcement learning is explicitly integrated
into the AI$\rho$ model. For $\rho\!=\!\mu$ it maximizes the true
expected credit, whereas the AI$\xi$ model is a universal,
environment independent, reinforcement learning algorithm.

There is another type of learning method: Supervised learning by
presentation of examples (EX). Many problems learned by this
method are association problems of the following type. Given some
examples $x\!\in\!R\subset\!X$, the system should reconstruct, from
a partially given $x'$, the missing or corrupted parts, i.e.
complete $x'$ to $x$ such that relation $R$ contains $x$. In many
cases, $X$ consists of pairs $(z,v)$, where $v$ is the possibly
missing part.

\paragraph{Applications/Examples:}
Learning functions by presenting $(z,f(z))$ pairs and asking for
the function value of $z$ by presenting $(z,?)$ also falls into
this category.

A basic example is learning properties of geometrical objects
coded in some way. E.g.\ if there are 18 different objects
characterized by their size (small or big), their colors (red,
green or blue) and their shapes (square, triange, circle), then
$(object,property)\!\in\!\!R$ if the $object$ possesses the
$property$. Here, $R$ is a relation which is not the graph of a
single valued function.

When teaching a child, by pointing to objects and saying ''this is
a tree'' or ''look how green'' or ''how beautiful'', one
establishes a relation of $(object,property)$ pairs in $R$.
Pointing to a (possibly different) tree later and asking ''what is
this ?'' corresponds to a partially given pair $(object,?)$, where
the missing part ''?'' should be completed by the
child saying ''tree''.

A final example we want to give is chess. We have seen that, in
principle, chess can be learned by reinforcement learning. In the
extreme case the environment only provides credit $c\!=\!1$ when
the system wins. The learning rate is completely inacceptable from
a practical point of view. The reason is the very low amount of
information feedback. A more practical method of teaching chess is
to present example games in the form of sensible
$(board\mbox{-}state,move)$
sequences. They contain information about legal and good moves
(but without any explanation). After several games have been presented, the
teacher could ask the system to make its own move by presenting
$(board\mbox{-}state,?)$ and then evaluate the answer of the system.

\paragraph{Supervised learning with the AI$\mu/\xi$ model:}
Let us define the EX model as follows: The environment presents
inputs
$x'_k = z_kv_k \equiv (z_k,v_k) \in R\!\cup\!(Z\!\times\!\{?\}) \subset
 Z\!\times\!(Y\!\cup\!\{?\}) = X'$
to the system in cycle $k$. The system is expected to output $y_{k+1}$
in the next cycle, which is evaluated with $c_{k+1}\!=\!1$ if $(z_k,y_{k+1})\!\in\!R$ and 0
otherwise. To simplify the discussion, an output $y_k$ is expected
and evaluated even when $v_k(\neq?)$ is given. To complete the
description of the environment, the probability distribution
$\mu_R(\pb{x'_1...x'_n})$ of the examples $x'_i$ (depending on $R$)
has to be given. Wrong examples should not occur, i.e.\ $\mu_R$
should be 0 if $x_i'\!\not\in\!R$ for some $1\!\leq\!i\!\leq\!n$.
The relations $R$ might also be probability distributed with
$\sigma(\pb R)$. The example prior probability in this case is
\beq\label{exmudef}
  \mu(\pb{x'_1...x'_n}) \;=\;
  \sum_R \mu_R(\pb{x'_1...x'_n})\!\cdot\!\sigma(\pb R)
\eeq
The knowledge of the valuation $c_k$ on output $y_k$
restricts the possible relations $R$, consistent with
$R(z_k,y_{k+1})\!=\!c_{k+1}$, where $R(z,y)\!:=\!1$ if $(z,y)\!\in\!R$ and 0
otherwise. The prior probability for the input sequence
$x_1...x_n$ if the output sequence is $y_1...y_n$, is
therefore
\beqn
  \mu^{AI}(y_1\pb x_1...y_n\pb x_n) \;=\;
  \sum_{R:\forall 1\leq i< n[R(z_i,y_{i+1})=c_{i+1}]}
  \mu_R(\pb{x'_1...x'_n})\!\cdot\!\sigma(\pb R)
\eeqn
where $x_i\!=\!c_ix'_i$ and $x'_{i-1}\!=\!z_iv_i$ with $v_i\!\in\!Y\!\cup\!\{?\}$.
In the I/O sequence $y_1x_1y_2x_2...=y_1c_1z_2v_2y_2c_2z_3v_3...$
the $c_1y_1$ are dummies, after which regular behaviour starts.

The AI$\mu$ model is optimal by construction of $\mu^{AI}$. For
computable prior $\mu_R$ and $\sigma$, we expect a near optimal
behavior of the universal AI$\xi$ model if $\mu_R$ additionally satisfies some
separability property. In the following, we give some motivation
why the AI$\xi$ model takes into account the supervisor
information contained in the examples and why it learns faster than by
reinforcement.

We keep $R$ fixed and assume
$\mu_R(x'_1...x'_n)\!=\!\mu_R(x'_1)\!\cdot...\cdot\!\mu_R(x'_n)\!\neq\!0
\Leftrightarrow x'_i\!\in\!R\!\cup\!(Z\!\times\!\{?\})\;\forall i$
to simplify the discussion. Short codes $q$ contribute mostly to
$\xi^{AI}(y_1\pb x_1...y_n\pb x_n)$. As $x'_1...x'_n$ is
distributed according to the computable probability distribution
$\mu_R$, a short code of $x'_1...x'_n$ for large enough $n$ is a
Huffman coding w.r.t.\ the distribution $\mu_R$. So we expect
$\mu_R$ and hence $R$ coded in the dominant contributions to
$\xi^{AI}$ in some way, where the plausible assumption was made
that the $y$ on the input tape do not matter. Much more than one
bit per cycle will usually be learned, hence, relation $R$ can be
learned in $n\!\ll\!K(R)$ cycles by appropriate examples. This
coding of $R$ in $q$ evolves independently of the feedbacks $c$.
To maximize the feedback $c_k$, the system has to learn to output
a $y_{k+1}$ with $(z_k,y_{k+1})\!\in\!R$. The system has to invent
a program extension $q'$ to $q$, which extracts $z_k$ from
$x_k\!=\!z_kv_k$ and searches for and outputs a $y_{k+1}$ with
$(z_k,y_{k+1})\!\in\!R$. As $R$ is already coded in $q$, $q'$ can
re-use this coding of $R$ in $q$. The size of the extension $q'$
is, therefore, of $O(1)$. To learn this $q'$, the system requires
feedback $c$ with information content of $O(1)\!=\!K(q')$ only.

Let us compare this with reinforcement learning, where only $x'_k\!=\!(z_k,?)$
pairs are presented. A coding of $R$ in a short code $q$ for
$x'_1...x'_n$ is of no use and will therefore be absent. Only the
credits $c$ force the system to learn $R$. $q'$ is therefore
expected to be of size $K(R)$. The information content in the
$c's$ must be of the order $K(R)$. In practice, there are often only very few
$c_k\!=\!1$ at the beginning of the learning phase and the
information content in $c_1...c_n$ is much less than $n$ bits. The
required number of cycles to learn $R$ by reinforcement is,
therefore, at least but in many cases much larger than $K(R)$.

Although AI$\xi$ was never designed or told to learn
supervised, it learns how to take advantage of the examples from
the supervisor.  $\mu_R$ and $R$ are learned from the examples, the
credits $c$ are not necessary for this process. The remaining task
of learning how to learn supervised is then a simple task of
complexity $O(1)$, for which the credits $c$ are necessary.

\section{Other AI Classes}\label{secOther}
\ifprivate
\begin{itemize}\parskip=0ex\parsep=0ex\itemsep=0ex
\item Function Inversion
\item Building analogies
\item Delayed SP
\item Artificial Life
\end{itemize}
\fi

\paragraph{Other aspects of intelligence:}
In AI, a variety of general ideas and methods have been developed.
In the last sections, we have seen how several problem classes can
be formulated within AI$\xi$. As we claim universality of the
AI$\xi$ model, we want to enlight which of, and how the other AI
methods are incorporated in the AI$\xi$ model, by looking its
structure. Some methods are directly included,
others are or should be emergent. We do not claim the following
list to be complete.

{\it Probability theory} and {\it utility theory} are the heart of
the AI$\mu/\xi$ models. The probabilities are the true/universal
behaviours of the environment. The utility function is what we
called total credit, which should be maximized. Maximization of an
expected utility function in a probabilistic environment is
usually called {\it sequential decision theory}, and is explicitly integrated
in full generality in our model. This includes probabilistic (a
generalization of deterministic) {\it reasoning}, where the
object of reasoning are not true or false statements, but the
prediction of the environmental behaviour. {\it Reinforcement
Learning}
is explicitly built in, due to the credits. Supervised learning is
an emergent phenomenon (section \ref{secEX}). {\it Algorithmic
information theory} leads us to use $\xi$ as a universal estimate
for the prior probability $\mu$.

For horizon $>\!1$, the alternative series of expectimax series
in (\ref{facydot}) and the process of selecting maximal
values can be interpreted as abstract {\it planning}. This expectimax
series also includes {\it informed search}, in the case of AI$\mu$, and {\it
heuristic search}, for AI$\xi$, where $\xi$ could be interpreted as
a heuristic for $\mu$. The minimax strategy of {\it game playing}
in case of AI$\mu$ is also subsumed. The AI$\xi$ model converges
to the minimax strategy if the environment is a minimax player but
it can also take advantage of environmental players with limited
rationality. {\it Problem solving} occurs (only) in the form of
how to maximize the expected future credit.

{\it Knowledge} is accumulated by AI$\xi$ and is stored in some
form not specified further on the working tape. Any kind of
information in any representation on the inputs $y$ is
exploited. The problem of {\it knowledge engineering} and
representation appears in the form of how to train the AI$\xi$
model. More practical aspects, like {\it language or image
processing} have to be learned by AI$\xi$ from scratch.

Other theories, like {\it fuzzy logic}, {\it possibility theory},
{\it Dempster-Shafer theory}, ... are partly outdated and partly
reducible to Bayesian probability theory \cite{Che85}. The
interpretation and effects of the evidence gap
$g\!:=\!1\!-\!\sum_{x_k}\xi(y\!x_{<k}y\!\pb x_k)\!>\!0$ in $\xi$ may
be similar to those in Dempster-Shafer theory. Boolean logical
reasoning about the external world plays, at best, an emergent
role in the AI$\xi$ model.

Other methods, which don't seem to be contained in the AI$\xi$ model
might also be emergent phenomena. The AI$\xi$ model has to
construct short codes of the environmental behaviour, the
AI$\xi^{\tilde t\tilde l}$ (see next section) has to construct
short action programs. If we would analyze and interpret these
programs for realistic environments, we might find some of the
unmentioned or unused or new AI methods at work in these
algorithms. This is, however, pure speculation at this point. More
important: when trying to make AI$\xi$ practically usable,
some other AI methods, like genetic algorithms or neural nets,
may be useful.

The main thing we wanted to point out is that the AI$\xi$ model
does not lack any important known property of intelligence or
known AI methodology. What {\it is} missing, however, are computational
aspects, which are addressed, in the next section.

\newpage
\section{Time Bounds and Effectiveness}\label{secTime}

\paragraph{Introduction:}
Until now, we have not bothered with the non-computability of the
universal probability distribution $\xi$. As all universal models
in this paper are based on $\xi$, they are not effective in this
form. In this section, We will outline how the previous models and
results can be modified/generalized to the time-bounded case.
Indeed, the situation is not as bad as it could be. $\xi$ and $C$
are enumerable and $\hh y_k$ is still approximable or computable
in the limit. There exists an algorithm, that will produce a
sequence of outputs eventually converging to the exact output $\hh
y_k$, but we can never be sure whether we have already reached it.
Besides this, the convergence is extremely slow, so this type of
asymptotic computability is of no direct (practical) use, but will
nevertheless, be important later.

Let $\tilde p$ be a program which calculates within a reasonable
time $\tilde t$ per cycle, a reasonable intelligent output, i.e.
$\tilde p(\hh x_{<k})\!=\!\hh y_{1:k}$. This sort
of computability assumption, that a general purpose computer of
sufficient power is able to behave in an intelligent way, is
the very basis of AI, justifying the
hope to be able to construct systems which eventually reach and outperform
human intelligence. For a contrary viewpoint see \cite{Pen89}. It
is not necessary to discuss here, what is meant by 'reasonable
time/intelligence' and 'sufficient power'. What we are interested
in, in this section, is whether there is a computable version
AI$\xi^{\tilde t}$ of the AI$\xi$ system which is superior or equal to any
$p$ with computation time per cycle of at most $\tilde t$.
With 'superior', we mean 'more intelligent', so what we
need is an order relation (like) (\ref{aiorder}) for intelligence.

The best result we could think of would be an AI$\xi^{\tilde t}$
with computation time $\leq\!\tilde t$ at least as intelligent as
any $p$ with computation time $\leq\!\tilde t$. If AI is possible
at all, we would have reached the final goal, the construction of
the most intelligent algorithm with computation $\leq\!\tilde t$.
Just as there is no universal measure in the set of computable
measures (within time $\tilde t$), such an AI$\xi^t$ may
neither exist.

What we can realistically hope to construct, is an AI$\xi^{\tilde
t}$ system of computation time $c\!\cdot\!\tilde t$ per cycle for
some constant $c$. The idea is to run all programs $p$ of length
$\leq\!\tilde l\!:=\!l(\tilde p)$ and time $\leq\!\tilde t$ per
cycle and pick the best output. The total computation time is
$2^{\tilde l}\!\cdot\!\tilde t$, hence $c=2^{\tilde l}$. This sort
of idea of 'typing monkeys' with one of them eventually writing
Shakespeare, has been applied in various forms and contexts in
theoretical computer science. The realization of this {\it best
vote} idea, in our case, is not straightforward and will be
outlined in this section. An idea related to this, is that of basing the
decision on the majority of algorithms. This 'democratic vote'
idea has been used in \cite{LiWa89,Vov92} for sequence prediction,
and is referred to as 'weighted majority' there.

\paragraph{Time limited probability distributions:}
In the literature one can find time limited versions of Kolmogorov
complexity \cite{Dal73,Ko86} and the time limited universal
semimeasure \cite{LiVi91,LiVi93}. In the following, we
utilize and adapt the latter and see how far we get. One way to define a
time-limited universal chronological semimeasure is
as a sum over all enumerable chronological semimeasures
computable within time $\tilde t$ and of size at most $\tilde l$
similar to the unbounded case (\ref{xirhodef}).
\beq\label{aixitl}
  \xi^{\tilde t\tilde l}(y\!\pb x_{1:n})
  \;:=\; \nq\sum_{\quad\rho\;:\;l(\rho)\leq\tilde l\;\wedge\;t(\rho)\leq\tilde t}
  \nq\nq 2^{-l(\rho)}\rho(y\!\pb x_{1:n})
\eeq
Let us assume that the true environmental prior probability $\mu^{AI}$
is equal to or sufficiently accurately approximated by a $\rho$ with
$l(\rho)\!\leq\!\tilde l$ and $t(\rho)\!\leq\!\tilde t$ with $\tilde
t$ and $\tilde l$ of reasonable size. There are several AI
problems that fall into this class. In function minimization of
section \ref{secFM}, the computation of $f$ and $\mu^{FM}$ are
usually feasible. In many cases, the sequences of section \ref{secSP}
which should be predicted, can be easily calculated when $\mu^{SP}$
is known. In a classifier problem, the
probability distribution $\mu^{CF}$, according to which examples
are presented, is, in many cases, also elementary. But not all AI
problems are of this 'easy' type. For the strategic games of section
\ref{secSG}, the environment is usually, itself, a highly
complex strategic player with a difficult to calculate $\mu^{SG}$
that is difficult to calculate,
although one might argue that the environmental player may have
limited capabilities too. But it is easy to think of a difficult
to calculate physical (probabilistic) environment like the
chemistry of biomolecules.

The number of interesting applications makes this restricted class
of AI problems, with time and space bounded environment
$\mu^{\tilde t\tilde l}$, worth being studied. Superscripts to a
probability distribution except for $\xi^{\tilde t\tilde l}$
indicate their length and maximal computation time. $\xi^{\tilde
t\tilde l}$ defined in (\ref{aixitl}), with a yet to be determined
computation time, multiplicatively dominates all $\mu^{\tilde
t\tilde l}$ of this type. Hence, an AI$\xi^{\tilde t\tilde l}$
model, where we use $\xi^{\tilde t\tilde l}$ as prior probability,
is universal, relative to all AI$\mu^{\tilde t\tilde l}$ models in
the same way as AI$\xi$ is universal to AI$\mu$ for all enumerable
chronological semimeasures $\mu$. The $\maxarg_{y_k}$ in
(\ref{ydotxi}) selects a $y_k$ for which $\xi^{\tilde t\tilde l}$
has the highest expected utility $C_{km_k}$, where $\xi^{\tilde
t\tilde l}$ is the weighted average over the $\rho^{\tilde t\tilde
l}$. $\hh y_k^{AI\xi^{\tilde t\tilde l}}$ is determined by a
weighted majority. We expect $AI\xi^{\tilde t\tilde l}$ to
outperform all (bounded) $AI\rho^{\tilde t\tilde l}$, analog to the
unrestricted case.

In the following we analyze the computability properties of
$\xi^{\tilde t\tilde l}$ and AI$\xi^{\tilde t\tilde l}$,
i.e.\ of $\hh y_k^{AI\xi^{\tilde t\tilde l}}$. To compute
$\xi^{\tilde t\tilde l}$ according to the definition
(\ref{aixitl}) we have to enumerate all chronological enumerable semimeasures
$\rho^{\tilde t\tilde l}$ of length $\leq\!\tilde l$
and computation time $\leq\!\tilde t$. This can be done similarly to
the unbounded case (\ref{ccsm1}-\ref{ccsm3}). All $2^{\tilde l}$
enumerable functions of length $\leq\!\tilde l$, computable within time
$\tilde t$ have to be converted to chronological probability
distributions. For this, one has to evaluate each function for
$|X|\!\cdot\!k$ different arguments. Hence,
$\xi^{\tilde t\tilde l}$ is computable within time\footnote{We
assume that a TM can be simulated by another in linear time.}
$
  t(\xi^{\tilde t\tilde l}(y\!\pb x_{1:k})) \!=\!
  O(|X|\!\cdot\!k\!\cdot\!2^{\tilde l}\!\cdot\!\tilde t)
$.
The computation time of $\hh y_k^{AI\xi^{\tilde t\tilde l}}$
depends on the size of $X$, $Y$ and $m_k$.
$\xi^{\tilde t\tilde l}$ has to be
evaluated $|Y|^{h_k}|X|^{h_k}$ times in (\ref{ydotxi}).
It is possible to
optimize the algorithm and perform the computation within time
\beq\label{tyaixi}
  t(\hh y_k^{AI\xi^{\tilde t\tilde l}}) \;=\;
  O(|Y|^{h_k}|X|^{h_k}\!\cdot\!2^{\tilde l}\!\cdot\!\tilde t)
\eeq
per cycle. If we assume that the computation time of $\mu^{\tilde
t\tilde l}$ is exactly $\tilde t$ for all arguments, the brute
force time $\bar t$ for calculating the sums and maxs in
(\ref{ydotrec}) is $\bar t(\hh y_k^{AI\mu^{\tilde t\tilde
l}})\!\geq\!|Y|^{h_k}|X|^{h_k}\!\cdot\!\tilde t$. Combining this
with (\ref{tyaixi}), we get
\beqn
  t(\hh y_k^{AI\xi^{\tilde t\tilde l}}) \;=\;
  O(2^{\tilde l}\!\cdot\!
  \bar t(\hh y_k^{AI\mu^{\tilde t\tilde l}}))
\eeqn
This result has the proposed structure, that there is a universal
AI$\xi^{\tilde t\tilde l}$ system with computation time
$2^{\tilde l}$ times the computation time of a special
AI$\mu^{\tilde t\tilde l}$ system.

Unfortunately, the class of AI$\mu^{\tilde t\tilde l}$ systems
with brute force evaluation of $\hh y_k$, according to
(\ref{ydotrec}) is completely uninteresting from a practical point
of view. E.g. in the context of chess, the above result says that
the AI$\xi^{\tilde t\tilde l}$ is superior within time $2^{\tilde
l}\!\cdot\!\tilde t$ to any brute force minimax strategy of computation time
$\tilde t$. Even if the factor of $2^{\tilde l}$ in computation
time would not matter, the AI$\xi^{\tilde t\tilde l}$ system is,
nevertheless practically useless, as a brute force minimax chess
player with reasonable time $\tilde t$ is a very poor player.

Note, that in the case of sequence prediction ($h_k\!=\!1$,
$|Y|\!=\!|X|\!=\!2$) the computation time of $\rho$ coincides with
that of $\hh y_k^{AI\rho}$ within a factor of 2. The class
AI$\rho^{\tilde t\tilde l}$ includes {\it all} non-incremental
sequence prediction algorithms of size $\leq\!\tilde l$ and
computation time $\leq\!\tilde t/2$. With non-incremental, we mean
that no information of previous cycles is taken into account for
the computation of $\hh y_k$ of the current cycle.

The shortcomings (mentioned and unmentioned ones) of this
approach are cured in the next subsection, by deviating from the
standard way of defining a timebounded $\xi$ as a sum over functions or
programs.

\paragraph{The idea of the best vote algorithm:}
A general cybernetic or AI system is a chronological program
$p(x_{<k})=y_{1:k}$. This form, introduced in section
\ref{secAIfunc}, is general enough to include any AI system (and
also less intelligent systems).
In the following, we are interested in programs $p$ of length
$\leq\!\tilde l$ and computation time $\leq\!\tilde t$ per cycle.
One important point in the time-limited setting is that $p$ should be
incremental, i.e. when computing $y_k$ in cycle $k$, the
information of the previous cycles stored on the working tape can
be re-used. Indeed, there is probably no practically interesting,
non-incremental AI system at all.

In the following, we construct a policy $p^\best$, or more
precisely, policies $p_k^\best$ for every cycle $k$ that
outperform all time and length limited AI systems $p$. In cycle k,
$p_k^\best$ runs all $2^{\tilde l}$ programs $p$ and selects the
one with the best output $y_k$. This is a 'best vote' type of
algorithm, as compared to the 'weighted majority' like algorithm of the
last subsection. The ideal measure for the quality of the output
would be the $\xi$ expected credit
\beq
 C_{km}(p|\hh y\!\hh x_{<k}) \;:=\; \sum_{q\in\hh Q_k}2^{-l(q)}C_{km}(p,q)
 \quad,\quad
  C_{km}(p,q) \;:=\; c(x_k^{pq})+...+c(x_m^{pq})
\eeq
The program $p$ which maximizes $C_{km_k}$ should be
selected. We have dropped the normalization $\cal N$ unlike in
(\ref{cxi}), as it is independent of $p$ and
does not change the order relation which we are solely interested
in here. Furthermore, without normalization, $C_{km}$ is enumerable,
which will be important later.

\paragraph{Extended chronological programs:}
In the (functional form of the) AI$\xi$ model it was convenient to
maximize $C_{km_k}$ over all $p\!\in\!\hh P_k$,
i.e. all $p$ consistent with the current history $\hh y\!\hh x_{<k}$.
This was no restriction, because for every
possibly inconsistent program $p$ there exists a program $p'\!\in\!\hh P_k$ consistent
with the current history and identical to $p$ for all future
cycles $\geq\!k$. For the time limited best vote algorithm
$p^\best$ it would be too restrictive to demand $p\!\in\!\hh
P_k$. To prove universality, one has to compare {\it all} $2^{\tilde l}$
algorithms in every cycle, not just the consistent ones. An
inconsistent algorithm may become the best one in later cycles.
For inconsistent programs we have to include the $\hh y_k$ into the
input, i.e. $p(\hh y\!\hh x_{<k})\!=\!y_{1:k}^p$
with $\hh y_i\!\neq\!y_i^p$ possible. For $p\!\in\!\hh P_k$ this
was not necessary, as $p$ knows the output $\hh y_k\equiv y_k^p$ in
this case. The $c_i^{pq}$ in the definition of $C_{km}$ are the
valuations emerging in the I/O sequence, starting with $\hh
y\!\hh x_{<k}$ (emerging from $p^\best$) and then continued
by applying $p$ and $q$ with $\hh y_i\!:=\!y_i^p$ for
$i\!\geq\!k$.

Another problem is that we need $C_{km_k}$ to select the best
policy, but unfortunately $C_{km_k}$ is uncomputable. Indeed, the
structure of the definition of $C_{km_k}$ is very similar to that
of $\hh y_k$, hence a brute force approach to approximate
$C_{km_k}$ requires too much computation time as for $\hh y_k$. We
solve this problem in a similar way, by supplementing each $p$ with
a program that estimates $C_{km_k}$ by $w_k^p$ within time
$\tilde t$. We combine the calculation of $y_k^p$ and $w_k^p$ and
extend the notion of a chronological program once again to
\beq\label{extprog}
  p(\hh y\!\hh x_{<k}) \;=\; w_1^py_1^p...w_k^py_k^p
\eeq
with chronological order $w_1^py_1^p\hh y_1\hh x_1
w_2^py_2^p\hh y_2\hh x_2...$.

\paragraph{Valid approximations:}
$p$ might suggest any output $y_k^p$ but it is not allowed to rate
it with an arbitrarily high $w_k^p$ if we want $w_k^p$ to be a reliable
criterion for selecting the best $p$. We demand that no policy is
allowed to claim that it is better than it actually is. We define
a (logical) predicate VA($p$) called {\it valid approximation}, which
is true if, and only if, $p$ always satisfies
$w_k^p\!\leq\!C_{km_k}(p)$, i.e. never overrates itself.
\beq\label{vadef}
  \mbox{VA}(p) \;\equiv\;
  \forall k\forall w_1^py_1^p\hh y_1\hh x_1...w_k^py_k^p :
  p(\hh y\!\hh x_{<k}) \!=\! w_1^py_1^p...w_k^py_k^p
  \Rightarrow
  w_k^p\!\leq\!C_{km_k}(p|\hh y\!\hh x_{<k})
\eeq
In the following, we restrict our attention to programs $p$, for which
VA($p$) can be proved in some formal axiomatic system.
A very important point is that $C_{km_k}$ is enumerable.
This ensures the existence of sequences of
program $p_1, p_2, p_3, ...$ for which VA($p_i$) can be proved and
$\lim_{i\to\infty}w_k^{p_i}\!=\!C_{km_k}(p)$
for all $k$ and all I/O sequences. The approximation is not
uniform in $k$, but this does not matter as the selected $p$ is allowed to change
from cycle to cycle.

Another possibility would be to consider only those $p$ which check
$w_k^p\!\leq\!C_{km_k}(p)$ online in every cycle, instead of
the pre-check VA($p$), either by constructing a proof (on the working
tape) for this special case, or it is already evident by the
construction of $w_k^p$. In cases where $p$ cannot guarantee
$w_k^p\!\leq\!C_{km_k}(p)$ it sets $w_k\!=\!0$ and, hence, trivially
satisfies $w_k^p\!\leq\!C_{km_k}(p)$. On the other hand, for these
$p$ it is also no problem to prove VA($p$) as one has simply to
analyze the internal structure of $p$ and recognize that $p$ shows
the validity internally itself, cycle by cycle, which is easy by
assumption on $p$. The cycle by cycle check is, therefore, a special
case of the pre-proof of VA($p$).

\paragraph{Effective intelligence order relation:}
In section \ref{secAIxi} we have introduced an intelligence order
relation $\succeq$ on AI systems, based on the expected credit
$C_{km_k}(p)$. In the following we need an order relation
$\succeq^c$ based on the claimed credit $w_k^p$ which might
be interpreted as an approximation to $\succeq$. We call $p$
{\it effectively more or equally intelligent} than $p'$ if
\bqa\label{effaiord}
  p\succeq^c\!p' \;:\Leftrightarrow\;
  \forall k\forall \hh y\!\hh x_{<k}
  \exists w_{1:n}w'_{1:n} : \\
  p(\hh y\!\hh x_{<k}) \!=\! w_1\!*...w_k\!* \;\wedge\;
  p'(\hh y\!\hh x_{<k}) \!=\! w_1'\!*...w_k'\!* \;\wedge\;
  w_k\!\geq\!w_k'
\eqa
i.e.\ if $p$ always claims higher credit estimate $w$ than $p'$.
$\succeq^c$ is a co-enumerable partial order relation on extended
chronological programs. Restricted to valid approximations
it orders the policies w.r.t.\ the quality of their outputs {\it
and} their ability to justify their outputs with high $w_k$.

\paragraph{The universal time bounded AI$\xi^{\tilde t\tilde l}$ system:}
In the following we, describe the algorithm $p^\best$ underlying
the universal time bounded AI$\xi^{\tilde t\tilde l}$ system. It
is essentially based on the selection of the best algorithms
$p_k^\best$ out of the time ${\tilde t}$ and length ${\tilde l}$
bounded $p$, for which there exists a proof of VA($p$) with length
$\leq\!l_P$.

\begin{enumerate}\parskip=0ex\parsep=0ex\itemsep=0ex
\item Create all binary strings of length $l_P$ and interpret each
as a coding of a mathematical proof in the same formal logic system in
which VA($\cdot$) has been formulated. Take those strings
which are proofs of VA($p$) for some $p$ and keep the
corresponding programs $p$.
\item Eliminate all $p$ of length $>\!\tilde l$.
\item Modify all $p$ in the following way: all output $w_k^py_k^p$
is temporarily written on an auxiliary tape. If $p$ stops in $\tilde t$
steps the internal 'output' is copied to the output tape. If $p$
does not stop after $\tilde t$ steps a stop is forced and $w_k\!=\!0$
and some arbitrary $y_k$ is written on the output tape. Let $P$ be
the set of all those modified programs.
\item Start first cycle: $k\!:=\!1$.
\item\label{pbestloop} Run every $p\!\in\!P$ on extended input
$\hh y\!\hh x_{<k}$, where all outputs are redirected to some auxiliary
tape:
$p(\hh y\!\hh x_{<k})\!=\!w_1^py_1^p...w_k^py_k^p$.
\item Select the program $p$ with highest claimed credit $w_k^p$:
$p_k^\best\!:=\!\maxarg_pw_k^p$.
\item Write $\hh y_k\!:=\!y_k^{p_k^\best}$ to the output tape.
\item Receive input $\hh x_k$ from the environment.
\item Begin next cycle: $k\!:=\!k\!+\!1$, goto step
\ref{pbestloop}.
\end{enumerate}

It is easy to see that the following theorem holds.

\paragraph{Main theorem:}
Let $p$ be any extended chronological (incremental) program like
(\ref{extprog}) of length $l(p)\!\leq\!\tilde l$ and computation
time per cycle $t(p)\!\leq\!\tilde t$, for which there exists a
proof of VA($p$) defined in (\ref{vadef}) of length $\leq\!l_P$.
The algorithm $p^\best$ constructed in the last subsection,
depending on $\tilde l$, $\tilde t$ and $l_P$ but not on $p$, is
effectively more or equally intelligent, according to $\succeq^c$
defined in (\ref{effaiord}) than any such $p$. The size of
$p^\best$ is $l(p^\best)\!=\!O(\ln(\tilde l\!\cdot\!\tilde
t\!\cdot\! l_P))$, the setup-time is
$t_{setup}(p^\best)\!=\!O(l_P\!\cdot\!2^{l_P})$, the computation
time per cycle is $t_{cycle}(p^\best)\!=\!O(2^{\tilde
l}\!\cdot\!\tilde t)$.

Roughly speaking, the theorem says, that if there exists a
computable solution to some AI problem at all, the explicitly
constructed algorithm $p^\best$ is such a solution. Although this
theorem is quite general, there are some limitations and open
questions which we discuss in the following.

\paragraph{Limitations and open questions:}
\begin{itemize}\parskip=0ex\parsep=0ex
\item Formally, the total computation time of $p^\best$ for cycles
$1...k$ increases linearly with $k$, i.e. is of order $O(k)$ with
a coefficient $2^{\tilde l}\!\cdot\!\tilde t$. The unreasonably
large factor $2^{\tilde l}$ is a well known drawback in
best/democratic vote models and will be taken without further comments, whereas the
factor ${\tilde t}$ can be assumed to be of reasonable size. If we
don't take the limit $k\!\to\!\infty$ but consider reasonable $k$,
the practical usefulness of the timebound on $p^\best$ is somewhat
limited, due to the additional additive constant
$O(l_P\!\cdot\!2^{l_P})$. It is much larger than
$k\!\cdot\!2^{\tilde l}\!\cdot\!\tilde t$ as typically
$l_P\!\gg\!l($VA$(p))\!\geq\!l(p)\!\equiv\!\tilde l$.
\item $p^\best$ is superior only to those $p$ which justify their
outputs (by large $w_k^p$). It might be possible that there are
$p$ which produce good outputs $y_k^p$ within reasonable time, but
it takes an unreasonably long time to justify their outputs by
sufficiently high $w_k^p$. We do not think that (from a certain
complexity level onwards) there are policies where the process of
constructing a good output is completely separated from some sort
of justification process. But this justification might not be
translatable (at least within reasonable time) into a reasonable
estimate of $C_{km_k}(p)$.
\item The (inconsistent) programs $p$ must be able to continue
strategies started by other policies. It might happen that a
policy $p$ steers the environment to a direction for which it is
specialized. A 'foreign' policy might be able to displace $p$
only between loosely bounded episodes. There is probably no
problem for factorizable $\mu$. Think of a chess game, where it is
usually very difficult to continue the game/strategy of a
different player. When the game is over, it is usually advantageous
to replace a player by a better one for the next game. There might
also be no problem for sufficiently separable $\mu$.
\item There might be (efficient) valid approximations $p$ for which
VA($p$) is true but not provable, or for which only a very long
($>\!l_P$) proof exists.
\end{itemize}

\paragraph{Remarks:}
\begin{itemize}\parskip=0ex\parsep=0ex
\item The idea of suggesting outputs and justifying them by proving
credit bounds implements one aspect of human thinking. There are
several possible reactions to an input. Each reaction possibly has
far reaching consequences. Within a limited time one tries to estimate the
consequences as well as possible. Finally,
each reaction is valued and the best one is selected. What
is inferior to human thinking is, that the estimates $w_k^p$ must
be rigorously proved and the proofs are constructed by blind
extensive search, further, that {\it all} behaviours $p$ of length
$\leq\!\tilde l$ are checked. It is inferior 'only' in the sense of
necessary computation time but not in the sense of the quality of
the outputs.
\item In practical applications there are often cases with
short and slow programs $p_s$ performing some task $T$, e.g.
the computation of the digits of $\pi$, for which there also exist
long and quick programs $p_l$ too. If it is not too difficult to
prove that this long program is equivalent to the short one, then it is
possible to prove $K(T)\!\leq\!l(p_s)$ within time $t(p_l)$.
Similarly, the method of proving bounds $w_k$ for $C_{km_k}$ can
give high lower bounds without explicitly executing these short
and slow programs, which mainly contribute to $C_{km_k}$.
\item Dovetailing all length and time-limited programs is a well
known elementary idea (typing monkeys). The crucial part
which has been developed here, is the selection criterion for the
most intelligent system.
\item By construction of AI$\xi^{\tilde t\tilde l}$ and due to the enumerability
of $C_{km_k}$, ensuring arbitrary close approximations of
$C_{km_k}$ we expect that the behaviour of AI$\xi^{\tilde t\tilde l}$
converges to the behaviour of AI$\xi$ in the limit $\tilde
t,\tilde l\!\to\!\infty$ in a sense.
\item Depending on what you know/assume that a program $p$ of size
$\tilde l$ and computation time per cycle $\tilde t$ is able to
achieve, the computable AI$\xi^{\tilde t\tilde l}$ model will have the
same capabilities. For the strongest assumption of the existence of a Turing
machine, which outperforms human intelligence, the AI$\xi^{\tilde
t\tilde l}$ will do too, within the same time frame up to a (unfortunately
very large) constant factor.
\end{itemize}

\newpage
\section{Outlook \& Discussion}\label{secOutlook}
This section contains some discussion of otherwise unmentioned
topics and some (more personal) remarks. It also serves as an outlook
to further research.

\paragraph{Miscellaneous:}
\begin{itemize}
\item In game theory \cite{Osb94} one often wants to model the situation of
      simultaneous actions, whereas the AI$\xi$ models has
      serial I/O. Simultaneity can be simulated by withholding the
      environment from the current system's output $y_k$, until
      $x_k$ has been received by the system. Formally, this means
      that $\xi(y\!x_{<k}y\!\pb x_k)$ is independent of $y_k$.
      The AI$\xi$ system is already of simultaneous type in an
      abstract view if the behaviour $p$ is interpreted as the action.
      In this sense, AI$\xi$ is the action $p^\best$ which maximizes
      the utility function (credit), under the assumption that the environment
      acts according to $\xi$. The situation is different from
      game theory as the environment is not modeled to be a second
      'player' that tries to optimize his own utility although it might
      actually be a rational player (see section \ref{secSG}).
\item In various examples we have chosen differently specialized
      input and output spaces $X$ and $Y$. It should be clear
      that, in principle, this is unnecessary, as large enough spaces $X$
      and $Y$, e.g. $2^{32}$ bit, serve every need and can always
      be Turing reduced to the specific presentation needed internally by the
      AI$\xi$ system itself. But it is clear that using a generic
      interface, such as camera and monitor for, learning
      tic-tac-toe for example, adds the task of learning vision and drawing.
\end{itemize}

\paragraph{Outlook:}
\begin{itemize}
\item Rigorous proofs for credit bounds are the major theoretical challenge are
      -- general ones as well as tighter bounds for
      special environments $\mu$. Of special importance are suitable (and
      acceptable) conditions to $\mu$, under which $\hh y_k$ and
      finite credit bounds exist for infinite $Y$, $X$ and $m_k$.
\item A direct implementation of the
      AI$\xi^{\tilde t\tilde l}$ model is ,at best, possible for toy
      environments due to the large factor $2^{\tilde l}$ in
      computation time. But there are other applications of the AI$\xi$ theory.
      We have seen in several examples how to integrate problem classes
      into the AI$\xi$ model. Conversely, one can downscale the
      AI$\xi$ model by using more restricted forms of $\xi$.
      This could be done in the same way as the theory of universal
      induction has been downscaled with many insights
      to the Minimum Description Length principle
      \cite{LiVi92,Ris89} or to the domain of finite automata \cite{Fed92}.
      The AI$\xi$ model might similarly serve as a super model or as the
      very definition of (universal unbiased) intelligence, from
      which specialized models could be derived.
\item With a reasonable computation time, the AI$\xi$ model
      would be a solution of AI (see next point if you disagree).
      The AI$\xi^{\tilde t\tilde l}$ model was the first step,
      but the elimination of the factor $2^{\tilde l}$ without giving up
      universality will (almost certainly) be a very difficult task.
      One could try to select programs $p$ and prove VA($p$) in a
      more clever way than by mere enumeration, to improve performance
      without destroying
      universality. All kinds of ideas like, genetic algorithms,
      advanced theorem provers and many more could be incorporated. But now we
      are in trouble. We seem to have transferred the AI
      problem just to a different level. This shift has some
      advantages (and also some disadvantages) but presents, in no way, a
      solution.
      Nevertheless, we want to stress that we have reduced the AI
      problem to (mere) computational questions.
      Even the most general other systems the author is aware of, depend on some
      (more than computational) assumptions about the
      environment or it is far from clear whether they are, indeed, universal and optimal.
      Although computational
      questions are themselves highly complicated, this reduction is a
      non-trivial result. A formal theory of something, even if
      not computable, is often a great step toward solving a
      problem and has also merits of its own, and AI should not be different (see previous item).
\item Many researchers in AI believe that intelligence is something
      complicated and cannot be condensed into a few formulas.
      It is more a combining of enough {\it methods} and much explicit
      {\it knowledge} in the right way. From a theoretical point of
      view, we disagree as the AI$\xi$ model is simple and seems to serve all
      needs. From a practical point of view we agree to the following extent.
      To reduce the computational burden one should
      provide special purpose algorithms ({\it methods}) from the
      very beginning, probably many of them related to reduce
      the complexity of the input and output spaces $X$ and $Y$ by
      appropriate preprocessing {\it methods}.
\item There is no need to incorporate extra {\it knowledge} from the very
      beginning. It can be presented in the first few cycles in
      {\it any} format. As long as the algorithm to interpret the data
      is of size $O(1)$, the AI$\xi$ system will 'understand' the data
      after a few cycles (see section \ref{secEX}). If the
      environment $\mu$ is complicated but extra knowledge
      $z$ makes $K(\mu|z)$ small, one can show that the bound
      (\ref{eukdist}) reduces to $\1d2\ln 2\!\cdot\!K(\mu|z)$
      when $x_1\!\equiv\!z$, i.e.\
      when $z$ is presented in the first cycle. The special
      purpose algorithms could be presented in $x_1$, too, but it
      would be cheating to say that no special purpose algorithms
      had been implemented in AI$\xi$. The boundary between
      implementation and training is unsharp in the AI$\xi$ model.
\item We have not said much about the training
      process itself, as it is not specific to the AI$\xi$ model
      and has been discussed in literature in various forms and
      disciplines. A serious discussion would be out of place.
      To repeat a truism, it is, of course,
      important to present enough knowledge $x'_k$ and evaluate
      the system output $y_k$ with $c_k$ in a reasonable way.
      To maximize the information content in the credit, one should
      start with simple tasks and give positive reward
      $c_k\!=\!1$ to approximately half of the outputs $y_k$.
\end{itemize}

\paragraph{The big questions:}
This subsection is devoted to the {\it big} questions of AI in
general and the AI$\xi$ model in particular with a personal touch.

\begin{itemize}
\item There are two possible objections to AI in general and,
      therefore, also against AI$\xi$ in particular we want
      to comment on briefly. Non-computable physics (which is not too
      weird) could make Turing computable AI impossible. As at least the
      world that is relevant for humans seems mainly to be computable
      we do not believe that it is necessary to integrate non-computable
      devices into an AI system. The (clever and nearly convincing) 'G\"odel'
      argument by Penrose \cite{Pen89} that non-computational physics
      {\it must} exist and {\it is} relevant to the brain, has (in our opinion convincing)
      loopholes.
\item A more serious problem is the evolutionary information
      gathering process. It has been shown that the
      'number of wisdom' $\Omega$ contains a very compact
      tabulation of $2^n$ undecidable problems in its very first
      $n$ binary digits \cite{Cha91}. $\Omega$ is only enumerable
      with computation time increasing more rapidly with $n$, than any
      recursive function.
      The enormous computational power of evolution could
      have developed and coded something like $\Omega$ into
      our genes, which significantly guides human reasoning.
      In short: Intelligence could be something complicated
      and evolution toward it from an even cleverly designed
      algorithm of size $O(1)$ could be too slow. As evolution has
      already taken place, we could add the information from our
      genes or brain structure to any/our AI system, but this means that
      the important part is still missing and a simple formal definition
      of AI is principally impossible.
\item For the probably {\it biggest question} about {\it consciousness}
      we want to give a physical analogy. Quantum (field) theory is
      the most accurate and universal physical theory ever
      invented. Although already developed in the 1930ies the {\it
      big} question regarding the interpretation of the wave function collapse
      is still open. Although extremely interesting from a
      philosophical point of view, it is completely irrelevant from
      a practical point of view\footnote{In the theory of everything, the
      collapse might become of 'practical' importance and must or will be
      solved.}.
      We believe the same to be true
      for {\it consciousness} in the field of Artificial
      Intelligence. Philosophically highly interesting but
      practically unimportant. Whether consciousness {\it will} be
      explained some day is another question.
\end{itemize}

\newpage
\section{Conclusions}\label{secCon}
All tasks which require intelligence to be solved can naturally be
formulated as a maximization of some expected utility in the
framework of agents. We gave a functional (\ref{pbestfunc}) and an
iterative (\ref{ydotrec}) formulation of such a decision theoretic
agent, which is general enough to cover all AI problem classes,
as has been demonstrated by several examples. The main remaining
problem is the unknown prior probability distribution $\mu^{AI}$
of the environment(s). Conventional learning algorithms are
unsuitable, because they can neither handle large (unstructured)
state spaces, nor do they converge in the theoretically minimal
number of cycles, nor can they handle non-stationary environments
appropriately. On the other hand, the universal semimeasure $\xi$
(\ref{xidef}), based on ideas from algorithmic information theory,
solves the problem of the unknown prior distribution for induction
problems. No explicit learning procedure is necessary, as $\xi$
automatically converges to $\mu$. We unified the theory of
universal sequence prediction with the decision theoretic agent by
replacing the unknown true prior $\mu^{AI}$ by an appropriately
generalized universal semimeasure $\xi^{AI}$. We gave strong
arguments that the resulting AI$\xi$ model is the most
intelligent, parameterless and environmental/application independent model
possible. We defined an intelligence order relation
(\ref{aiorder}) to give a rigorous meaning to this claim.
Furthermore, possible solutions to the horizon problem have been
discussed. We outlined for a number of problem classes in sections
\ref{secSP}--\ref{secEX}, how the AI$\xi$ model can solve them.
They include sequence prediction, strategic games, function
minimization and, especially, how AI$\xi$ learns to learn
supervised. The list could easily be extended to other problem
classes like classification, function inversion and many others.
The major drawback of the AI$\xi$ model is that it is
uncomputable, or more precisely, only asymptotically computable,
which makes an implementation impossible. To overcome this
problem, we constructed a modified model AI$\xi^{\tilde t\tilde
l}$, which is still effectively more intelligent than any other
time $\tilde t$ and space $\tilde l$ bounded algorithm. The
computation time of AI$\xi^{\tilde t\tilde l}$ is of the order
$\tilde t\!\cdot\!2^{\tilde l}$. Possible further research has
been discussed. The main directions could be to prove general
and special credit bounds, use AI$\xi$ as a super model and
explore its relation to other specialized models and finally
improve performance with or without giving up universality.

\newpage
\addcontentsline{toc}{section}{Literature}
\parskip=0ex plus 1ex minus 1ex

\end{document}